\documentclass[journal]{IEEEtran}
\usepackage{amsmath,amsfonts}
\usepackage{amssymb}
\usepackage{algorithmic}
\usepackage{algorithm}
\usepackage{textcomp}
\usepackage{stfloats}
\usepackage{url}
\usepackage{verbatim}
\usepackage{graphicx}
\usepackage{cite}
\usepackage[T1]{fontenc} 
\usepackage{threeparttable} 
\usepackage{multirow} 
\usepackage{booktabs}
\usepackage{makecell} 
\usepackage{subcaption}
\usepackage[colorlinks,linkcolor=blue]{hyperref} 

\usepackage{pifont}
\usepackage{array}
\usepackage{color}
\usepackage{graphics}
\usepackage{xcolor}
\newcommand{\etal}{\emph{et~al.~}}
\newcommand{\ie}{\emph{i.e.},~}
\newcommand{\wrt}{\emph{w.r.t.~}}
\newcommand{\eg}{\emph{e.g.},~}
\newcommand{\etc}{\emph{etc}}
\newcommand{\checked}{\checkmark}

\begin{document}

\title{A Comprehensive Study on the Robustness of Image Classification and Object Detection in Remote Sensing: Surveying and Benchmarking}

\author{Shaohui~Mei,~\IEEEmembership{Senior Member,~IEEE},
        Jiawei~Lian,~\IEEEmembership{Graduate Student Member,~IEEE},
        Xiaofei~Wang, 
        Yuru~Su, 
        Mingyang~Ma,~\IEEEmembership{Graduate Student Member,~IEEE},
        and~Lap-Pui~Chau,~\IEEEmembership{Fellow,~IEEE}

\thanks{This work was supported in part by the National Natural Science Foundation of China (62171381 and 62201445). (Corresponding author: Shaohui Mei.)}

\thanks{Shaohui Mei, Jiawei Lian, Xiaofei Wang, Yuru Su, and Mingyang Ma are with the School of Electronics and Information, Northwestern Polytechnical University, Xi'an 710129, China (Email: meish@nwpu.edu.cn; lianjiawei@mail.nwpu.edu.cn; wangxiaofei2022@mail.nwpu.edu.cn; suyuru\underline{~}nwpu@mail.nwpu.edu.cn; mamingyang@mail.nwpu.edu.cn).

Lap-Pui Chau is with the Department of Electronic and Information Engineering, The Hong Kong Polytechnic University, Hong Kong, China (lap-pui.chau@polyu.edu.hk).
}
}

\markboth{Journal of \LaTeX\ Class Files,~Vol.~14, No.~8, August~2021}%
{Shell \MakeLowercase{\textit{et al.}}: A Sample Article Using IEEEtran.cls for IEEE Journals}


\maketitle

\begin{abstract}
Deep neural networks (DNNs) have found widespread applications in interpreting remote sensing (RS) imagery. 
However, it has been demonstrated in previous works that DNNs are susceptible and vulnerable to different types of noises, particularly adversarial noises. 
Surprisingly, there has been a lack of comprehensive studies on the robustness of RS tasks, prompting us to undertake a thorough survey and benchmark on the robustness of DNNs in RS.
This manuscript conducts a thorough study of both the natural robustness and adversarial robustness of DNNs in RS tasks. 
Specifically, we systematically and extensively survey the robustness of DNNs from various perspectives such as noise type, attack domain, attacker's knowledge, \etc., encompassing typical applications such as object detection, image classification, \etc.
Building upon this foundation, we further develop a rigorous framework for testing the robustness of models, which entails the construction of noised datasets, robustness testing, and evaluation. 
Under the proposed framework, we perform a meticulous and systematic examination of the robustness of typical deep learning algorithms in the context of object detection and image classification applications.
Through survey and benchmark, we uncover insightful and intriguing findings, which shed light on the relationship between adversarial noise crafting and model training, yielding a deeper understanding of the susceptibility and limitations of various DNNs-based models, and providing guidance for the development of more resilient and robust models.
\end{abstract}


\section{Introduction}
\label{Section1}

The proliferation of remote sensing (RS) technologies has remarkably augmented the volume and fidelity of RS imagery, which is critical and influential for characterizing diverse features of the earth's surface. 
As a consequence, the automated and intelligent processes of satellite or aerial images have become indispensable for earth observation and analysis. 
The significance of RS image (RSI) interpretation such as image classification and object detection is paramount and extends to a multitude of applications, encompassing but not limited to environmental monitoring, intelligent transportation, urban planning, and disaster management.
In response to the pressing demand for automated analysis and comprehension of optical RSIs, there has been a surge in the development of diverse techniques for aerial detection \cite{mei2023rotation,li2020object,li2022deep,hou2022refined} over the past few years.

\begin{figure}[!t]
  \centering
  \includegraphics[width=0.999\linewidth]{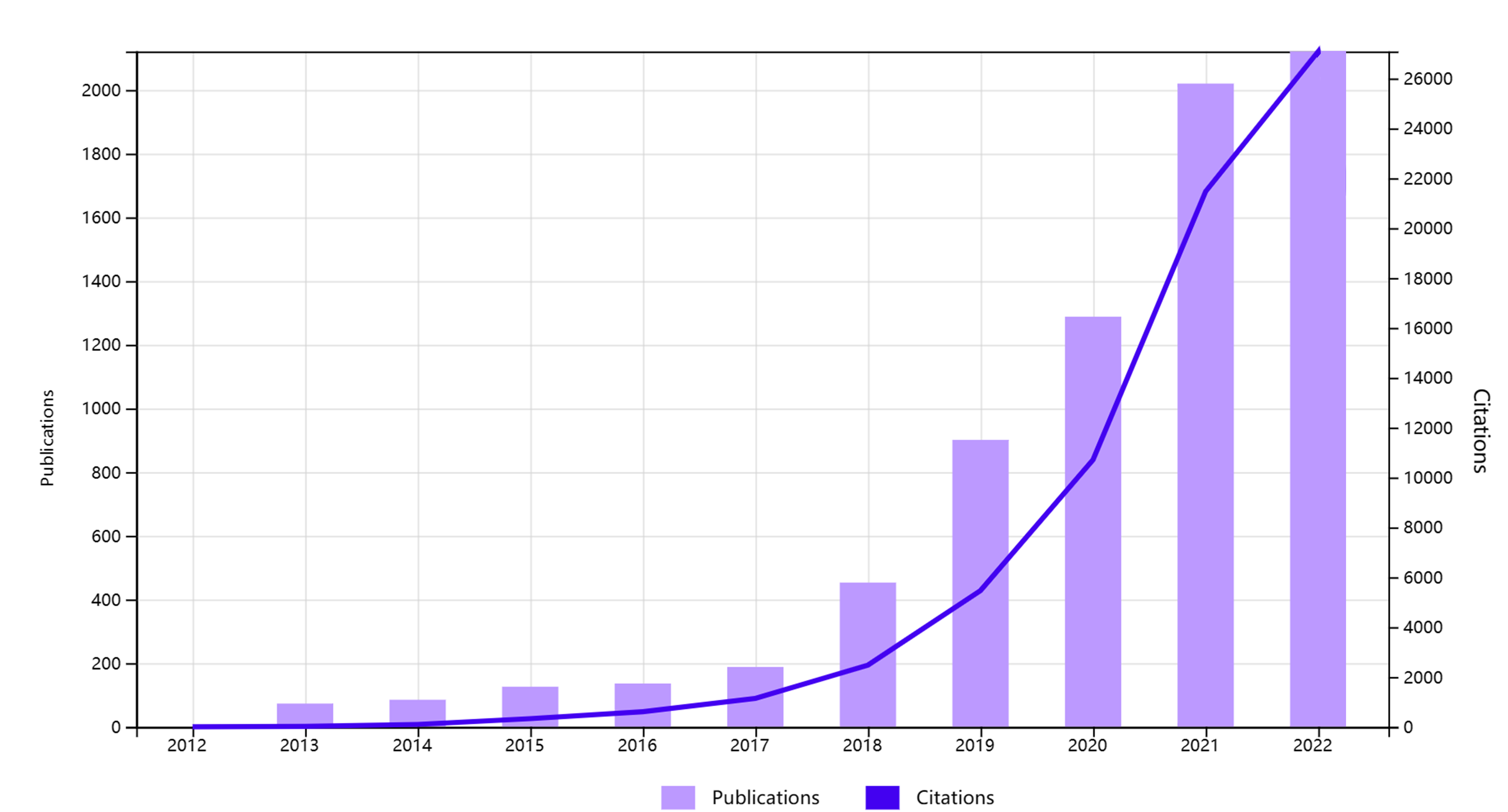}
  \caption{The temporal progression of publications and citations related to adversarial attacks. Data from Web of Science.}
  \label{fig:publications&citations}
\end{figure}

In recent years, algorithms based on deep learning (DL) technologies have emerged as the forerunners in top accuracy benchmark for a range of visual recognition tasks, \eg image classification \cite{he2016deep,krizhevsky2017imagenet}, object detection \cite{ren2015faster,redmon2018yolov3}, semantic segmentation \cite{he2017mask,ronneberger2015u}, \etc., owing to the remarkable feature representation capability of deep neural networks (DNNs). 
As a natural progression, DNNs have been widely adopted for the processing of optical RS imagery, with particular emphasis on image classification and object detection tasks. 
Undoubtedly, DNNs-based models \cite{zhang2023cof,li2023lightweight,cheng2021feature,lian2022fast} have emerged as a dominant approach, surpassing the performance of previous traditional methods by a significant margin.

\begin{figure}[!t]
  \centering
  \includegraphics[width=0.999\linewidth]{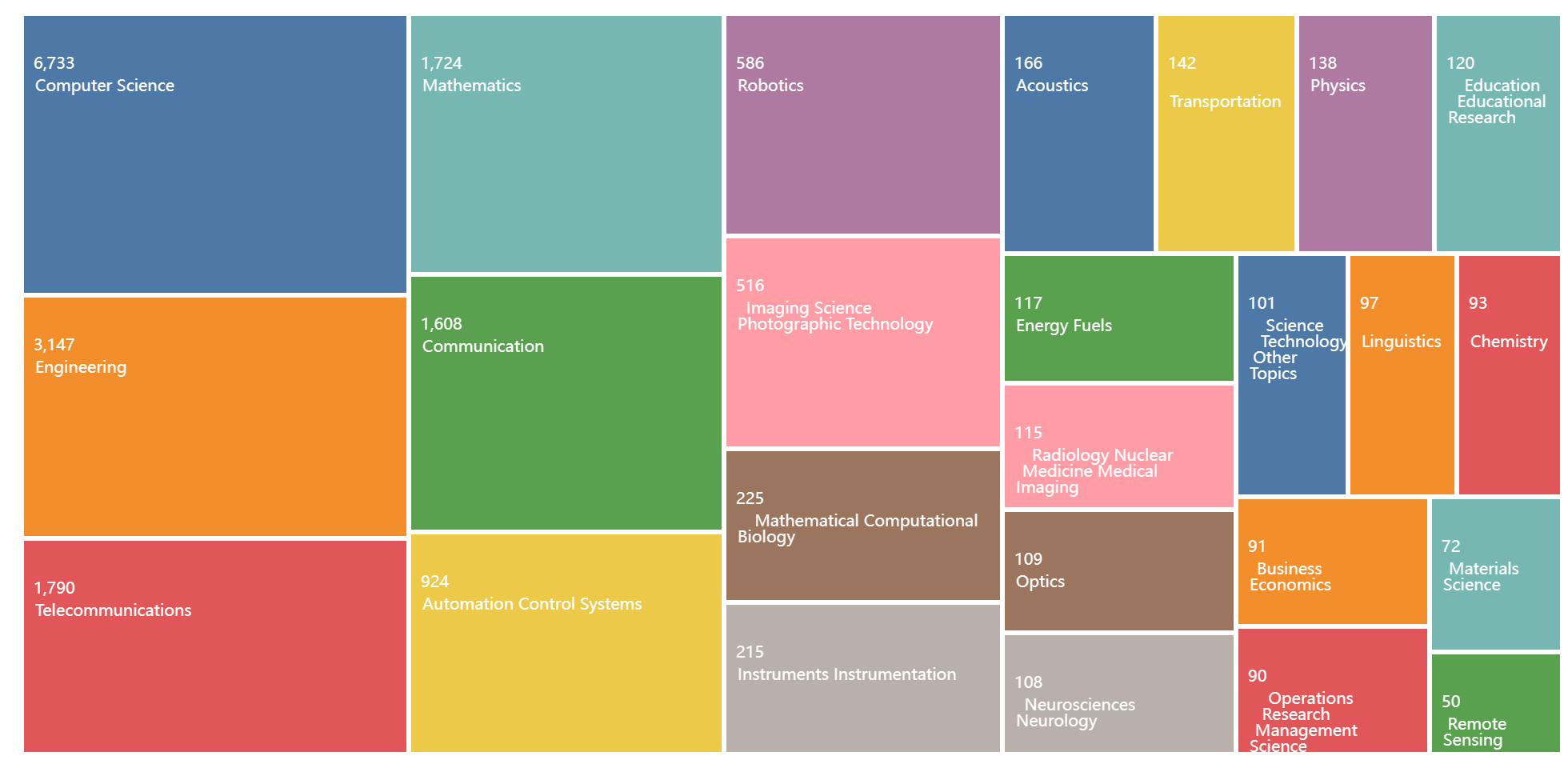}
  \caption{Research areas have been threatened by adversarial attacks. Data from Web of Science.}
  \label{fig:research_areas}
\end{figure}

However, good fortune brings misfortune on its train.
The utilization of DL in intelligent recognition brings forth notable advantages, yet it also introduces substantial security concerns. The black-box nature of DL has been the subject of critique due to its inherent lack of interpretability and transparency.
Furthermore, the susceptibility and vulnerability of DL models to adversarial examples have garnered significant attention within the academic community, prompting questions regarding the veracity of these models as reliable predictors. 
As a result, there are growing concerns that these models may merely be clever "Hans," achieving acceptable outcomes via flawed methods, which undermines the credibility and trustworthiness of DNNs-based systems.
The temporal progression of publications and citations related to adversarial attacks is shown in Fig. \ref{fig:publications&citations}.
Previous works \cite{szegedy2014intriguing,ian2015explaining} have demonstrated that DNNs are susceptible and vulnerable to adversarial examples, which involve the addition of carefully crafted imperceptible perturbations to benign images that can lead to erroneous predictions and pose a significant threat to both digital and physical applications \cite{kurakin2018adversarial,lian2023contextual,dong2018boosting,shi2022query} of DL. 
The research areas that have been threatened by adversarial attacks are detailed and exhibited in Fig. \ref{fig:research_areas}.
Furthermore, studies \cite{hendrycks2021natural,taori2020measuring,hendrycks2019benchmarking,hendrycks2021many} have also shown that DNNs can be easily disturbed by natural noises, indicating that DL systems are not inherently secure and robust. 

The phenomena mentioned above underscore the need for delving into the mechanism of adversarial attacks and improving the resilience and reliability of DL systems. 
In addition, it is beneficial to comprehensively benchmark the robustness of DNNs for better understanding and developing robust DNNs-based models, while none of the public surveys and benchmarks provide a comprehensive study on the robustness of image classification and object detection in RS.
We summarize the existing surveys and benchmarks as shown in Table \ref{table:taxonomy}.
Specifically, most existing related works \cite{wei2022physical,wu2022backdoorbench,tang2021robustart,wu2023adversarial,dong2020benchmarking,liu2023comprehensive,chakraborty2018adversarial,akhtar2018threat,qiu2019review,huang2020survey,akhtar2021advances,pintor2023imagenet,mi2023adversarial,zhuo2022attack,goldblum2022dataset,deng2021deep,machado2021adversarial,serban2020adversarial,sharif2019general} focus on surveying and benchmarking in computer vision (CV).
Only a few attempts \cite{wei2022physically,xu2022ai,kazmi2023adversarial} involve RS tasks.
For example, wei \etal \cite{wei2022physically} surveyed physical adversarial attacks and defenses in CV and briefly reviewed physical attacks in RS.
In \cite{xu2022ai}, the authors discussed the challenges and future trends of AI security for geoscience and RS while without further study on the robustness of DNNs-based methods in optical RSIs.
Work \cite{kazmi2023adversarial} attempts to comprehensively analyze the diversity of adversarial attacks in the context of autonomous aerial imaging and provides a literature review of adversarial attacks on aerial imagery processing but without further analysis of models' robustness.

Driven by the above requirements, this manuscript presents a comprehensive and rigorous study on the robustness of DNNs-based models and examines both natural and adversarial robustness in RS field. 
Technically, the study systematically explores the robustness of DNNs from multiple perspectives, including the type of noise, attack domain, and the attacker's knowledge. 
Moreover, it encompasses a wide range of applications such as object detection and image classification. 
To establish a solid foundation, we develop an elaborate framework that involves the creation of noised datasets, rigorous robustness testing, and thorough evaluation. 
Employing this framework, we conduct meticulous and systematic experiments on the robustness of common DL algorithms within the context of object detection and image classification tasks. 
Through widespread surveying and benchmarking, we reveal insightful and captivating discoveries that illuminate the intricate relationship between the crafting of adversarial noise and model training. 
These findings deepen our understanding of the vulnerability and constraints associated with various DNNs-based models, while also offering valuable guidance for the development of more resilient and robust models.

\begin{table}[t!]
\scriptsize
    \renewcommand{\arraystretch}{1.25}
\caption{Existing works related to survey and benchmark on models' robustness}
\label{table:taxonomy}
\centering
\setlength{\tabcolsep}{0.75mm}
\begin{threeparttable}
\begin{tabular*}{\hsize}{cccccccccc}
\hline\hline
\thead{\multirow{2}{*}{}} & \multirow{2}{*}{Year} & \multirow{2}{*}{Works} & \multirow{2}{*}{Journals}                                      & \multicolumn{2}{c}{Survey}       & \multicolumn{4}{c}{Benchmark} 
\\
                         & & & & CV & RS & IC & OD & AR & NR \\\hline
\multirow{20}{*}{Surveys}   & 2018                  & \cite{akhtar2018threat}          & IEEE Access                                                    & \checked               & \ding{53}              & \ding{53}                    & \ding{53}                & \ding{53}                      & \ding{53}                  \\
                            & 2018                  & \cite{huang2020survey}           & Computer Science Review                                        & \checked               & \ding{53}              & \ding{53}                    & \ding{53}                & \ding{53}                      & \ding{53}                  \\
                            & 2018                  & \cite{chakraborty2018adversarial}     & arXiv                                                          & \checked               & \ding{53}              & \ding{53}                    & \ding{53}                & \ding{53}                      & \ding{53}                  \\
                            & 2019                  & \cite{qiu2019review}             & Applied Science                                                & \checked               & \ding{53}              & \ding{53}                    & \ding{53}                & \ding{53}                      & \ding{53}                  \\
                            & 2020                  & \cite{serban2020adversarial}          & ACM Computing Surveys                                          & \checked               & \ding{53}              & \ding{53}                    & \ding{53}                & \ding{53}                      & \ding{53}                  \\
                            & 2021                  & \cite{akhtar2021advances}          & IEEE Access                                                    & \checked               & \ding{53}              & \ding{53}                    & \ding{53}                & \ding{53}                      & \ding{53}                  \\
                            & 2021                  & \cite{machado2021adversarial}         & ACM Computing Surveys                                          & \checked               & \ding{53}              & \ding{53}                    & \ding{53}                & \ding{53}                      & \ding{53}                  \\
                            & 2021                  & \cite{deng2021deep}            & TII                    & \checked               & \ding{53}              & \ding{53}                    & \ding{53}                & \ding{53}                      & \ding{53}                  \\
                            & 2022                  & \cite{wang2022survey}            & arXiv                                                          & \checked               & *              & \ding{53}                    & \ding{53}                & \ding{53}                      & \ding{53}                  \\
                            & 2022                  & \cite{huayu2022survey}              & INJOIT         & \checked               & \ding{53}              & \ding{53}                    & \ding{53}                & \ding{53}                      & \ding{53}                  \\
                            & 2022                  & \cite{sharma2022adversarial}       & Artificial Intelligence Review                                 & \checked               & \ding{53}              & \checked                    & \ding{53}                & \checked                      & \ding{53}                  \\
                            & 2022                  & \cite{goldblum2022dataset}        & TPAMI & \checked               & \ding{53}              & \ding{53}                    & \ding{53}                & \ding{53}                      & \ding{53}                  \\
                            & 2022                  & \cite{zhuo2022attack}            & TII                    & \ding{53}               & \ding{53}              & \ding{53}                    & \ding{53}                & \ding{53}                      & \ding{53}                  \\
                            & 2022                  & \cite{sharma2022adversarial}          & arXiv                                                          & \checked               & *              & \ding{53}                    & \ding{53}                & \ding{53}                      & \ding{53}                  \\
                            & 2022                  & \cite{wei2022physical}             & arXiv                                          & \checked               & \ding{53}              & \ding{53}                    & \ding{53}                & \ding{53}                      & \ding{53}                  \\
                            & 2022                  & \cite{wei2022physically}             & arXiv                                                          & \checked               & *              & \ding{53}                    & \ding{53}                & \ding{53}                      & \ding{53}                  \\
                            & 2022                  & \cite{xu2022ai}              & arXiv                                                          & *               & \checked              & \ding{53}                    & \ding{53}                & \ding{53}                      & \ding{53}                  \\
                            & 2022                  & \cite{mi2023adversarial}              & Neurocomputing                                                 & \checked               & \ding{53}              & \ding{53}                    & \ding{53}                & \ding{53}                      & \ding{53}                  \\
                            & 2023                  & \cite{li2023trustworthy}              & ACM Computing Surveys                                          & *               & \ding{53}              & \ding{53}                    & \ding{53}                & \ding{53}                      & \ding{53}                  \\
                            & 2023                  & \cite{wu2023adversarial}              & arXiv                                                          & \checked               & \ding{53}              & \ding{53}                    & \ding{53}                & \ding{53}                      & \ding{53}                  \\                            
                            & 2023                  & \cite{kazmi2023adversarial}              & ICAI                                                          & *               & \checked              & \ding{53}                    & \ding{53}                & \ding{53}                      & \ding{53}                  \\\hline
\multirow{9}{*}{Benchmarks} & 2020                  & \cite{dong2020benchmarking}            & CVPR                                                           & \ding{53}               & \ding{53}              & \checked                    & \ding{53}                & \checked                      & \ding{53}                  \\
                            & 2021                  & \cite{tang2021robustart}            & arXiv                                                          & \ding{53}               & \ding{53}              & \checked                    & \ding{53}                & \checked                      & \checked                  \\
                            & 2022                  & \cite{labarbarie2022benchmarking}      & IJCAI                                                          & \ding{53}               & \ding{53}              & \ding{53}                    & \checked                & \checked                      & \ding{53}                  \\
                            & 2022                  & \cite{wu2022backdoorbench}              & NIPS                                                           & \ding{53}               & \ding{53}              & \checked                    & \ding{53}                & \checked                      & \ding{53}                  \\
                            & 2022                  & \cite{hingun2022reap}          & arXiv                                                          & \ding{53}               & \ding{53}              & \ding{53}                    & \checked                & \checked                      & \ding{53}                  \\
                            & 2022                  & \cite{pintor2023imagenet}          & Pattern Recognition                                            & \ding{53}               & \ding{53}              & \checked                    & \ding{53}                & \checked                      & \ding{53}                  \\
                            & 2023                  & \cite{liu2023comprehensive}             & arXiv                                                   & \ding{53}               & \ding{53}              & \checked                    & \ding{53}                & \checked                      & \checked                  \\
                            & 2023                  & \cite{guo2023comprehensive}             & Pattern Recognition                                                             & \ding{53}               & \ding{53}              & \checked                    & \ding{53}                & \checked                      & \ding{53}                  \\
                            & 2023                  & \cite{dong2023benchmarking}            & CVPR                                                           & \ding{53}               & \ding{53}              & \ding{53}                    & \checked                & \ding{53}                      & \checked                  \\\hline
Both & 2023 & Ours & -                                                                                                                      & \checked               & \checked              & \checked                    & \checked                & \checked                      & \checked                 
\\                   
\hline\hline
\end{tabular*}
    \begin{tablenotes}
        \normalsize      
        \item "*" represents a brief introduction. “CV” "RS" "IC" "OD" "AR" and "NR" represent computer vision,	remote sensing,	image classification, object detection, adversarial robustness, and natural robustness, respectively.
    \end{tablenotes}
\end{threeparttable}
\end{table}

In summary, the main contributions of this article are four-fold as follows:

\begin{itemize}

    \item \textbf{Comprehensive survey.} 
    We conduct a comprehensive survey to systematically explore and assess the robustness of DNNs-based models.
    Our investigation encompasses diverse perspectives, including but not limited to the type of noise, attack domain, and the level of the attacker's knowledge. 
    Moreover, we extend our analysis to cover typical applications, such as object detection, image classification, \etc.
    
    \item \textbf{Rigorous benchmark.} 
    Building upon the foundation of the thorough survey, we also devise a rigorous framework that facilitates the evaluation of model robustness. 
    This framework includes several key components, namely the construction of datasets with added noise\footnote{\url{https://github.com/wangxiaofei2022/Robustness-Evaluation}}, robustness testing procedures, and robustness evaluation methods.
    
    \item \textbf{Systematical experiments.}  
    By employing the proposed framework, we meticulously and systematically investigate the robustness of typical DL algorithms within the specific domains of object detection and image classification. 
    Our examination involves rigorous testing and evaluation procedures, allowing us to gain comprehensive insights into the resilience of these algorithms under various conditions (see Table \ref{table:top5} for the top-5 robust models against different noises). 
    
    \item \textbf{In-depth analysis.} 
    Through extensive surveying and rigorous benchmarking, we have derived insightful and intriguing findings that shed light on the potential connection between adversarial perturbation generation and model training. 
    These findings contribute to a deeper understanding of the sensitivity and vulnerability exhibited by various DNNs-based models across different tasks. 
    By uncovering these relationships, our study offers valuable insights into the robustness of DNNs in the face of adversarial attacks and provides a foundation for developing more resilient and robust models in the future.
    
\end{itemize}

\begin{table*}[t!]
\footnotesize
    \renewcommand{\arraystretch}{1.25}
\caption{Top-5 robust models against different noises.}
\label{table:top5}
\centering
\setlength{\tabcolsep}{0.85mm}
\begin{threeparttable}
\begin{tabular*}{\hsize}{cccccccc}
\hline\hline
\thead{\multirow{2}{*}{\textbf{Task}}}          & \multirow{2}{*}{\textbf{Noises}}         & \multirow{2}{*}{\textbf{Datasets/Attacks}} & \multicolumn{5}{c}{\textbf{Model Performance Ranking}}                                           \\
                                                &                                              &                                            & \textbf{No.1}    & \textbf{No.2}     & \textbf{No.3}     & \textbf{No.4}     & \textbf{No.5}     \\\hline
\multirow{12}{*}{\textbf{\begin{tabular}[c]{@{}c@{}}Object\\ Detection\end{tabular}}}     & \textbf{Clean}       & DOTA                 & YOLOv5        & YOLOv3            & Swin Transformer        & Cascade R-CNN     & FoveaBox          \\\cline{2-8}
                                                & \multirow{4}{*}{\textbf{\begin{tabular}[c]{@{}c@{}}Adversarial\\ Noises\end{tabular}}} & Thys et al.  & Swin Transformer & YOLOv5   & Cascade R-CNN  & Mask R-CNN        & FreeAnchor        \\
                                                &                                              & APPA (on)                                  & Cascade R-CNN    & FreeAnchor        & YOLOv5            & Swin Transformer  & YOLOv3            \\
                                                &                                              & APPA (outside)                             & Faster R-CNN     & FreeAnchor        & Cascade R-CNN     & SSD               & RetinaNet         \\
                                                &                                              & CBA                                        & VFNet     & Faster R-CNN      & Mask R-CNN        & ATSS              & Swin Transformer  \\\cline{2-8}
                                                & \multirow{7}{*}{\textbf{\begin{tabular}[c]{@{}c@{}}Natural\\ Noises\end{tabular}}}    & Gaussian Noise       & YOLOv5     & YOLOv3      & VFNet      & FoveaBox   & TOOD              \\
                                                &                                              & Poisson Noise                              & YOLOv5           & YOLOv3            & Swin Transformer  & Cascade R-CNN     & FoveaBox          \\
                                                &                                              & SP Noise                                   & YOLOv5           & YOLOv3            & Swin Transformer  & SSD               & VFNet      \\
                                                &                                              & Random Noise                               & YOLOv3           & YOLOv5            & Swin Transformer  & VFNet      & ATSS              \\
                                                &                                              & Fog                                        & YOLOv5           & YOLOv3            & Swin Transformer  & SSD               & FoveaBox          \\
                                                &                                              & Rain                                       & YOLOv5           & YOLOv3            & Swin Transformer  & Cascade R-CNN     & ATSS              \\
                                                &                                              & Snow                                       & YOLOv5           & YOLOv3            & Swin Transformer  & Cascade R-CNN     & FoveaBox          \\\hline
\multirow{15}{*}{\textbf{\begin{tabular}[c]{@{}c@{}}Image\\ Classification\end{tabular}}} & \textbf{Clean}              & AID               & ResNet-152       & DenseNet-169      & Swin-T            & WRN-50-2          & Swin-S            \\\cline{2-8}
                                                & \multirow{7}{*}{\textbf{\begin{tabular}[c]{@{}c@{}}Adversarial\\ Noises\end{tabular}}} & AA   & \textbackslash{} & \textbackslash{}  & \textbackslash{}  & \textbackslash{}  & \textbackslash{}  \\
                                                &                                              & CW                                         & Vit-B/32         & Vit-B/16          & ResNeXt-50-32x4d  & ResNet-152        & WRN-101-2         \\
                                                &                                              & PGD-10                                     & ResNeXt-50-32x4d & ShuffleNetV2-x2.0 & DenseNet-121      & DenseNet-169      & MobileNetV3-L     \\
                                                &                                              & MIFGSM                                     & ResNeXt-50-32x4d & DenseNet-121      & ShuffleNetV2-x2.0 & DenseNet-169      & MobileNetV3-L     \\
                                                &                                              & FGSMs                                      & Vit-B/16         & Vit-B/32          & Swin-S            & WRN-101-2         & Swin-T            \\
                                                &                                              & FGSMm                                      & ResNeXt-50-32x4d & ResNeXt-101-32x8d & ResNet-152        & WRN-101-2         & WRN-50-2          \\
                                                &                                              & FGSMl                                      & ResNeXt-50-32x4d & ResNeXt-101-32x8d & ResNet-152        & WRN-101-2         & ResNet-50         \\\cline{2-8}
                                                & \multirow{7}{*}{\textbf{\begin{tabular}[c]{@{}c@{}}Natural\\ Noises\end{tabular}}}    & Gaussian Noise    & Vit-B/16    & ResNet-152    & Swin-T     & ResNeXt-50-32x4d  & ResNeXt-101-32x8d \\
                                                &                                              & Poisson Noise                              & ResNet-152       & Swin-T            & ResNeXt-101-32x8d & MobileNetV3-L     & Swin-S            \\
                                                &                                              & SP Noise                                   & Swin-T           & ResNeXt-50-32x4d  & ResNet-152        & Vit-B/16          & WRN-101-2         \\
                                                &                                              & Random Noise                               & ResNeXt-50-32x4d & Vit-B/32          & ResNeXt-101-32x8d & Vit-B/16          & WRN-101-2         \\
                                                &                                              & Fog                                        & Swin-T           & ResNet-101        & WRN-101-2         & ResNeXt-101-32x8d & MobileNetV3-L     \\
                                                &                                              & Rain                                       & Swin-T           & Swin-S            & ResNeXt-101-32x8d & ResNet-101        & ResNeXt-50-32x4d  \\
                                                &                                              & Snow                                       & Swin-T           & Swin-S            & ResNeXt-101-32x8d & MobileNetV3-L     & ResNeXt-50-32x4d 
\\                   
\hline\hline
\end{tabular*}
    \begin{tablenotes}
        \footnotesize      
        \item " \textbackslash{}" represents accuracy lower than 0.05$\%$.
        \item "on" and "outside" represent patches on and outside targets, respectively.
        \item Adversarial robustness is evaluated under white-box conditions.
    \end{tablenotes}
\end{threeparttable}
\end{table*}

The rest parts of this manuscript are organized as follows.
We first thoroughly survey the robustness of DNNs in CV and RS in Section \ref{Section2};
Secondly, the robustness of image classification and object detection is comprehensively benchmarked in the RS field in Section \ref{Section3};
Then, rigorous and extensive experimental results are provided in Section \ref{Section4};
Followed by some further discussions as presented in Section \ref{Section5};
Finally, we summarize this work in Section \ref{Section6}.

\section{Survey}
\label{Section2}

The integration of DNNs into safety-critical applications, such as autonomous driving \cite{yu2022dair,zhou2023bridging}, face recognition \cite{kim2022adaface,liu2023brain}, RS \cite{mei2023lightweight,mei2023rotation}, \etc., highlights the criticality of enhancing model robustness and developing resilient DL systems. 
As a result, there is a growing need to comprehensively evaluate the robustness of DL models for a better understanding of the factors affecting their resilience and facilitate further improvements in DNNs' robustness.
In this section, we first introduce the background knowledge of the adversarial attack.
Then, the robustness of DNNs-based methods is comprehensively surveyed in CV and RS, respectively.

\begin{figure*}[!t]
  \centering
  \includegraphics[width=0.95\linewidth]{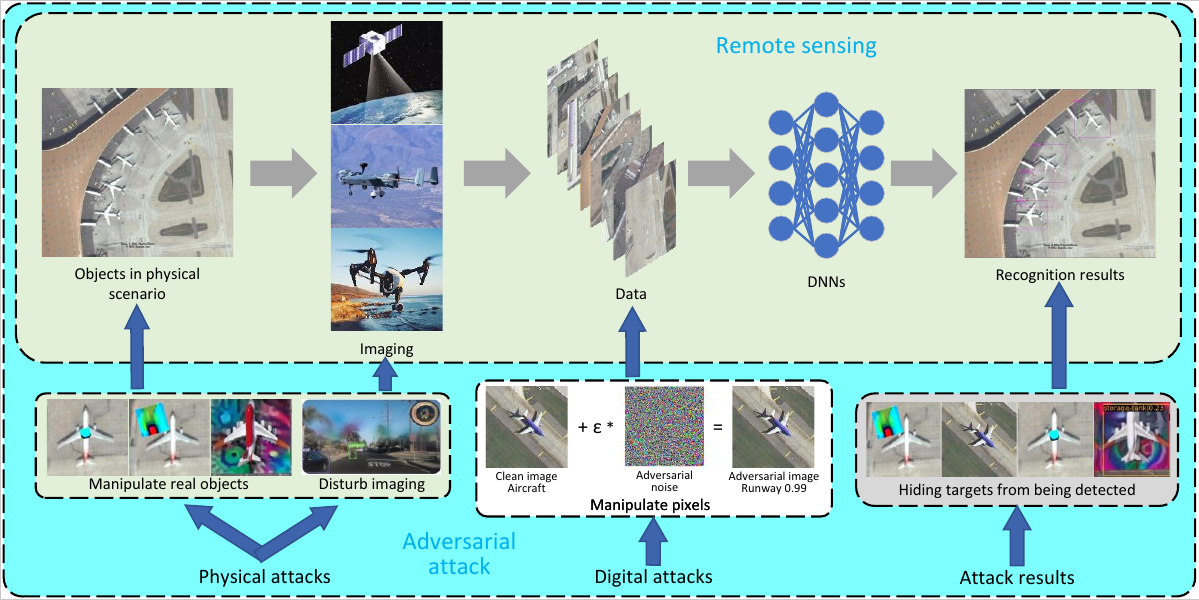}
  \caption{Comparison of digital attack and physical attack.}
  \label{fig:digital_physical_comparison}
\end{figure*}

\subsection{Background Knowledge}

The primary objective of DL is to enable models to learn from data in a manner that allows them to perform tasks similar to humans when confronted with new data.
Over the last decade, DL has made tremendous strides in numerous significant applications.
Although DL has delivered impressive results in practical applications, recent years have revealed a disturbing phenomenon where DL models may make abnormal predictions that are inconsistent with human intuition. 
For instance, a model could yield significantly different predictions on two visually similar images, with one being perturbed by malicious and imperceptible noises \cite{szegedy2014intriguing,ian2015explaining}, whereas a human's prediction would remain unaffected by such noises. 
We refer to this phenomenon as the adversarial phenomenon or adversarial attack, signifying the inherent adversarial relationship between DL models and human perception \cite{wu2023adversarial}.

The discovery of the adversarial phenomenon originated from image classification tasks in the digital realm.
As a consequence, the majority of existing research on adversarial attacks has been concentrated on image classification tasks in the digital domain \cite{szegedy2014intriguing,ian2015explaining,liu2022towards,shi2022query,ma2021simulating,mahmood2021robustness,ilyas2019adversarial,cheng2021perturbation}, \ie the so-called \textbf{digital attack}.
In comparison, \textbf{physical attack} happens in real physical world scenarios.
Consequently, in this section, we provide an overview of adversarial attacks, offering background knowledge on both digital attacks and physical attacks as illustrative examples. 
Digital attacks involve manipulating image pixel values in the digital domain after capturing an image using an imaging device. 
On the other hand, physical attacks involve tampering with the target to be disturbed before image capture. 
Although digital attack methods can easily fool various DL models in the digital domain, the generated digital perturbations lose their effectiveness in the real physical world because they are often imperceptible and cover the entire image, making them invisible to imaging devices. 
As a result, researchers are increasingly studying adversarial attacks that are applicable in the physical world. 
Physical attack methods have been proposed and used to attack intelligent systems such as autonomous driving \cite{thys2019fooling,wang2021dual,xiao2021improving,cheng2022physical}, face recognition\cite{pautov2019adversarial,sharif2016accessorize,wei2022simultaneously,wei2022adversarial}, RS \cite{lian2022benchmarking,lian2023contextual,lian2023cba,du2022physical}, security monitoring \cite{wu2020making,wang2019advpattern,ding2021towards}, \etc.

Typically, digital and physical attacks occur at different stages of an intelligent recognition task, as shown in Fig. \ref{fig:digital_physical_comparison}, which illustrates the difference between digital and physical attacks in the context of RS. 
It is observed that:

\begin{itemize}
    \item For physical attacks, the attacker manipulates either the actual targets or the imaging process itself to intentionally induce incorrect predictions;
    \item For digital attacks, the attacker directly modifies the pixel values of the image data captured by the imaging device to implement the attack.
\end{itemize}

\begin{figure}[!t]
  \centering
  \includegraphics[width=0.99\linewidth]{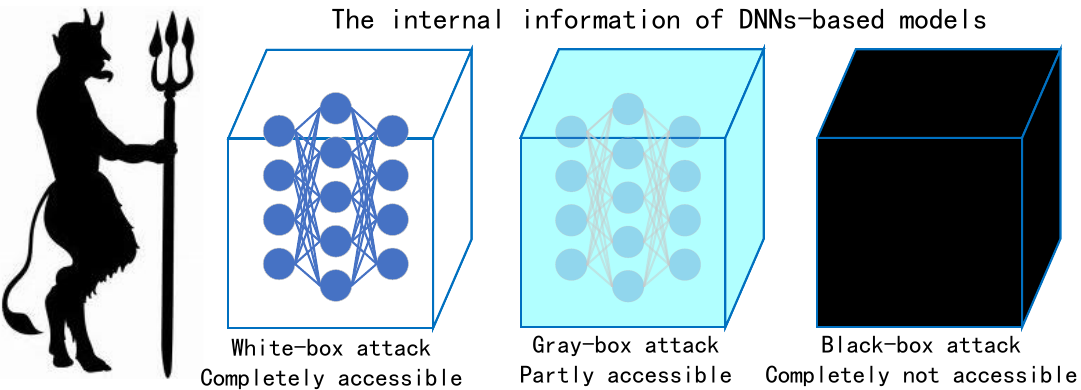}
  \caption{Categorization of adversarial attacks according to attackers' accessible information.}
  \label{fig:white_gray_black}
\end{figure}

In addition, adversarial attacks can be classified based on other attack characteristics.
Regarding the attacker's access to the victim model's information, adversarial attacks can be categorized into three types: \textbf{white-box attack}, \textbf{gray-box attack}, and \textbf{black-box attack}, as shown in Fig. \ref{fig:white_gray_black}. 
In white-box attacks, the attacker has full access to the internal information of the model, including its structure, parameters, gradients, and other relevant details. 
This comprehensive knowledge enables the attacker to craft sophisticated adversarial examples to deceive the model.
Gray-box attacks grant the attacker partial access to the internal information of the model. 
Although not as extensive as in white-box attacks, this limited access still provides valuable insights that can be leveraged for crafting effective adversarial examples.
In contrast, black-box attacks present unique challenges as the attacker lacks access to the specific parameters and structural details of the target model. 
Consequently, alternative techniques must be employed to generate adversarial examples in such scenarios. 
Transfer-based attacks, where knowledge gained from a substitute model is utilized to craft adversarial examples for the black-box model, are commonly employed. 
Additionally, gradient estimation methods based on query results can also be utilized to approximate the gradients of the black-box model and guide the generation of effective adversarial examples. 
These approaches showcase the ingenuity and adaptability of attackers in navigating the constraints imposed by black-box settings.
While several white-box attack methods \cite{zhang2022investigating,lian2023cba,wei2023efficient,chen2022adversarial} have been proposed, they typically demand extensive information about the victim model, rendering their practical applicability in real-world attack and defense scenarios quite challenging.
As a result, researchers in the field of adversarial machine learning have increasingly directed their attention towards black-box attack methods \cite{tu2019autozoom,wei2022sparse,shi2022query,wei2022simultaneously}, which are more suitable for real-world adversarial situations where the attacker only has limited knowledge of the target model.

\begin{figure*}[!t]
  \centering
  \includegraphics[width=0.9\linewidth]{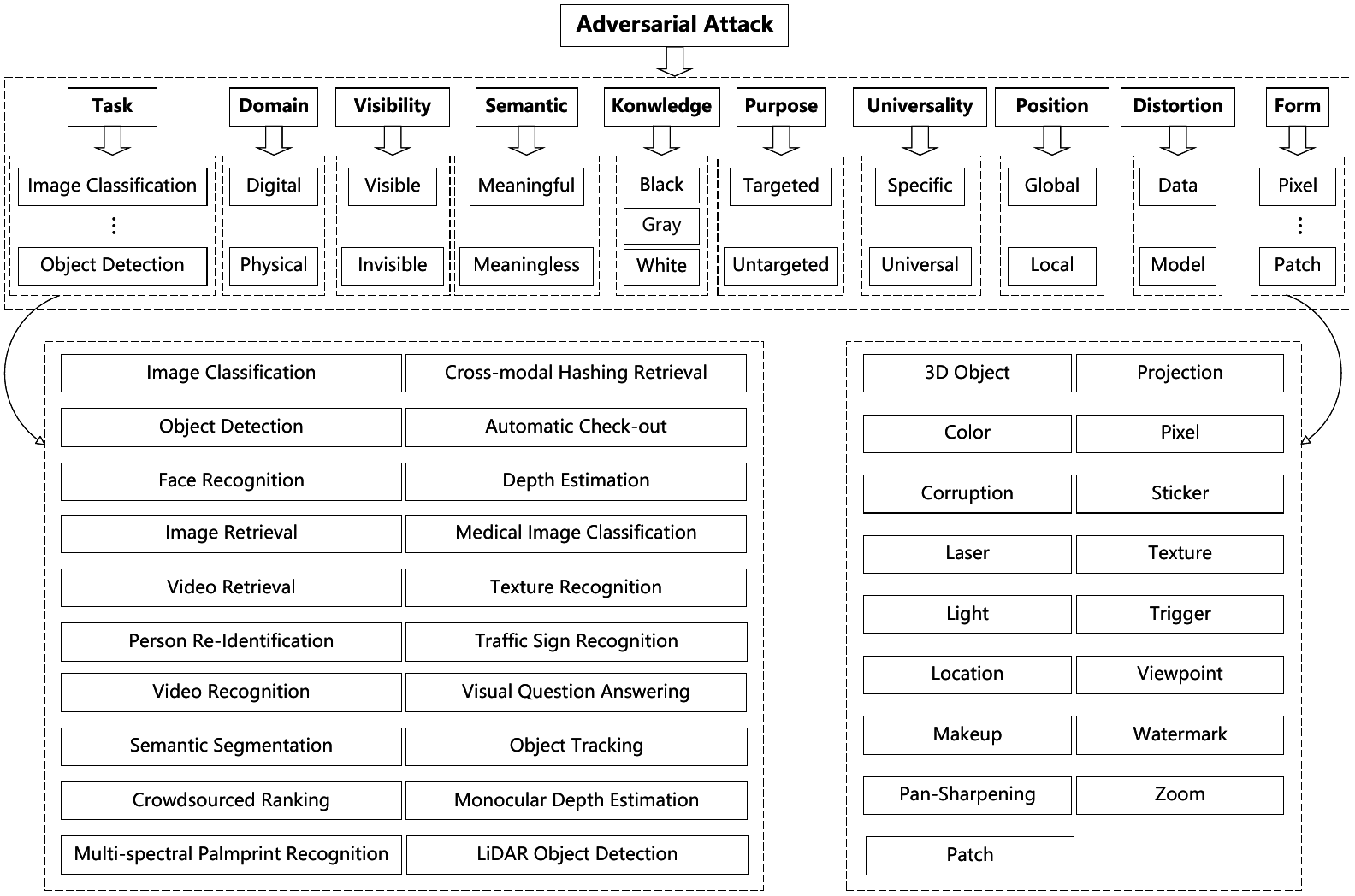}
  \caption{Taxonomy of adversarial attacks.}
  \label{fig:taxonomy}
\end{figure*}

\begin{figure*}
  \centering
  \begin{subfigure}{0.105\linewidth}
    \includegraphics[width=1\linewidth]{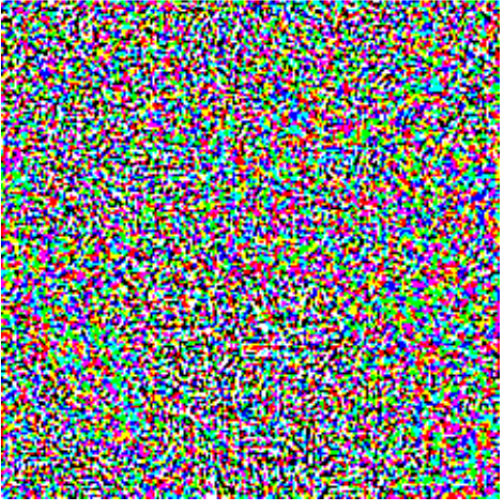}
    \captionsetup{font=scriptsize}
    \caption{Pixel \cite{ian2015explaining}}
  \end{subfigure}
  \begin{subfigure}{0.105\linewidth}
    \includegraphics[width=1\linewidth]{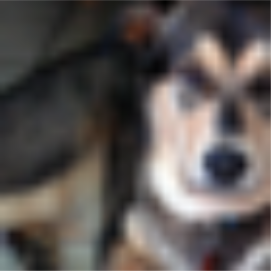}
    \captionsetup{font=scriptsize}
    \caption{Watermark \cite{li2020open}}
  \end{subfigure}
  \begin{subfigure}{0.105\linewidth}
    \includegraphics[width=1\linewidth]{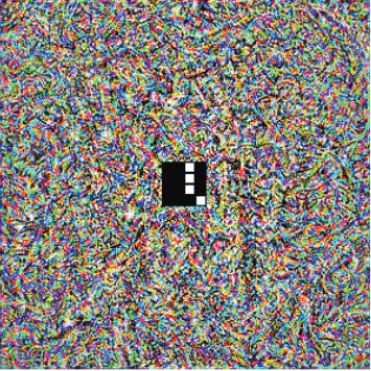}
    \captionsetup{font=scriptsize}
    \caption{Trigger \cite{elsayed2018adversarial}}
  \end{subfigure}
  \begin{subfigure}{0.105\linewidth}
    \includegraphics[width=1\linewidth]{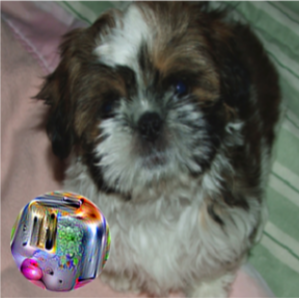}
    \captionsetup{font=scriptsize}
    \caption{Patch \cite{brown2017adversarial}}
  \end{subfigure}
  \begin{subfigure}{0.105\linewidth}
    \includegraphics[width=1\linewidth]{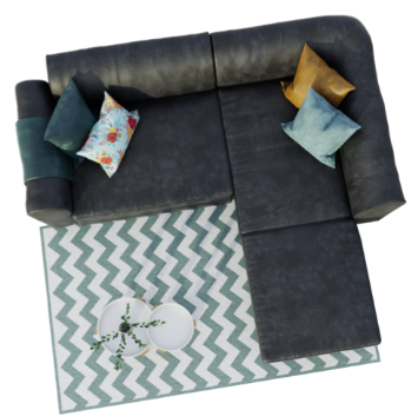}
    \captionsetup{font=scriptsize}
    \caption{Viewpoint \cite{dongviewfool}}
  \end{subfigure}
  \begin{subfigure}{0.105\linewidth}
    \includegraphics[width=1\linewidth]{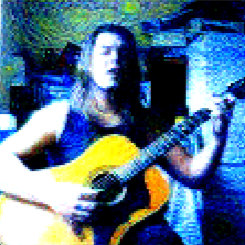}
    \captionsetup{font=scriptsize}
    \caption{Style \cite{cao2022stylefool}}
  \end{subfigure}
  \begin{subfigure}{0.105\linewidth}
    \includegraphics[width=1\linewidth]{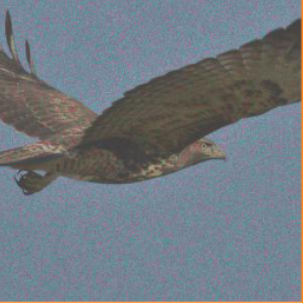}
    \captionsetup{font=scriptsize}
    \caption{Erosion \cite{huang2023erosion}}
  \end{subfigure}
  \begin{subfigure}{0.105\linewidth}
    \includegraphics[width=1\linewidth]{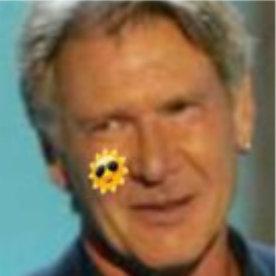}
    \captionsetup{font=scriptsize}
    \caption{Sticker \cite{wei2022adversarial}}
  \end{subfigure}
    \begin{subfigure}{0.105\linewidth}
    \includegraphics[width=1\linewidth]{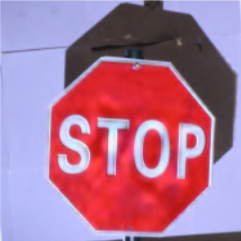}
    \captionsetup{font=scriptsize}
    \caption{Light \cite{gnanasambandam2021optical}}
  \end{subfigure}
  \begin{subfigure}{0.105\linewidth}
    \includegraphics[width=1\linewidth]{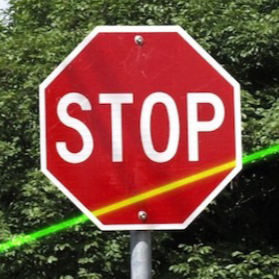}
    \captionsetup{font=scriptsize}
    \caption{Laser \cite{duan2021adversarial}}
  \end{subfigure}
  \begin{subfigure}{0.105\linewidth}
    \includegraphics[width=1\linewidth]{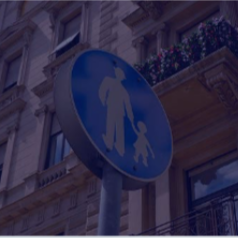}
    \captionsetup{font=scriptsize}
    \caption{Color \cite{hu2022adversarialColor}}
  \end{subfigure}
  \begin{subfigure}{0.105\linewidth}
    \includegraphics[width=1\linewidth]{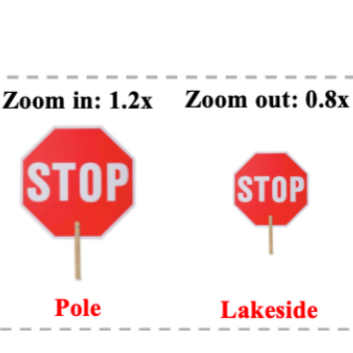}
    \captionsetup{font=scriptsize}
    \caption{Zoom \cite{hu2022adversarialZoom}}
  \end{subfigure}
  \begin{subfigure}{0.105\linewidth}
    \includegraphics[width=1\linewidth]{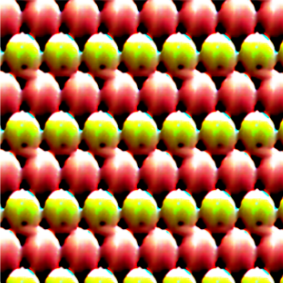}
    \captionsetup{font=scriptsize}
    \caption{Texture \cite{hu2022adversarialfooling}}
  \end{subfigure}
  \begin{subfigure}{0.105\linewidth}
    \includegraphics[width=1\linewidth]{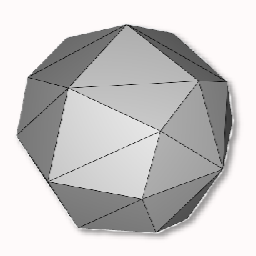}
    \captionsetup{font=scriptsize}
    \caption{3D object \cite{liu2023x}}
  \end{subfigure}
  \begin{subfigure}{0.105\linewidth}
    \includegraphics[width=1\linewidth]{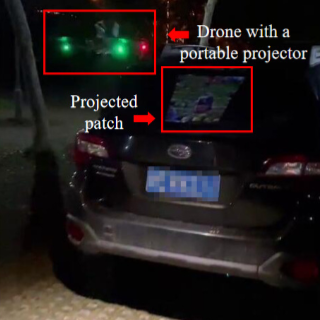}
    \captionsetup{font=scriptsize}
    \caption{Projection \cite{wen2023light}}
  \end{subfigure}
  \begin{subfigure}{0.105\linewidth}
    \includegraphics[width=1\linewidth]{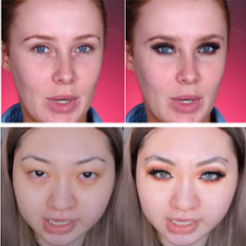}
    \captionsetup{font=scriptsize}
    \caption{Makeup \cite{zhu2019generating}}
  \end{subfigure}
  \begin{subfigure}{0.105\linewidth}
    \includegraphics[width=1\linewidth]{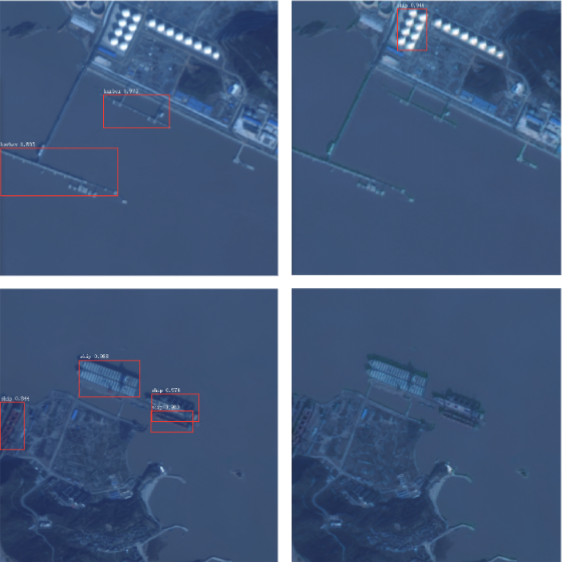}
    \captionsetup{font=scriptsize}
    \caption{PS \cite{yuan2021generating}}
  \end{subfigure}
  \begin{subfigure}{0.105\linewidth}
    \includegraphics[width=1\linewidth]{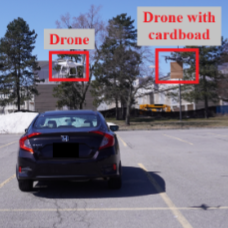}
    \captionsetup{font=scriptsize}
    \caption{Location \cite{zhu2021can}}
  \end{subfigure}
  \caption{Adversarial perturbations in different forms. Note that PS represents Pan-Sharpening.}
  \label{fig:perturbation_forms}
\end{figure*}

We have categorized adversarial attacks based on their distinct characteristics and strategies employed as shown in Fig. \ref{fig:taxonomy}.
Furthermore, we also display different forms of perturbations, as shown in Fig. \ref{fig:perturbation_forms}.
In this section, we illustrate the adversarial attack from different domains, \ie digital attacks and physical attacks.

\subsubsection{Formulation of digital attacks} 

Assuming the presence of an image classifier $ f(\boldsymbol{x}): \boldsymbol{x} \in \boldsymbol{X} \to y \in \boldsymbol{Y} $ that generates a prediction $y$ based on an input image $\boldsymbol{x}$, the primary aim of an adversarial attack is to generate an adversarial example $\boldsymbol{x}^{*}$ that closely resembles the clean example $\boldsymbol{x}$ but causes the image classifier $f(\boldsymbol{x})$ to make an incorrect prediction $y^*$.
From a technical standpoint, adversarial attack methods can be categorized as either \textbf{non-targeted} or \textbf{targeted}, depending on the attacker's motives.

\begin{figure}[!t]
  \centering
  \includegraphics[width=0.8\linewidth]{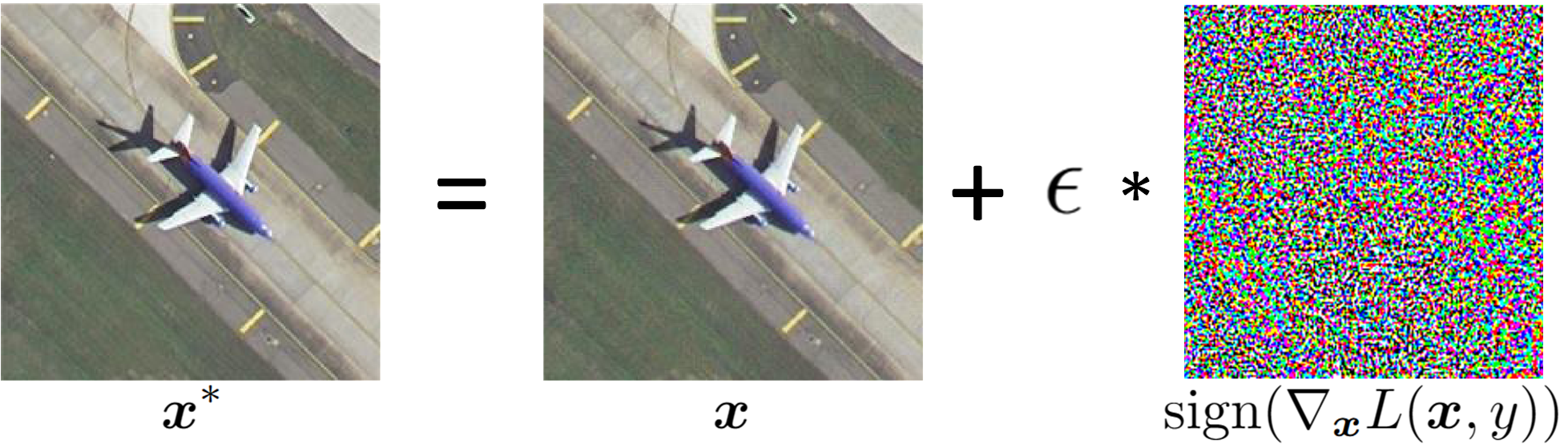}
  \caption{Formulation of digital attack with imperceptible perturbation.}
  \label{fig:formulation_digital_attack}
\end{figure}

Suppose an input image $\boldsymbol{x}$ is properly classified by a model such that its predicted label is $y$, \ie $f(\boldsymbol{x})=y$.
\textbf{Non-targeted attack} methods are designed to generate adversarial examples $\boldsymbol{x} ^ {*} $ by adding imperceptible perturbations to clean images $\boldsymbol{x}$, which mislead the classifier into making an incorrect prediction, \ie $f(\boldsymbol{x} ^ {*}) \neq y$. 
\textbf{Targeted attack} methods are designed to manipulate the classifier into predicting a specific label, such that $f(\boldsymbol{x}^{*}) = y^*$, where $y^*$ represents the target label specified by the attacker and $y^* \neq y$. 
These methods are intended to deceive the classifier into producing a specific output rather than simply causing a misclassification.
The $L_p$ norm is typically used as a measure of the visibility of the adversarial noise. 
In the case of digital attacks, as shown in Fig. \ref{fig:formulation_digital_attack}, the adversarial noise is required to be imperceptible to human vision, \ie less than or equal to a certain threshold value $\epsilon$, expressed as ${\Vert \boldsymbol{x}^* - \boldsymbol{x} \Vert}_p \le \epsilon$.

Current adversarial attack methods can be classified into two categories (adversarial optimization and distance optimization) according to the optimizing strategy adopted to generate the adversarial samples. 
In this article, we present the formulation of non-targeted attack methods, and the targeted version can be derived using a similar approach.

\textbf{Adversarial optimization attacks.}
Adversarial optimization attacks aim to elaborate adversarial examples $\boldsymbol{x}^*$ by maximizing a loss function $L(\boldsymbol{x}^*, y)$ with adversarial example $\boldsymbol{x}^*$ are within a certain distance $\epsilon$ of clean example $\boldsymbol{x}$, which is defined as:
\begin{equation}
    \label{eq_gradient_based}
    \mathop{\arg\max}\limits_{\boldsymbol{x}^*} L(\boldsymbol{x}^*, y) \quad s.t. \quad {\Vert \boldsymbol{x}^* - \boldsymbol{x} \Vert}_p \le \epsilon
\end{equation}
For example, fast gradient sign method (FGSM) \cite{ian2015explaining} crafts adversarial examples to subject to the $L_p$ norm constraint $ {\Vert \boldsymbol{x}^* - \boldsymbol{x} \Vert}_p \le \epsilon $, which is mathematically represented as:
\begin{equation}
    \label{eq_one-step1}
    \boldsymbol{x}^* = \boldsymbol{x} + \epsilon \cdot {\rm sign} (\nabla_{\boldsymbol{x}}L(\boldsymbol{x},y)),
\end{equation}
where $\nabla_{\boldsymbol{x}}L(\boldsymbol{x},y)$ is the gradient value of the objective loss $L(\boldsymbol{x},y)$ \wrt clean image $\boldsymbol{x}$, and $\rm sign()$ represents the sign function. 
A extension of FGSM algorithm is to satisfy the $L_2$ norm limitation $ {\Vert \boldsymbol{x}^* - \boldsymbol{x} \Vert}_2 \le \epsilon $ mathematically defined as:
\begin{equation}
    \label{eq_one-step2}
    \boldsymbol{x}^* = \boldsymbol{x} + \epsilon \cdot \frac{\nabla_{\boldsymbol{x}}L(\boldsymbol{x},y)}{{\Vert         \nabla_{\boldsymbol{x}}L(\boldsymbol{x},y) \Vert}_2}.
\end{equation}

The aforementioned adversarial optimization attacks are one-step methods.
Subsequently, multi-step methods, such as iterative FGSM (I-FGSM) \cite{kurakin2018adversarial}, momentum iterative FGSM (MI-FGSM) \cite{dong2018boosting}, projected gradient descent (PGD) \cite{madry2018towards}, Nesterov accelerated gradient (NI-FGSM) \cite{linnesterov}, AutoAttack (AA) \cite{croce2020reliable}, \etc., iteratively adopt one-step approaches multiple times with a small step size $\alpha$, which can be expressed as:
\begin{equation}
    \label{eq_iterative}
    \boldsymbol{x}^*_{t+1} = \boldsymbol{x}^*_{t} + \alpha \cdot {\rm sign} (\nabla_{\boldsymbol{x}}L(\boldsymbol{x}^*_{t},y)), \quad \boldsymbol{x}^*_0 = \boldsymbol{x}.
\end{equation}
To ensure that the adversarial perturbations generated are imperceptible to human observers, \ie satisfy the $L_p$ constraint, which can be achieved by simply clipping $\boldsymbol{x}^*_t$ into the $\epsilon$ vicinity of $\boldsymbol{x}$ or simply set $\alpha = \epsilon / T $ with $T$ being the number of iterations.

\textbf{Distance optimization attacks.} 
Distance optimization attacks, such as L-BFGS \cite{szegedy2014intriguing}, Deepfool \cite{moosavi2016deepfool}, C\&W \cite{carlini2017towards}, \etc., directly minimize the distance between the clean and adversarial examples, while ensuring that the adversarial examples are misclassified by the model, which can be mathematically expressed as:
\begin{equation}
    \label{eq_optimization-based}
    \mathop{\arg\min}\limits_{\boldsymbol{x}^*} \cdot {\Vert \boldsymbol{x}^* - \boldsymbol{x} \Vert}_p \quad s.t. \quad f(\boldsymbol{x}^*) \neq f(\boldsymbol{x}).
\end{equation}
For distance optimization attacks, they directly optimize the distance between an adversarial example and the corresponding benign example, thus the optimization of the $L_p$ norm is not necessary to be less than or equal to a particular threshold value.

\begin{figure}
  \centering
  \begin{subfigure}{0.49\linewidth}
    \includegraphics[width=1\linewidth]{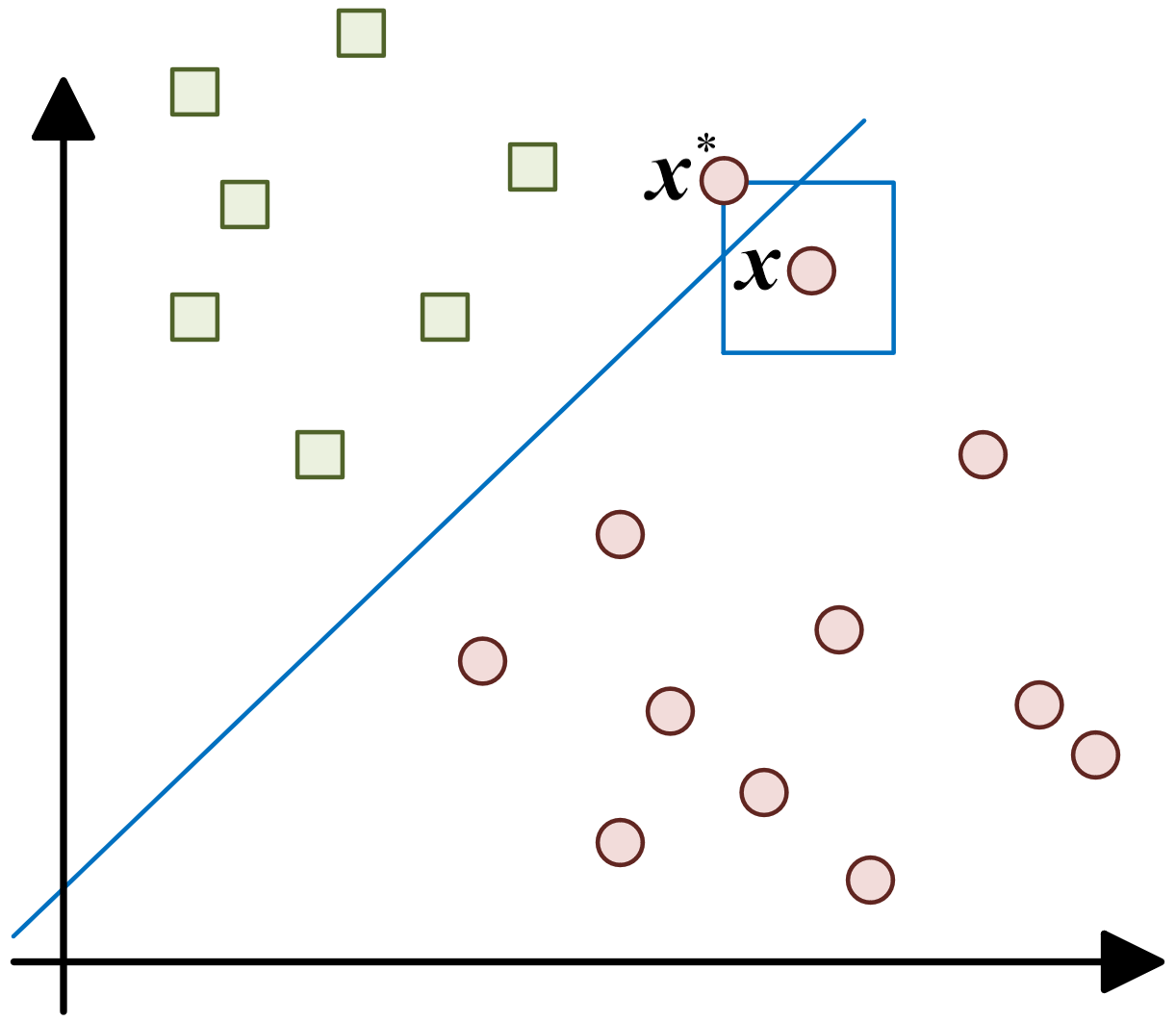}
    \caption{Adversarial optimization}
  \end{subfigure}
  \begin{subfigure}{0.49\linewidth}
    \includegraphics[width=1\linewidth]{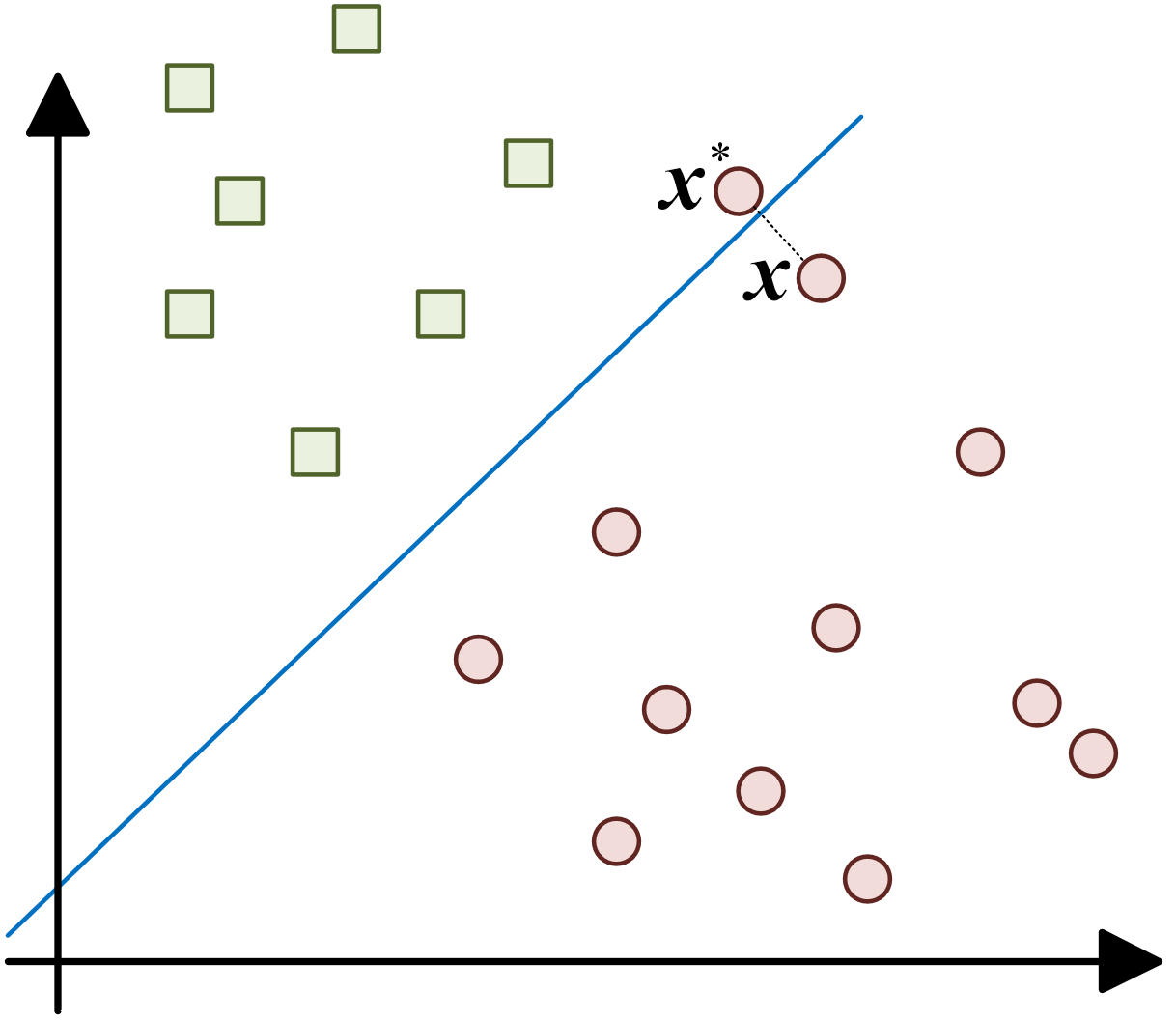}
    \caption{Distance optimization}
  \end{subfigure}  
  \caption{Comparison of adversarial optimization attack and distance optimization attack.}
  \label{fig:gradient&optimization_based_attack}
\end{figure}

To summarize, as shown in Fig. \ref{fig:gradient&optimization_based_attack}, adversarial optimization attacks aim to generate adversarial perturbations that are farthest from the decision boundary within the specified perturbation range. 
On the other hand, distance optimization attacks aim to minimize the size of the adversarial perturbation, \ie the distance between the adversarial and clean samples, for a given adversarial perturbation.
As a consequence, the adversarial perturbations generated by adversarial optimization attack methods are more effective in producing misclassifications, while the perturbations generated by distance optimization attack methods are more visually imperceptible.

\subsubsection{Formulation of physical attacks} 

\begin{figure}[!t]
  \centering
  \includegraphics[width=0.99\linewidth]{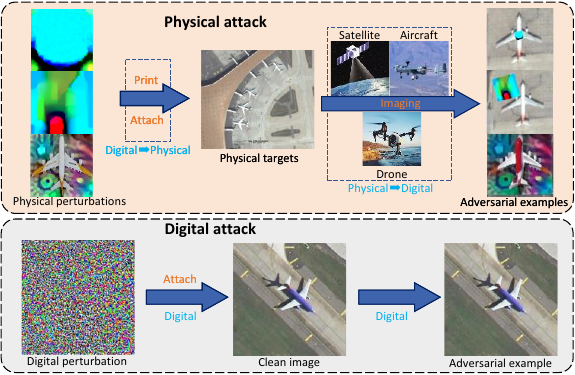}
  \caption{Cross-domain transformation. For physical attacks, the elaborated physical perturbations have to suffer cross-domain transformations, \ie digital to physical and physical to digital, to perform attacks in the physical world, while it is unnecessary for digital attacks.}
  \label{fig:across_domain_tranformation}
\end{figure}

As the study of adversarial attack problems has progressed, researchers have found that generating adversarial examples in the digital domain presents considerable difficulties in launching successful attacks in the physical domain.
Kurakin \etal \cite{kurakin2018adversarial} first discover that DNNs are also susceptible and vulnerable to adversarial attacks performed in real-world physical scenarios, \ie physical attacks.
Notably, physical attacks carried out in real-world settings are significantly more dangerous than digital ones. 
Consequently, the practical feasibility of adversarial attack methods in physical contexts has emerged as a crucial area of research in the domain of machine learning security.
However, physical attacks still face some challenges when compared to digital attacks, such as:

\begin{itemize}
    \item Physical attack methods should be able to withstand the impact of the imaging process, which mainly includes optical lenses, image sensors, processors, \etc;
    \item Adversarial perturbations created using physical attack methods should be robust enough to handle the impact of dynamic environments when they face transformations across different domains, as shown in Fig. \ref{fig:across_domain_tranformation};
    \item Adversarial perturbations for physical attacks should be as concealed as possible to avoid attention-grabbing anomalies.
    In contrast, digital attacks typically involve limited pixel-level modifications to images, which is hard to notice while it is challenging to make physical attacks unobtrusive.
\end{itemize}

\begin{figure}
  \centering
  \begin{subfigure}{0.24\linewidth}
    \includegraphics[width=1\linewidth]{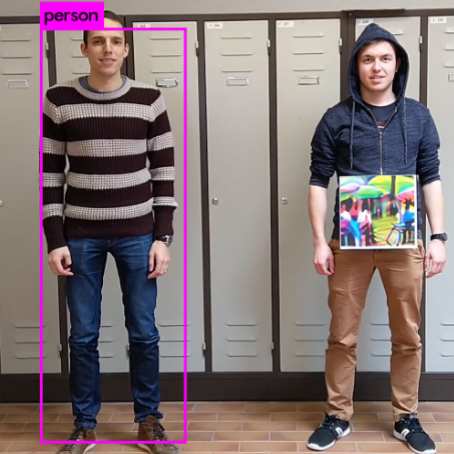}
    \captionsetup{font=scriptsize}
    \caption{Normal \cite{thys2019fooling}}
  \end{subfigure}
  \begin{subfigure}{0.24\linewidth}
    \includegraphics[width=1\linewidth]{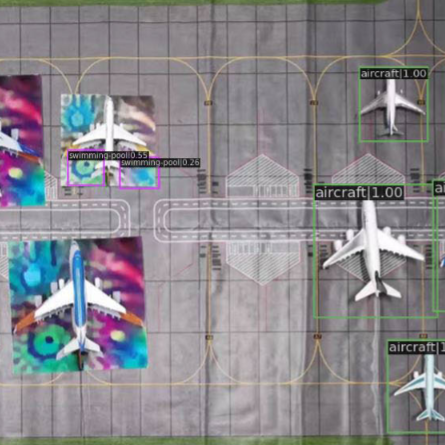}
    \captionsetup{font=scriptsize}
    \caption{Background \cite{lian2023cba}}
  \end{subfigure}
  \begin{subfigure}{0.24\linewidth}
    \includegraphics[width=1\linewidth]{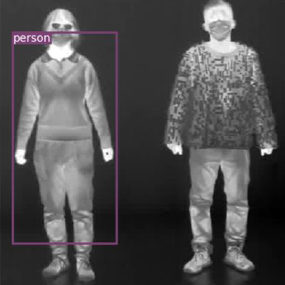}
    \captionsetup{font=scriptsize}
    \caption{Infrared \cite{zhu2022infrared}}
  \end{subfigure}
  \begin{subfigure}{0.24\linewidth}
    \includegraphics[width=1\linewidth]{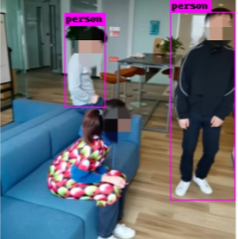}
    \captionsetup{font=scriptsize}
    \caption{Clothes \cite{hu2022adversarialfooling}}
  \end{subfigure}
  \begin{subfigure}{0.24\linewidth}
    \includegraphics[width=1\linewidth]{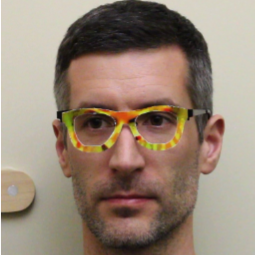}
    \captionsetup{font=scriptsize}
    \caption{Eyeglass \cite{sharif2016accessorize}}
  \end{subfigure}
  \begin{subfigure}{0.24\linewidth}
    \includegraphics[width=1\linewidth]{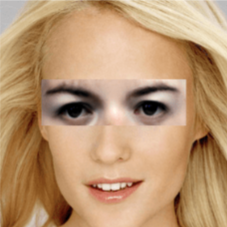}
    \captionsetup{font=scriptsize}
    \caption{Mask \cite{xiao2021improving}}
  \end{subfigure}
  \begin{subfigure}{0.24\linewidth}
    \includegraphics[width=1\linewidth]{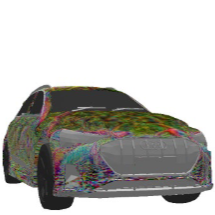}
    \captionsetup{font=scriptsize}
    \caption{3D \cite{wang2022fca}}
  \end{subfigure}
  \begin{subfigure}{0.24\linewidth}
    \includegraphics[width=1\linewidth]{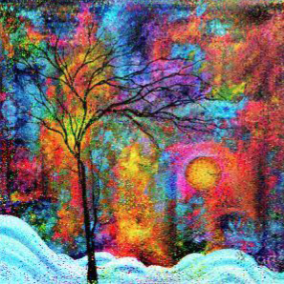}
    \captionsetup{font=scriptsize}
    \caption{Semantic \cite{guesmi2023advart}}
  \end{subfigure}
  \caption{Adversarial patches in different forms.}
  \label{fig:patch_forms}
\end{figure}

Consequently, numerous studies have aimed at assessing the physical adversarial robustness of DNNs in response to the concerns mentioned above during the past few years.
Physical attacks are executed in practical settings that encompass a diverse range of tasks conducted in physical scenarios.
Prior to executing a physical attack, it is imperative to fabricate the adversarial example properly. 
Attackers frequently prioritize the practicality of a given approach within a real-world setting, taking into account factors such as environmental interference, ease of manufacture, and avoidance of visual detection by human observers.
In this paper, we formulate physical attacks in patch form due to the widespread popularity of adversarial patches as an approach for implementing physical attacks in real-world scenarios.
we exhibit different forms of adversarial patches in Fig. \ref{fig:patch_forms}.

\begin{figure}[!t]
  \centering
  \includegraphics[width=0.99\linewidth]{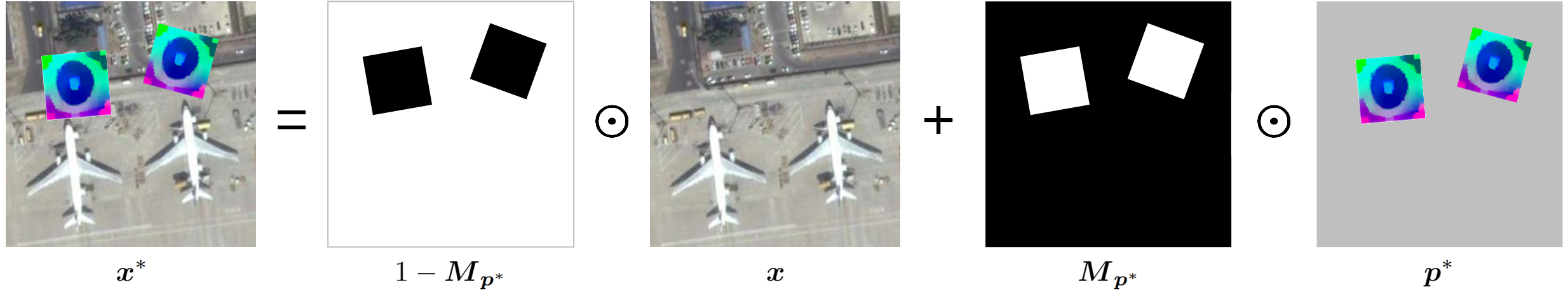}
  \caption{Formulation of patch-based physical attack.}
  \label{fig:formulation_physical_attack}
\end{figure}

In the context of digital adversarial attacks, global perturbations engendered throughout an entire image present substantial impediments to the practical execution of such assaults within real-world environments. 
In contrast, adversarial patches, which solely manipulate localized pixel regions, offer a more viable alternative. 
These patches can be conveniently produced via printing methods and directly adhered to the designated targets. 
A mask is commonly utilized to regulate the geometry of the disrupted area. 
Upon completing the optimization process for the adversarial patch within the digital domain, the tailored patch is subsequently crafted and strategically situated on the object's exterior surface or background area, as shown in Fig. \ref{fig:formulation_physical_attack}.
Mathematically, the adversarial example with adversarial patches can be formulated as:
\begin{equation}
    \label{eq_formulation_physical_attack}
    \boldsymbol{x}^* = (\boldsymbol{1}-\boldsymbol{M}_{\boldsymbol{p}^*}) \odot \boldsymbol{x} + \boldsymbol{M}_{\boldsymbol{p}^*} \odot \boldsymbol{p}^*,
\end{equation}
where $\odot$ and $\boldsymbol{p}^*$ denote Hadamard product and adversarial patches, respectively. 
Mask matrix $\boldsymbol{M}_{\boldsymbol{p}^*}$ is used to constrict the size, shape, and location of the adversarial patch, where the value of the patch position area is 1. $\boldsymbol{1}$ is a unit matrix with the same size as $\boldsymbol{M}_{\boldsymbol{p}^*}$.

\begin{figure}
  \centering
  \begin{subfigure}{0.188\linewidth}
    \includegraphics[width=1\linewidth]{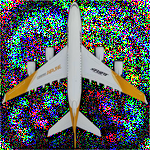}
    \caption{$\alpha_1$}
  \end{subfigure}
  \begin{subfigure}{0.188\linewidth}
    \includegraphics[width=1\linewidth]{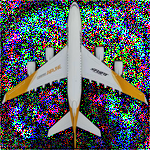}
    \caption{$\alpha_2$}
  \end{subfigure}
  \begin{subfigure}{0.188\linewidth}
    \includegraphics[width=1\linewidth]{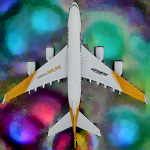}
    \caption{$\alpha_3$}
  \end{subfigure}
  \begin{subfigure}{0.188\linewidth}
    \includegraphics[width=1\linewidth]{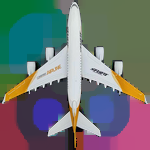}
    \caption{$\alpha_4$}
  \end{subfigure}
  \begin{subfigure}{0.188\linewidth}
    \includegraphics[width=1\linewidth]{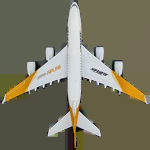}
    \caption{$\alpha_5$}
  \end{subfigure}
  \caption{Adversarial perturbations crafted with different weights of TV ($\alpha_5$ \textgreater{} $\alpha_4$ \textgreater{} $\alpha_3$ \textgreater{} $\alpha_2$ \textgreater{} $\alpha_1$) \cite{lian2023cba}.}
  \label{fig:tv}
\end{figure}

To address the challenge of capturing value discrepancies between neighboring pixels by image acquisition devices, Total Variation (TV), as delineated in \cite{sharif2016accessorize}, is usually incorporated into the objective function of physical attacks. 
The inclusion of $L_{tv}$ serves to ensure that the optimization process favors adversarial patches characterized by smooth patterns and gradual color transitions, as shown in Fig. \ref{fig:tv}.
TV can be mathematically defined as:
\begin{equation}
    \label{eq_Ltv}
    L_{tv} = \sum\limits_{i,j} \sqrt{(p_{i+1,j}-p_{i,j})^2 + (p_{i,j+1}-p_{i,j})^2},
\end{equation}
where $p_{i,j}$ denotes the pixel value situated at the $i$th row and $j$th column within the adversarial perturbations.

Owing to the color alterations that occur when transitioning the adversarial patch from the digital domain to the physical domain, the non-printability score (NPS) outlined in \cite{sharif2016accessorize} is frequently employed to evaluate the fidelity with which the colors in the adversarial patch can be reproduced in the physical world. 
This metric serves as an indicator of the distance between the digital representation of the adversarial patch and its physical manifestation when produced using a standard printer.
$L_{nps}$ is written as:
\begin{equation}
    \label{eq_Lnps}
    L_{nps} = \sum\limits_{i,j} \min_{c_{print} \in C} \mid p_{i,j} - c_{print} \mid,
\end{equation}
where $c_{print}$ represents an individual color within the set of physically printable colors, denoted as $C$. By incorporating $L_{nps}$ as part of the loss, the pixel values of the generated adversarial patch are biased towards printable colors from the set $C$, thereby promoting the reproducibility of the patch in the physical domain.

\begin{figure}[!t]
  \centering
  \includegraphics[width=0.8\linewidth]{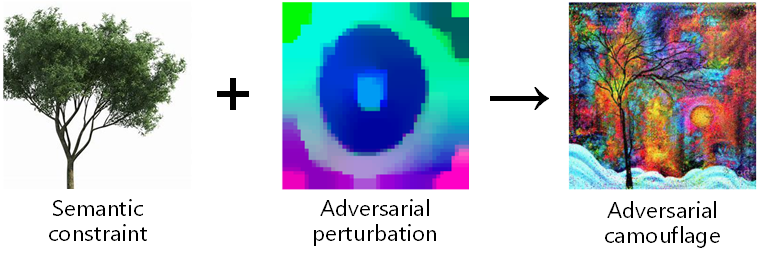}
  \caption{Adversarial camouflage.}
  \label{fig:adversarial_camouflage}
\end{figure}

Last but not least, camouflage loss $L_{cam}$ can be added to improve the invisibility of the adversarial patches to human visual perception.
From an academic standpoint, the rationale for employing camouflage loss stems from the observation that carefully crafted adversarial patches often exhibit vibrant hues and unconventional patterns. 
By incorporating camouflage loss into the optimization process, it becomes possible to generate adversarial patches that seamlessly blend with natural things, as shown in Fig. \ref{fig:adversarial_camouflage}, ensuring that the resultant perturbations remain inconspicuous while retaining their effectiveness in adversarial settings.
Technically, the $L_p$ norm is often used as the camouflage metric to measure the distance between adversarial patches and natural images.

In summary, the total objective function of physical attacks in patch form can be derived from the combination of the aforementioned parts and adversarial loss $L_{adv}$ (similar to digital attacks).
The total loss is depicted as:
\begin{equation}
    \label{eq_total_loss}
    L = L_{adv} + \alpha \cdot L_{tv} + \beta \cdot L_{nps} + \gamma \cdot L_{cam},
\end{equation}
where $\alpha$, $\beta$, and $\gamma$ are adopted to scale different components of the total loss.

\subsection{Survey of Robustness in CV}

In the following subsections, we provide a comprehensive examination of adversarial attacks as they pertain to the domain of CV, encompassing a variety of tasks including, but not limited to, image classification and object detection. 
By conducting an in-depth review of the pertinent literature, we aim to elucidate the underlying principles, methodologies, and implications of these attacks, thereby contributing to a more robust understanding of their role and significance within the broader context of CV research.

\subsubsection{Image Classification}
\label{Subsubsection2.2.1}

In the present section, we provide an overview of adversarial attacks in image classification, with a particular emphasis on both digital and physical attack methodologies.
The majority of research on adversarial attacks has focused on the digital domain, as the attacks were initially discovered in this context.

\begin{table*}[t!]
\scriptsize
    \renewcommand{\arraystretch}{1.25}
\caption{Digital attacks against image classification.}
\label{table:digital_classification}
\centering
\setlength{\tabcolsep}{0.35mm}
\begin{threeparttable}
\begin{tabular*}{\hsize}{ccccccccccccc}
\hline\hline
\thead{\textbf{Methods}} & {\textbf{Venue}} & {\textbf{Tasks}} & {\textbf{Domain}} & {\textbf{Visibility}} & {\textbf{Semantics}} & {\textbf{Knowledge}} & {\textbf{Purpose}} & {\textbf{Universality}} & {\textbf{Position}} & {\textbf{Distortion}} & {\textbf{Form}} 
\\ \hline
{EDA\cite{chen2018ead}}               & {AAAI 2018}            & {Image Classification} & {Digital}         & {Invisible}           & {Meaningless}        & {Black}              & {Targeted}         & {Specific}              & {Global}            & {Data}                & {Pixel}         \\
{AutoZOOM\cite{tu2019autozoom}}          & {AAAI 2019}            & {Image Classification} & {Digital}         & {Invisible}           & {Meaningless}        & {Black}              & {Both}             & {Specific}              & {Global}            & {Data}                & {Pixel}         \\
{Li et al.\cite{li2020open}}         & {NlPS 2020}            & {Image Classification} & {Digital}         & {Visible}             & {Meaningless}        & {White}              & {Targeted}         & {Universal}             & {Local}             & {Both}                & {Watermark}     \\
{TA-LBF\cite{bai2021targeted}}            & {ICLR 2021}            & {Image Classification} & {Digital}         & {/}                   & {/}                  & {White}              & {Targeted}         & {Specific}              & {/}                 & {Model}               & {/}             \\
{HPT\cite{bai2022hardly}}               & {ECCV 2022}            & {Image Classification} & {Digital}         & {Invisible}           & {Meaningless}        & {White}              & {Targeted}         & {Specific}              & {Global}            & {Both}                & {Pixel}         \\
{SSA/TSA\cite{bai2022versatile}}           & {arXiv 2022}           & {Image Classification} & {Digital}         & {Invisible}           & {Meaningless}        & {White}              & {Targeted}         & {Specific/Universal}    & {Local}             & {Both}                & {Trigger}       \\
{HMCAM\cite{wang2020hamiltonian}}             & {TPAMI 2020}           & {Image Classification} & {Digital}         & {Invisible}           & {Meaningless}        & {Both}               & {Both}             & {Specific}              & {Global}            & {Data}                & {Pixel}         \\
{BASES\cite{cai2022blackbox}}             & {NIPS 2022}            & {Image Classification} & {Digital}         & {Invisible}           & {Meaningless}        & {Black}              & {Both}             & {Specific}              & {Global}            & {Data}                & {Pixel}         \\
{$A^3$\cite{liu2022practical}}                & {CVPR 2022}            & {Image Classification} & {Digital}         & {Invisible}           & {Meaningless}        & {White}              & {Untargeted}       & {Specific}              & {Global}            & {Data}                & {Pixel}         \\
{L-BFGS\cite{szegedy2014intriguing}}            & {ICLR 2014}            & {Image Classification} & {Digital}         & {Invisible}           & {Meaningless}        & {White}              & {Targeted}         & {Specific}              & {Global}            & {Data}                & {Pixel}         \\
{FGSM\cite{ian2015explaining}}              & {ICLR 2015}            & {Image Classification} & {Digital}         & {Invisible}           & {Meaningless}        & {White}              & {Targeted}         & {Specific}              & {Global}            & {Data}                & {Pixel}         \\
{DeepFool\cite{moosavi2016deepfool}}          & {CVPR 2016}            & {Image Classification} & {Digital}         & {Invisible}           & {Meaningless}        & {White}              & {Untargeted}       & {Specific}              & {Global}            & {Data}                & {Pixel}         \\
{JSMA\cite{papernot2016limitations}}              & {EuroS\&P 2016}        & {Image Classification} & {Digital}         & {Invisible}           & {Meaningless}        & {White}              & {Targeted}         & {Specific}              & {Global}            & {Data}                & {Pixel}         \\
{UPSET/ANGRI\cite{sarkar2017upset}}       & {arXiv 2017}           & {Image Classification} & {Digital}         & {Visible}             & {Meaningless}        & {Black}              & {Targeted}         & {Universal}             & {Global}            & {Data}                & {Pixel}         \\
{Papernot et al.\cite{papernot2017practical}}   & {ACM ASIACCS 2017}     & {Image Classification} & {Digital}         & {Invisible}           & {Meaningless}        & {Black}              & {Untargeted}       & {Specific}              & {Global}            & {Data}                & {Pixel}         \\
{UAP\cite{moosavi2017universal}}               & {CVPR 2017}            & {Image Classification} & {Digital}         & {Invisible}           & {Meaningless}        & {White}              & {Untargeted}       & {Universal}             & {Global}            & {Data}                & {Pixel}         \\
{ZOO\cite{chen2017zoo}}               & {ACM AISec 2017}       & {Image Classification} & {Digital}         & {Invisible}           & {Meaningless}        & {Black}              & {Both}             & {Specific}              & {Global}            & {Data}                & {Pixel}         \\
{C\&W\cite{carlini2017towards}}              & {S\&P 2017}            & {Image Classification} & {Digital}         & {Invisible}           & {Meaningless}        & {White}              & {Targeted}         & {Specific}              & {Global}            & {Data}                & {Pixel}         \\
{LaVAN\cite{karmon2018lavan}}             & {ICML 2018}            & {Image Classification} & {Digital}         & {Visible}             & {Meaningless}        & {White}              & {Targeted}         & {Universal}             & {Local}             & {Data}                & {Patch}         \\
{PGD\cite{madry2018towards}}               & {ICLR 2018}            & {Image Classification} & {Digital}         & {Invisible}           & {Meaningless}        & {Both}               & {Targeted}         & {Specific}              & {Global}            & {Data}                & {Pixel}         \\
{MI-FGSM\cite{dong2018boosting}}           & {CVPR 2018}            & {Image Classification} & {Digital}         & {Invisible}           & {Meaningless}        & {Both}               & {Both}             & {Specific}              & {Global}            & {Data}                & {Pixel}         \\
{Ilyas et al.\cite{ilyas2018black}}      & {ICML 2018}            & {Image Classification} & {Digital}         & {Invisible}           & {Meaningless}        & {Black}              & {Targeted}         & {Specific}              & {Global}            & {Data}                & {Pixel}         \\
{Ilyas et al.\cite{ilyas2019prior}}      & {ICLR 2019}            & {Image Classification} & {Digital}         & {Invisible}           & {Meaningless}        & {Black}              & {Untargeted}       & {Specific}              & {Global}            & {Data}                & {Pixel}         \\
{Su et al.\cite{su2019one}}         & {TEVC 2019}            & {Image Classification} & {Digital}         & {Invisible}           & {Meaningless}        & {Black}              & {Both}             & {Specific}              & {Local}             & {Data}                & {Pixel}         \\
{Ilyas et al.\cite{ilyas2019adversarial}}      & {NIPS 2019}            & {Image Classification} & {Digital}         & {Invisible}           & {Meaningless}        & {White}              & {Targeted}         & {Specific}              & {Global}            & {Data}                & {Pixel}         \\
{NATTACK\cite{li2019nattack}}           & {ICML 2019}            & {Image Classification} & {Digital}         & {Invisible}           & {Meaningless}        & {Black}              & {Targeted}         & {Specific}              & {Global}            & {Data}                & {Pixel}         \\
{Du et al.\cite{du2020query}}         & {ICLR 2020}            & {Image Classification} & {Digital}         & {Invisible}           & {Meaningless}        & {Black}              & {Both}             & {Specific}              & {Global}            & {Data}                & {Pixel}         \\
{MGAAttack\cite{wang2020mgaattack}}         & {ACM MM 2020}          & {Image Classification} & {Digital}         & {Invisible}           & {Meaningless}        & {Black}              & {Both}             & {Specific}              & {Global}            & {Data}                & {Pixel}         \\
{HopSkipJumpAttack\cite{chen2020hopskipjumpattack}} & {S\&P 2020}            & {Image Classification} & {Digital}         & {Invisible}           & {Meaningless}        & {Black}              & {Both}             & {Specific}              & {Global}            & {Data}                & {Pixel}         \\
{SAPF\cite{fan2020sparse}}              & {ECCV 2020}            & {Image Classification} & {Digital}         & {Invisible}           & {Meaningless}        & {White}              & {Targeted}         & {Specific}              & {Global}            & {Data}                & {Pixel}         \\
{AoA\cite{chen2020universal}}               & {TPAMI 2020}           & {Image Classification} & {Digital}         & {Invisible}           & {Meaningless}        & {Both}               & {Untargeted}       & {Specific}              & {Global}            & {Data}                & {Pixel}         \\
{Mahmood et al.\cite{mahmood2021robustness}}    & {ICCV 2021}            & {Image Classification} & {Digital}         & {Invisible}           & {Meaningless}        & {Both}               & {Both}             & {Specific}              & {Global}            & {Data}                & {Pixel}         \\
{SurFree\cite{maho2021surfree}}           & {CVPR 2021}            & {Image Classification} & {Digital}         & {Invisible}           & {Meaningless}        & {Black}              & {Both}             & {Specific}              & {Global}            & {Data}                & {Pixel}         \\
{Simulator Attack\cite{ma2021simulating}}  & {CVPR 2021}            & {Image Classification} & {Digital}         & {Invisible}           & {Meaningless}        & {Black}              & {Both}             & {Specific}              & {Global}            & {Data}                & {Pixel}         \\
{AdvDrop\cite{duan2021advdrop}}           & {ICCV 2021}            & {Image Classification} & {Digital}         & {Invisible}           & {Meaningless}        & {White}              & {Both}             & {Specific}              & {Global}            & {Data}                & {Pixel}         \\
{MCG\cite{yin2023generalizable}}               & {TPAMI 2023}           & {Image Classification} & {Digital}         & {Invisible}           & {Meaningless}        & {Black}              & {Both}             & {Specific}              & {Global}            & {Data}                & {Pixel}         \\
{CLPA\cite{zhao2022clpa}}              & {AAAI 2022}            & {Image Classification} & {Digital}         & {/}                   & {/}                  & {Both}               & {/}                & {/}                     & {/}                 & {Model}               & {/}             \\
{CISA\cite{shi2022query}}              & {TPAMI 2022}           & {Image Classification} & {Digital}         & {Invisible}           & {Meaningless}        & {Black}              & {Untargeted}       & {Specific}              & {Global}            & {Data}                & {Pixel}         \\
{GA\cite{liu2022towards}}                & {arXiv 2022}           & {Image Classification} & {Digital}         & {Invisible}           & {Meaningless}        & {Black}              & {/}                & {Specific}              & {Global}            & {Data}                & {Pixel}         \\
{ILA\cite{guo2022intermediate}}               & {TPAMI 2022}           & {Image Classification} & {Digital}         & {Invisible}           & {Meaningless}        & {Black}              & {Untargeted}       & {Specific}              & {Global}            & {Data}                & {Pixel}         \\
{MI-FAGSM\cite{wan2023average}}          & {TMM 2023}             & {Image Classification} & {Digital}         & {Invisible}           & {Meaningless}        & {Both}               & {/}                & {Specific}              & {Global}            & {Data}                & {Pixel}         \\
{PS Attack\cite{zhang2023improving}}         & {IS 2023}              & {Image Classification} & {Digital}         & {Invisible}           & {Meaningless}        & {Black}              & {Both}             & {Specific}              & {Global}            & {Data}                & {Pixel}         \\
{EA\cite{huang2023erosion}}                & {TIP 2023}             & {Image Classification} & {Digital}         & {Invisible}           & {Meaningless}        & {Black}              & {/}                & {Specific}              & {Global}            & {Data}                & {Corruption}    \\
{Shafahi et al.\cite{shafahi2018poison}}    & {NIPS 2018}            & {Image Classification} & {Digital}         & {/}                   & {/}                  & {White}              & {Targeted}         & {/}                     & {/}                 & {Model}               & {/}             \\
{Elsayed et al.\cite{elsayed2018adversarial}}    & {ICLR 2019}            & {Image Classification} & {Digital}         & {Visible}             & {Meaningless}        & {White}              & {Targeted}         & {Universal}             & {Local}             & {Both}                & {Trigger}       \\
{ISSBA\cite{li2021invisible}}             & {ICCV 2021}            & {Image Classification} & {Digital}         & {Invisible}           & {Meaningless}        & {White}              & {Targeted}         & {Specific}              & {Global}            & {Both}                & {Trigger}       \\
{Tao et al.\cite{tao2022better}}        & {CVPR 2022}            & {Image Classification} & {Digital}         & {Visible}             & {Meaningless}        & {White}              & {Targeted}         & {Universal}             & {Global}            & {Both}                & {Trigger}       \\
{Noppel et al.\cite{noppel2022disguising}}     & {S\&P 2023}            & {Image Classification} & {Digital}         & {Visible}             & {Meaningless}        & {White}              & {Targeted}         & {Universal}             & {Local}             & {Both}                & {Trigger}       \\
{Liu et al.\cite{liu2017delving}}        & {ICLR 2017}            & {Image Classification} & {Digital}         & {Invisible}           & {Meaningless}        & {Both}               & {Both}             & {Specific}              & {Global}            & {Data}                & {Pixel}         \\
{Demontis et al.\cite{demontis2019adversarial}}   & {USENIX Security 2019} & {Image Classification} & {Digital}         & {Invisible}           & {Meaningless}        & {Both}               & {Targeted}         & {/}                     & {/}                 & {Data}                & {Pixel}         \\
{ILA\cite{huang2019enhancing}}               & {ICCV 2019}            & {Image Classification} & {Digital}         & {Invisible}           & {Meaningless}        & {Both}               & {Targeted}         & {Specific}              & {Global}            & {Data}                & {Pixel}         \\
{Xie et al.\cite{xie2019improving}}        & {CVPR 2019}            & {Image Classification} & {Digital}         & {Invisible}           & {Meaningless}        & {Both}               & {Untargeted}       & {Specific}              & {Global}            & {Data}                & {Pixel}         \\
{TTP\cite{naseer2021generating}}               & {ICCV 2021}            & {Image Classification} & {Digital}         & {Invisible}           & {Meaningless}        & {Both}               & {Targeted}         & {Specific}              & {Global}            & {Data}                & {Pixel}         \\
{PRGF\cite{dong2021query}}              & {TPAMI 2021}           & {Image Classification} & {Digital}         & {Invisible}           & {Meaningless}        & {Both}               & {Untargeted}       & {Specific}              & {Global}            & {Data}                & {Pixel}         \\
{Admix\cite{wang2021admix}}             & {ICCV 2021}            & {Image Classification} & {Digital}         & {Invisible}           & {Meaningless}        & {Both}               & {Untargeted}       & {Specific}              & {Global}            & {Data}                & {Pixel}         \\
{FIA\cite{wang2021feature}}               & {ICCV 2021}            & {Image Classification} & {Digital}         & {Invisible}           & {Meaningless}        & {Both}               & {Untargeted}       & {Specific}              & {Global}            & {Data}                & {Pixel}         \\
{SMI-FGSM\cite{wang2022enhancing}}          & {arXiv 2022}           & {Image Classification} & {Digital}         & {Invisible}           & {Meaningless}        & {Both}               & {Untargeted}       & {Specific}              & {Global}            & {Data}                & {Pixel}         \\
{NAA\cite{zhang2022improving}}               & {CVPR 2022}            & {Image Classification} & {Digital}         & {Invisible}           & {Meaningless}        & {Both}               & {Untargeted}       & {Specific}              & {Global}            & {Data}                & {Pixel}         \\
{TSAA\cite{he2021transferable}}              & {arXiv 2021}           & {Image Classification} & {Digital}         & {Invisible}           & {Meaningless}        & {Both}               & {Both}             & {Specific}              & {Global}            & {Data}                & {Pixel}         \\
{SVRE\cite{xiong2022stochastic}}              & {CVPR 2022}            & {Image Classification} & {Digital}         & {Invisible}           & {Meaningless}        & {Both}               & {Untargeted}       & {Specific}              & {Global}            & {Data}                & {Pixel}         \\
{Top-k Attack\cite{zhang2022investigating}}      & {CVPR 2022}            & {Image Classification} & {Digital}         & {Invisible}           & {Meaningless}        & {Both}               & {Both}             & {Specific}              & {Global}            & {Data}                & {Pixel}         \\
{DSM\cite{yang2022boosting}}               & {arXiv 2022}           & {Image Classification} & {Digital}         & {Invisible}           & {Meaningless}        & {Both}               & {Both}             & {Specific}              & {Global}            & {Data}                & {Pixel}  
\\                 
\hline\hline
\end{tabular*}
    \begin{tablenotes}
        \footnotesize      
        \item "/"  represents "Not applicable".
    \end{tablenotes}
\end{threeparttable}
\end{table*}

\ding{172} \textbf{Digital attack.} 

\textbf{White-box attacks:} Szegedy \etal \cite{szegedy2014intriguing} first reveal that DNNs establish input-output associations characterized by a considerable degree of discontinuity. 
More precisely, their findings indicate that the application of a subtle and imperceptible perturbation, identified by maximizing the network's prediction error, can effectively induce DNNs' misclassification.
FGSM \cite{ian2015explaining} was the first adversarial optimization attack method, in which only one step was moved from benign image $\boldsymbol{x}$ following the sign of gradient with the step size $\epsilon$ to obtain adversarial image $\boldsymbol{x}^*$.
In \cite{moosavi2016deepfool}, the proposed DeepFool algorithm effectively generates perturbations that deceive DNNs and initially evaluates the resilience of state-of-the-art (SOTA) deep classifiers to adversarial perturbations on large-scale datasets.
Papernot \etal \cite{papernot2016limitations} present a formalization of the adversarial space \wrt DNNs and introduce a novel set of algorithms that generate adversarial examples through a comprehensive comprehension of the input-output mapping of DNNs.
\cite{sarkar2017upset} achieves targeted deception of high-performance image classifiers through the development of two innovative attack techniques. The first technique (Universal Perturbations for Steering to Exact Targets, UPSET) generates universal perturbations for specific target classes, while the second technique (Antagonistic Network for Generating Rogue Images, ANGRI) generates perturbations that are specific to individual images.
The authors of \cite{moosavi2017universal} demonstrate the existence of universal (image-agnostic) and invisible adversarial noise, which reveals important geometric correlations among the high-dimensional decision boundary of classifiers.
Moreover, the universal adversarial noises can generalize well across DNNs.
In \cite{karmon2018lavan}, the researchers explore the case where the noise is allowed to be visible but confined to a small, localized patch of the image, without covering any of the main object(s) in the image, named Localized and Visible Adversarial Noise (LaVAN).
\cite{ilyas2019adversarial} indicates that the existence of non-robust features is directly responsible for the emergence of adversarial examples and confirms their widespread prevalence in commonly used datasets.
Fan \etal \cite{fan2020sparse} investigate sparse adversarial attacks, which focus on generating adversarial perturbations on select regions of a benign image.
In research \cite{mahmood2021robustness}, the authors investigate the resilience of Vision Transformers to adversarial examples and their transferability between CNNs and transformers.
Paper \cite{ma2021understanding} offers a more comprehensive comprehension of adversarial examples concerning medical images.
Research \cite{tang2019adversarial} introduces a new form of adversarial attack that is capable of deceiving classifiers through significant alterations. 
For instance, even after significant changes to a face, well-trained DNNs are still can identify both the adversarial and original example as the same person.
To find out if the performance of DNNs decreases even for the images only lose a little information, Duan \etal \cite{duan2021advdrop} propose AdvDrop to craft adversarial examples by dropping part information of images.
Akhtar \etal \cite{akhtar2021attack} present a practical adversarial attack that can perform targeted fooling of deep visual classifiers on a per-class basis. Furthermore, they adapt this attack to interpret deep representations.
To avoid losing useful statistical information in boundary attacks, \cite{sang2022enhancing} investigates and enhances boundary attacks by restricting the perturbation direction in a square shape in the geometrical presentation of the image.
Research \cite{deng2022frequency} introduces a frequency-tuned universal attack method that employs the frequency domain to generate universal perturbations to attack DNNs-based texture recognition systems.
Wan \etal \cite{wan2023average} devise a new average gradient-based adversarial attack, in which a dynamic set of adversarial examples is constructed in each iteration by utilizing the gradient of each iteration to calculate the average gradient.
Perceptual Sensitive Attack (PS Attack) \cite{zhang2023improving} is introduced to avoid perturbations that are easily spotted by human eyes.

\textbf{Black-box attacks:} 
To avoid the demands for knowledge of either the model internals or its training data, the authors in \cite{papernot2017practical} introduce the first practical demonstration of an attacker controlling a remotely hosted DNN with no such knowledge by observing labels given by the DNN to chosen inputs.
In work \cite{chen2017zoo}, Zeroth Order Optimization (ZOO) is devised to directly estimate the gradients of the proxy model for crafting adversarial examples.
Specifically, They employ a combination of zeroth-order stochastic coordinate descent, dimension reduction, hierarchical attack, and importance sampling techniques to effectively fool black-box models.
By introducing three novel attack algorithms that can successfully penetrate both distilled and undistilled neural networks, Carlini \etal \cite{carlini2017towards} establish that defensive distillation does not notably enhance the resilience of neural networks.
To strengthen black-box attack efficacy, Dong \etal \cite{dong2018boosting} propose a momentum iterative FGSM (MI-FGSM) by integrating the momentum term into the iterative process of noise optimization, which can stabilize update directions and escape from poor local maxima during the optimization, resulting in more transferable adversarial examples.
Ilyas \etal \cite{ilyas2018black} establish three practical threat models that more precisely reflect the nature of many real-world classifiers: the query-limited model, the partial-information model, and the label-only model. 
Furthermore, they propose novel attack strategies that can deceive classifiers under these more restrictive threat models.
In the paper \cite{mopuri2018generalizable}, the authors propose a novel and data-free method for generating universal adversarial perturbations that can be applied across multiple vision tasks. 
Technically, their approach involves corrupting the extracted features at multiple layers to achieve fooling, which makes the objective generalizable and applicable to image-agnostic perturbations for various vision tasks, including object recognition, semantic segmentation, and depth estimation.
Work \cite{ilyas2019prior} proposes a framework that integrates and unifies a substantial portion of the existing research on black-box attacks and shows how to enhance the performance of black-box attacks by introducing gradient priors as a new factor in the problem.
Su \etal \cite{su2019one} analyze an attack in an extremely limited scenario where only one pixel can be modified. 
To achieve this, they propose a novel method for generating one-pixel adversarial perturbations based on differential evolution (DE). 
Moreover, this method requires minimal adversarial information, making it a black-box attack, and is capable of fooling a wider range of networks due to the inherent characteristics of DE.
Article \cite{li2019nattack} introduces a black-box adversarial attack algorithm that can successfully bypass both standard DNNs and those generated by various recently developed defense techniques, in which adversarial examples are drawn from the probability density distribution over a small region centered around the inputs.
In \cite{du2020query}, a meta attack is devised to attack a targeted model with few queries.
\cite{wang2020mgaattack} strengthens query efficiency by leveraging the advantages of both transfer-based and scored-based approaches and addressing a discretized problem through the utilization of a simple yet highly efficient microbial genetic algorithm (MGA).
The authors in \cite{chen2020hopskipjumpattack} present the HopSkipJumpAttack family of algorithms, which rely on a novel estimate of the gradient direction obtained through binary information at the decision boundary.
SurFree is presented in \cite{maho2021surfree} to decrease the number of queries by focusing on targeted trials along varied directions, guided by precise indications of the geometric properties of the decision boundaries.
The objective of study \cite{ma2021simulating} is to train a generalizable surrogate model, termed "Simulator," capable of emulating the behavior of an unknown target model.
To mitigate the query cost, the authors of \cite{yin2023generalizable} suggest using feedback information obtained from past attacks, \ie example-level adversarial transferability. 
By considering each attack on a benign example as an individual task, they construct a meta-learning framework that involves training a meta-generator to produce perturbations based on specific benign examples.
The authors of research \cite{shi2022query} introduce a novel framework for conducting query-efficient black-box adversarial attacks by integrating transfer-based and decision-based approaches. 
They also elucidate the correlation between the present noise and sampling variance, the compression monotonicity of noise, and the impact of transition functions on decision-based attacks.
Guo \etal \cite{guo2022intermediate} introduce an intermediate-level attack, which establishes a direct linear mapping from the intermediate-level discrepancies, \ie between adversarial features and benign features, to prediction loss of the adversarial example.
To strengthen attacks' transferability against black-box defenses, \cite{huang2023erosion} propose a novel transferable attack capable of defeating various black-box defenses and sheds light on their security limitations.

Finally, we summarize digital attacks against image classification (\cite{chen2018ead,tu2019autozoom,li2020open,bai2021targeted,bai2022hardly,bai2022versatile,wang2020hamiltonian,cai2022blackbox,liu2022practical,szegedy2014intriguing,ian2015explaining,moosavi2016deepfool,papernot2016limitations,sarkar2017upset,papernot2017practical,moosavi2017universal,chen2017zoo,carlini2017towards,karmon2018lavan,madry2018towards,dong2018boosting,ilyas2018black,ilyas2019prior,su2019one,ilyas2019adversarial,li2019nattack,du2020query,wang2020mgaattack,chen2020hopskipjumpattack,fan2020sparse,chen2020universal,mahmood2021robustness,maho2021surfree,ma2021simulating,duan2021advdrop,yin2023generalizable,zhao2022clpa,shi2022query,liu2022towards,guo2022intermediate,wan2023average,zhang2023improving,huang2023erosion,shafahi2018poison,elsayed2018adversarial,li2021invisible,tao2022better,noppel2022disguising,liu2017delving,demontis2019adversarial,huang2019enhancing,xie2019improving,naseer2021generating,dong2021query,wang2021admix,wang2021feature,wang2022enhancing,zhang2022improving,he2021transferable,xiong2022stochastic,zhang2022investigating,yang2022boosting}) in Table \ref{table:digital_classification}.

\ding{173} \textbf{Physical attack}

In \cite{kurakin2018adversarial}, the authors first demonstrate that machine learning systems are vulnerable to adversarial examples even in physical world scenarios and propose a basic iterative method (BIM).
Brown \etal \cite{brown2017adversarial} propose a method for generating universal, robust, targeted adversarial perturbations in patch form that can be deployed in the real world.
These adversarial patches can be printed, attached, photographed, and then presented to image classifiers for successful attacks.
Subsequently, adversarial patches are broadly adopted in various physical attacks \cite{lian2022benchmarking,lian2023cba,thys2019fooling,hu2021naturalistic,wang2022fca,wei2022simultaneously,wei2022adversarial}.
To better understand adversarial examples in the physical world, Eykholt \etal \cite{eykholt2018robust} propose a general physical attack method, Robust Physical Perturbations (RP2), to elaborate robust visual adversarial perturbations under dynamic physical conditions.
\cite{athalye2018synthesizing} provides evidence for the existence of robust 3D adversarial objects, and introduces the first algorithm Expectation Over Transformation (EOT) capable of synthesizing examples that remain adversarial across a chosen distribution of transformations.
\cite{zeng2019adversarial} focuses specifically on the subset of adversarial examples that correspond to meaningful changes in 3D physical properties, such as rotation, translation, illumination conditions, \etc.
To alleviate unrealistic distortions of adversarial patterns, Duan \etal \cite{duan2020adversarial} introduces a novel technique called Adversarial Camouflage (AdvCam), which involves crafting and camouflaging physical-world adversarial examples in natural styles that appear legitimate to human observers.
 In \cite{feng2021meta}, Feng \etal propose Meta-Attack by formulating physical attacks as a few-shot learning problem to improve the optimization efficiency of physical dynamic simulations.
\cite{gnanasambandam2021optical} propose an optical adversarial attack, which uses structured illumination to alter the appearance of the target objects to deceive image classifiers without physically touching the targeted objects, \eg moving or painting the targets of interest.
Duan \etal \cite{duan2021adversarial} demonstrates that DNNs can be easily deceived using only a laser beam.
Research \cite{doan2022tnt} uncovers the presence of an intriguing category of spatially constrained, physically feasible adversarial examples, \ie Universal NaTuralistic adversarial paTches (TnTs). 
TnTs are crafted by examining the full range of spatially bounded adversarial examples and the natural input space within generative adversarial networks (GANs).

Finally, we summarize physical attacks against image classification (\cite{dong2022viewfool,akhtar2021attack,brown2017adversarial,kurakin2018adversarial,athalye2018synthesizing,jan2019connecting,duan2020adversarial,feng2021meta,gnanasambandam2021optical,duan2021adversarial,hu2022adversarialColor,liu2019perceptual,hu2022adversarialZoom,mathov2022enhancing,byun2022improving}) in Table \ref{table:physical_classification}.

\begin{table*}[t!]
\scriptsize
    \renewcommand{\arraystretch}{1.25}
\caption{Physical attacks against image classification.}
\label{table:physical_classification}
\centering
\setlength{\tabcolsep}{1.0mm}
\begin{threeparttable}
\begin{tabular*}{\hsize}{ccccccccccccc}
\hline\hline
\thead{\textbf{Methods}} & {\textbf{Venue}} & {\textbf{Tasks}} & {\textbf{Domain}} & {\textbf{Visibility}} & {\textbf{Semantics}} & {\textbf{Knowledge}} & {\textbf{Purpose}} & {\textbf{Universality}} & {\textbf{Position}} & {\textbf{Distortion}} & {\textbf{Form}} 
\\ \hline
{ViewFool\cite{dong2022viewfool}}       & NIPS 2022           & Image Classification & Physical & /         & /           & Both  & Untargeted & Specific  & /      & Data & Viewpoint \\
{Akhtar et al.\cite{akhtar2021attack}}  & TPAMI 2021          & Image Classification & Both     & Both      & Meaningless & White & Targeted   & Specific  & Global & Data & Pixel     \\
{Brown et al.\cite{brown2017adversarial}}   & arXiv 2017          & Image Classification & Physical & Visible   & Meaningless & White & Targeted   & Universal & Local  & Data & Patch     \\
{Kurakin et al.\cite{kurakin2018adversarial}} & AISS 2018           & Image Classification & Physical & Visible   & Meaningless & Black & /          & Specific  & Global & Data & Pixel     \\
{EOT\cite{athalye2018synthesizing}}            & ICML 2018           & Image Classification & Physical & Visible   & Meaningless & White & Targeted   & Universal & Local  & Data & Patch     \\
{D2P\cite{jan2019connecting}}            & AAAI 2019           & Image Classification & Physical & Visible   & Meaningless & White & Targeted   & Universal & Global & Data & Pixel     \\
{AdvCam\cite{duan2020adversarial}}         & CVPR 2020           & Image Classification & Both     & Visible   & Meaningful  & White & Both       & Universal & Local  & Data & Style     \\
{Meta-Attack\cite{feng2021meta}}    & ICCV 2021           & Image Classification & Both     & Visible   & Meaningless & White & Targeted   & Specific  & Global & Data & Pixel     \\
{OPAD\cite{gnanasambandam2021optical}}           & ICCV 2021           & Image Classification & Physical & Visible   & Meaningless & Both  & Both       & Universal & Local  & Data & Light     \\
{AdvLB\cite{duan2021adversarial}}          & CVPR 2021           & Image Classification & Both     & Visible   & Meaningless & Black & Untargeted & Universal & Local  & Data & Laser     \\
{AdvCF\cite{hu2022adversarialColor}}          & arXiv 2022          & Image Classification & Physical & Visible   & Meaningless & Both  & Targeted   & Universal & Global & Data & Color     \\
{AdvZL\cite{hu2022adversarialZoom}}          & arXiv 2022          & Image Classification & Physical & /         & /           & White & Untargeted & Specific  & /      & Data & Zoom      \\
{PS-GAN\cite{liu2019perceptual}}         & AAAI 2019           & Image Classification & Physical & Visible   & Meaningless & Both  & Untargeted & Universal & Local  & Data & Patch     \\
{Mathov et al.\cite{mathov2022enhancing}}  & Neurocomputing 2022 & Image Classification & Both     & Visible   & Meaningless & White & /          & Universal & Local  & Data & Patch     \\
{ODI\cite{byun2022improving}}            & CVPR 2022           & Image Classification & Both     & Invisible & Meaningless & Both  & Targeted   & Specific  & Global & Data & Pixel
\\                 
\hline\hline
\end{tabular*}
    \begin{tablenotes}
        \footnotesize      
        \item "/"  represents "Not applicable".
    \end{tablenotes}
\end{threeparttable}
\end{table*}

\subsubsection{Object Detection}

In this section, we offer a comprehensive examination of adversarial attacks pertaining to object detection, focusing specifically on digital and physical attack strategies.
Given the practicality of adversarial attacks in object detection tasks, much of the current research focuses on physical attacks.

\ding{172} \textbf{Digital attack}

\textbf{White-box attacks:} 
In \cite{xie2017adversarial}, the authors extend the concept of adversarial examples to the domains of semantic segmentation and object detection, which are notably more challenging tasks.
Specifically, they introduce a novel algorithm called Dense Adversary Generation (DAG), which optimizes a loss function over a set of pixels or proposals to generate adversarial perturbations. 
To reduce the number of perturbed pixels, \cite{wu2020dpattack} presents a new technique known as the Diffused Patch Attack (DPAttack), which leverages diffused patches in the form of asteroid-shaped or grid-shaped patterns to deceive object detectors. 
This attack only modifies a small number of pixels in the image.
Research \cite{zhang2020contextual} introduces a novel approach called Contextual Adversarial Perturbation (CAP), which targets contextual information of objects in order to degrade the recognition accuracy of object detectors.
Zhang \etal \cite{zhang2021fooling} introduce a novel Half-Neighbor Masked Projected Gradient Descent (HNM-PGD) approach, capable of generating potent perturbations to deceive various detectors while adhering to stringent limitations.
\cite{shi2021adversarial} presents a new and distinctive patch configuration comprised of four intersecting lines. 
The proposed patch shape is shown to be a powerful tool for influencing deep convolutional feature extraction with limited pixel availability.
To ensure the stability of the ensemble attack, Huang \etal \cite{huang2021rpattack} present a gradient balancing technique that prevents any single detector from being over-optimized during the training process.
Furthermore, they propose a novel patch selection and refining mechanism that identifies the most crucial pixels for the attack, while gradually eliminating irrelevant perturbations.

\textbf{Black-box attacks:} 
Liu \etal \cite{liu2018dpatch} introduce DPATCH, a black-box adversarial-patch-based attack designed to target popular object detectors, such as Faster R-CNN \cite{ren2015faster} and YOLO \cite{jocher2020yolov5,redmon2018yolov3,redmon2017yolo9000}.
In contrast to the original adversarial patch, which only manipulates the image-level classifier, the DPATCH simultaneously targets both the bounding box regression and object classification of the object detector in order to disable their predictions.
\cite{liu2020efficient} introduces Efficient Warm Restart Adversarial Attack for Object Detection, which comprises three modules: 
Efficient Warm Restart Adversarial Attack, which selects the most appropriate top-k pixels for the attack;
Connecting Top-k pixels with Lines, which outlines the strategy for connecting two top-k pixels to minimize the number of changed pixels and reduce the number of patches;
Adaptive Black Box Optimization, which leverages white box models to improve the performance of the black box adversarial attack.
To fool context-aware detectors, Cai \etal \cite{cai2022zero} introduce the pioneering method for producing context-consistent adversarial attacks that can elude the context-consistency check of black-box object detectors working on intricate and natural scenes.

Finally, we summarize digital attacks against object detection (\cite{liu2018dpatch,wu2020dpattack,zhang2020contextual,zhang2021fooling,shi2021adversarial,huang2021rpattack,chow2020adversarial,cai2022zero}) in Table \ref{table:digital_detection}.

\begin{table*}[t!]
\scriptsize
    \renewcommand{\arraystretch}{1.25}
\caption{Digital attacks against object detection.}
\label{table:digital_detection}
\centering
\setlength{\tabcolsep}{1.7mm}
\begin{threeparttable}
\begin{tabular*}{\hsize}{ccccccccccccc}
\hline\hline
\thead{\textbf{Methods}} & {\textbf{Venue}} & {\textbf{Tasks}} & {\textbf{Domain}} & {\textbf{Visibility}} & {\textbf{Semantics}} & {\textbf{Knowledge}} & {\textbf{Purpose}} & {\textbf{Universality}} & {\textbf{Position}} & {\textbf{Distortion}} & {\textbf{Form}} 
\\ \hline
DPATCH\cite{liu2018dpatch} & arXiv 2019   & Object Detection & Digital & Visible   & Meaningless & Black & Both       & Universal & Local  & Data & Patch \\
DPAttack\cite{wu2020dpattack}                      & arXiv 2020   & Object Detection & Digital & Visible   & Meaningless & White & Untargeted & Specific  & Local  & Data & Patch \\
CAP\cite{zhang2020contextual}                           & ICME 2020    & Object Detection & Digital & Invisible & Meaningless & White & Untargeted & Specific  & Global & Data & Pixel \\
HNM-PGD\cite{zhang2021fooling}                       & arXiv 2021   & Object Detection & Digital & Visible   & Meaningless & White & Untargeted & Specific  & Local  & Data & Patch \\
DTTACK\cite{shi2021adversarial}                        & ICASSP 2021  & Object Detection & Digital & Visible   & Meaningless & White & Targeted   & Specific  & Local  & Data & Patch \\
RPAttack\cite{huang2021rpattack}                      & ICME 2021    & Object Detection & Digital & Invisible & Meaningless & White & Untargeted & Specific  & Local  & Data & Patch \\
TOG\cite{chow2020adversarial}                           & TPS-ISA 2020 & Object Detection & Digital & Invisible & Meaningless & Both  & Both       & Specific  & Global & Data & Pixel \\
ZQA\cite{cai2022zero}                           & CVPR 2022    & Object Detection & Digital & Invisible & Meaningless & Black & /          & Specific  & Global & Data & Pixel
\\                 
\hline\hline
\end{tabular*}
    \begin{tablenotes}
        \footnotesize      
        \item "/"  represents "Not applicable".
    \end{tablenotes}
\end{threeparttable}
\end{table*}

\ding{173} \textbf{Physical attack}

Lu \etal \cite{lu2017adversarial} present a construction that effectively deceives two commonly used detectors, Faster RCNN \cite{ren2015faster} and YOLO 9000 \cite{redmon2017yolo9000}, in the physical world.
\cite{song2018physical} extend physical attacks to object detection by implementing a Disappearance Attack, which causes a stop sign to "disappear" either by covering the sign with an adversarial poster or by adding adversarial stickers onto the sign.
The work \cite{chen2019shapeshifter} introduces ShapeShifter, and demonstrates that the EOT approach, initially proposed to improve the resilience of adversarial perturbations in image classification, can be effectively adapted to the object detection domain.
\cite{thys2019fooling} proposes a method for generating adversarial patches that can effectively conceal individuals from person detectors. 
This method is particularly designed for targets with a high degree of intra-class variety, such as persons.
In \cite{zhang2019camou}, the authors present an intriguing experimental investigation of physical adversarial attacks on object detectors in real-world scenarios. 
Specifically, they explore the efficacy of learning a camouflage pattern to obscure vehicles from being detected by SOTA detectors based on DNNs.
To generate visually natural patches with strong attacking ability, Liu \etal \cite{liu2019perceptual} present a novel Perceptual-Sensitive Generative Adversarial Network (PS-GAN) that can simultaneously enhance the visual authenticity and the attacking potential of the adversarial patch.
Wang \etal \cite{wang2019advpattern} take the first attempt to implement robust physical-world attacks against  person re-identification systems based on DNNs.
They propose advPattern to generate adversarial patches on clothes, which can hide people from being detected.
In \cite{huang2020universal}, the authors study physical attacks against object detectors in the wild.
They propose the Universal Physical Camouflage Attack (UPC), which involves learning an adversarial pattern capable of effectively attacking all instances of a given object category.
Wu \etal \cite{wu2020making} present a systematic study of the transferability of adversarial attacks on SOTA object detection frameworks.
To avoid direct access to targets of interest, \cite{zolfi2021translucent} presents a novel contactless and translucent patch containing a carefully crafted pattern, which is placed over the lens of the camera to deceive SOTA object detectors.
Zhu \etal \cite{zhu2021fooling} first demonstrate the feasibility of using two types of patches to launch an attack on YOLOv3-based infrared pedestrian detectors.
Following the previous work \cite{zhu2021fooling}, \cite{zhu2022infrared} propose the infrared adversarial clothing by simulating the process from cloth to clothing in the digital world and then designing the adversarial "QR code" pattern. 
\cite{hu2022adversarial} introduces a novel approach called Adversarial Texture (AdvTexture) for conducting multi-angle attacks against person detectors. 
AdvTexture enables the coverage of clothes with arbitrary shapes, rendering individuals wearing such clothes invisible to person detectors from various viewing angles.
In \cite{suryanto2022dta}, the authors introduce the Differentiable Transformation Attack (DTA), which enables the creation of patterns that can effectively hide the object from detection, while also taking into account the impact of various transformations that the object may undergo. 
Wang \etal \cite{wang2023transpatch} introduce a novel training pipeline called TransPatch to optimize the training efficiency of adversarial patches. 
To avoid generating conspicuous and attention-grabbing patterns, \cite{hu2021naturalistic} propose to create physical adversarial patches by leveraging the image manifold of a pre-trained GAN.
Inspired by the viewpoint that attention is indicative of the underlying recognition process, \cite{wang2021dual} proposes the Dual Attention Suppression (DAS) attack to craft visually-natural physical adversarial camouflages. 
The DAS achieves strong transferability by suppressing both model and human attention, thereby enhancing the efficacy of the attack.
In \cite{shapira2022attacking}, the researchers propose a novel targeted and universal attack against the SOTA object detector using a label-switching technique. 
The attack aims to fool the object detector into misclassifying a specific target object as another object category chosen by the attacker.
Mathov \etal \cite{mathov2022enhancing} introduce a novel framework that leverages 3D modeling to generate adversarial patches for a pre-existing real-world scene. 
By employing a 3D digital approximation of the scene, their methodology effectively simulates the real-world environment.
To bridge the divide between digital and physical attacks, Wang \etal \cite{wang2022fca} utilize the entire 3D surface of a vehicle to propose a resilient Full-coverage Camouflage Attack (FCA) that effectively deceives detectors.
A universal background adversarial attack method \cite{xu2022universal} is devised to fool DNNs-based object detectors. 
The proposed method involves placing target objects onto a universal background image and manipulating the local pixel data surrounding the target objects in a way that renders them unrecognizable by object detectors.
The focus of the study \cite{han2022physical} is on the lane detection system, a crucial component in numerous autonomous driving applications, such as navigation and lane switching.
The researchers design and realize the first physical backdoor attacks on such systems.
Zhang \etal \cite{zhang2022transferable} propose a novel approach for producing physically feasible adversarial camouflage to achieve transferable attacks on detection models.
Study \cite{zhong2022shadows} explores a new category of optical adversarial examples, generated by a commonly occurring natural phenomenon, shadows. 
They aim to employ these shadow-based perturbations to achieve naturalistic and inconspicuous physical-world adversarial attacks in black-box settings.
A systematic pipeline is introduced in \cite{jia2022fooling} to produce resilient physical adversarial examples that can effectively deceive real-world object detectors.
Zhu \etal \cite{zhutpatch} present TPatch, a physical adversarial patch that is triggered by acoustic signals. 
TPatch differs from other adversarial patches in that it remains benign under ordinary circumstances but can be activated to initiate hiding, altering, or creating attacks via a deliberate distortion introduced through signal injection attacks directed at cameras.
To improve the optimizing stability and efficiency, the study \cite{guesmi2023advart} presents a fresh and lightweight framework that generates naturalistic adversarial patches systematically, without relying on GANs.
In paper\cite{liu2023x}, the authors conduct the first investigation towards adversarial attacks that are directed at X-ray prohibited item detection and demonstrate the grave hazards posed by such attacks in this context of paramount safety significance.

Finally, we summarize physical attacks against object detection (\cite{chen2019shapeshifter,huang2020universal,hu2021naturalistic,wang2022fca,xu2022universalviaBackground,zhang2022transferable,guesmi2023advart,lu2017adversarial,thys2019fooling,zhang2019camou,wu2020making,zolfi2021translucent,zhu2021fooling,wang2023transpatch,zhu2022infrared,hu2022adversarialfooling,suryanto2022dta,liu2023x,song2018physical,shapira2022attacking,chan2023baddet,huang2022t,zhang2023boosting,wen2023light}) in Table \ref{table:physical_detection}.

\begin{table*}[t!]
\scriptsize
    \renewcommand{\arraystretch}{1.25}
\caption{Physical attacks against object detection.}
\label{table:physical_detection}
\centering
\setlength{\tabcolsep}{0.6mm}
\begin{threeparttable}
\begin{tabular*}{\hsize}{ccccccccccccc}
\hline\hline
\thead{\textbf{Methods}} & {\textbf{Venue}} & {\textbf{Tasks}} & {\textbf{Domain}} & {\textbf{Visibility}} & {\textbf{Semantics}} & {\textbf{Knowledge}} & {\textbf{Purpose}} & {\textbf{Universality}} & {\textbf{Position}} & {\textbf{Distortion}} & {\textbf{Form}} 
\\ \hline
ShapeShifter\cite{chen2019shapeshifter} & ECML 2019        & Object Detection & Physical & Visible & Meaningless & Both  & Targeted   & Universal & Local & Data & Patch     \\
UPC\cite{huang2020universal}                                 & CVPR 2020        & Object Detection & Physical & Visible & Meaningful  & White & Both       & Universal & Local & Data & Patch     \\
Hu et al.\cite{hu2021naturalistic}                           & ICCV 2021        & Object Detection & Physical & Visible & Meaningful  & White & Untargeted & Universal & Local & Data & Patch     \\
FCA\cite{wang2022fca}                                 & AAAI 2022        & Object Detection & Physical & Visible & Meaningless & Both  & B & Universal & Local & Data & Patch     \\
Xu et al.\cite{xu2022universalviaBackground}                           & ACNS 2022        & Object Detection & Physical & Visible & Meaningless & Both  & Untargeted & Universal & Local & Data & Patch     \\
Zhang et al.\cite{zhang2022transferable}                        & arXiv 2022       & Object Detection & Physical & Visible & Meaningless & Black & Untargeted & Universal & Local & Data & Patch     \\
AdvART\cite{guesmi2023advart}                              & arXiv            & Object Detection & Physical & Visible & Meaningful  & White & Untargeted & Universal & Local & Data & Patch     \\
Lu et al.\cite{lu2017adversarial}                           & arXiv            & Object Detection & Both     & Visible & Meaningless & White & Both       & Universal & Both  & Data & Pixel     \\
Thys et al.\cite{thys2019fooling}                         & CVPR 2019        & Object Detection & Physical & Visible & Meaningless & White & Untargeted & Universal & Local & Data & Patch     \\
CAMOU\cite{zhang2019camou}                               & ICLR 2019        & Object Detection & Physical & Visible & Meaningless & Black & Untargeted & Universal & Local & Data & Patch     \\
Wu et al.\cite{wu2020making}                           & ECCV 2020        & Object Detection & Physical & Visible & Meaningless & White & Untargeted & Universal & Local & Data & Patch     \\
Translucent Patch\cite{zolfi2021translucent}                   & CVPR 2021        & Object Detection & Both     & Visible & Meaningless & Both  & Untargeted & Universal & Local & Data & Patch     \\
Zhu et al.\cite{zhu2021fooling}                          & AAAI 2021        & Object Detection & Both     & Visible & Meaningless & White & Untargeted & Universal & Local & Data & Patch     \\
TransPatch\cite{wang2023transpatch}                          & ECCV 2022        & Object Detection & Physical & Visible & Meaningless & White & Untargeted & Universal & Local & Data & Patch     \\
Zhu et al.\cite{zhu2022infrared}                          & CVPR 2022        & Object Detection & Physical & Visible & Meaningless & White & Untargeted & Universal & Local & Data & Patch     \\
AdvTexture\cite{hu2022adversarialfooling}                          & CVPR 2022        & Object Detection & Physical & Visible & Meaningless & White & Untargeted & Universal & Local & Data & Texture   \\
DTA\cite{suryanto2022dta}                                 & CVPR 2022        & Object Detection & Physical & Visible & Meaningless & White & Untargeted & Universal & Local & Data & Patch     \\
X-Adv\cite{liu2023x}                               & arXiv 2023       & Object Detection & Physical & Visible & Meaningless & White & Untargeted & Universal & Local & Data & 3D Object \\
Eykholt et al.\cite{song2018physical}                      & USENIX WOOT 2018 & Object Detection & Physical & Visible & Meaningless & White & Both       & Universal & Local & Data & Patch     \\
UTLSP\cite{shapira2022attacking}                               & arXiv 2022       & Object Detection & Physical & Visible & Meaningless & White & Targeted   & Universal & Local & Data & Patch     \\
BadDet\cite{chan2023baddet}                              & ECCV 2022        & Object Detection & Both     & Visible & Meaningful  & White & Both       & Universal & Local & Both & Trigger   \\
T-SEA\cite{huang2022t}                               & arXiv 2022       & Object Detection & Both     & Visible & Meaningless & Both  & /          & Universal & Local & Data & Patch     \\
Zhang et al.\cite{zhang2023boosting}                        & PR 2023          & Object Detection & Both     & Visible & Meaningless & Both  & Untargeted & Universal & Local & Data & Patch 
\\                 
Wen et al.\cite{wen2023light} & ICASSP 2023 & Object Detection & Physical & Visible & Meaningless & Both & Untargeted & Universal & Local & Data & Patch/Projection
\\
\hline\hline
\end{tabular*}
    \begin{tablenotes}
        \footnotesize      
        \item "/"  represents "Not applicable".
    \end{tablenotes}
\end{threeparttable}
\end{table*}

\subsubsection{Face Recognition}

In this section, we undertake a thorough assessment of adversarial attacks in the context of face recognition. 
The practicality of adversarial attacks in face recognition tasks has resulted in a significant focus on physical attacks in current research on this topic.

Zhu \etal \cite{zhu2019generating} introduce a novel method to elaborate adversarial examples for attacking well-trained face recognition models. 
Their approach involves applying makeup effects to facial images through two GANs-based sub-networks: the Makeup Transfer Sub-network and Adversarial Attack Sub-network.
\cite{dong2019efficient} aims to investigate the robustness of current face recognition models in the decision-based black-box attack scenario.
Sharif \etal \cite{sharif2016accessorize} concentrates on the attack of facial biometric systems, which are extensively used for surveillance and access control. 
They introduce a new attack method that is both physically realizable and inconspicuous, enabling an attacker to circumvent identification or impersonate another individual. 
The authors of \cite{nguyen2020adversarial} investigate the possibility of performing real-time physical attacks on face recognition systems through the use of adversarial light projections.
In study \cite{xiao2021improving}, the researchers conduct a comprehensive evaluation of the robustness of face recognition models against adversarial attacks using patches in the black-box setting. 
In contrast to previous methods that rely on designing perturbations, Wei \etal \cite{wei2022adversarial} achieve physical attacks by manipulating the position and rotation angle of stickers pasted onto faces.
Paper\cite{wei2022simultaneously} addresses the importance of position and perturbation in adversarial attacks by proposing a novel method that optimizes both factors simultaneously. 
By doing so, they achieve a high attack success rate in the black-box setting. 
To comprehensively evaluate physical attacks against face recognition systems, \cite{yang2022controllable} introduce a framework that employs 3D-face modeling to simulate complex transformations of faces in the physical world, thus creating a digital counterpart of physical faces. 
This generic framework enables users to control various face variations and physical conditions, making it possible to conduct reproducible evaluations comprehensively.
In study \cite{zheng2023robust}, the authors investigate the adversarial robustness of face recognition systems against sticker-based physical attacks, aiming to gain a better understanding of the system's vulnerabilities.
To increase the imperceptibility of attacks, Lin \etal \cite{lin2022real} propose a physical adversarial attack using full-face makeup, as its presence on the human face is a common occurrence.
Singh \etal \cite{singh2022powerful} present a new smoothness loss and a patch-noise combo for the physical attack against face recognition systems.
\cite{yang2023towards} aims to devise a more dependable technique that can holistically assess the adversarial resilience of commercial face recognition systems from end to end.
To achieve this goal, they propose the design of Adversarial Textured 3D Meshes (AT3D) with the intricate topology on a human face. 
The AT3D can be 3D-printed and then worn by the attacker to evade the facial recognition defenses.

Finally, we summarize adversarial attacks against face recognition (\cite{sharif2016accessorize,zhu2019generating,dong2019efficient,nguyen2020adversarial,xiao2021improving,wei2022adversarial,wei2022simultaneously,yang2022controllable,lin2022real,singh2022powerful,zheng2023robust,yang2023towards,wenger2021backdoor}) in Table \ref{table:attack_fr}.

\begin{table*}[t!]
\scriptsize
    \renewcommand{\arraystretch}{1.25}
\caption{Adversarial attacks against face recognition.}
\label{table:attack_fr}
\centering
\setlength{\tabcolsep}{1.4mm}
\begin{threeparttable}
\begin{tabular*}{\hsize}{ccccccccccccc}
\hline\hline
\thead{\textbf{Methods}} & {\textbf{Venue}} & {\textbf{Tasks}} & {\textbf{Domain}} & {\textbf{Visibility}} & {\textbf{Semantics}} & {\textbf{Knowledge}} & {\textbf{Purpose}} & {\textbf{Universality}} & {\textbf{Position}} & {\textbf{Distortion}} & {\textbf{Form}} 
\\ \hline
Sharif et al.\cite{sharif2016accessorize} & CCS 2016    & Face Recognition & Physical & Visible & Meaningless & Both  & Both     & Universal & Both   & Data & Patch   \\
Zhu et al.\cite{zhu2019generating}                           & ICIP 2019   & Face Recognition & Digital  & Visible & Meaningful  & White & Targeted & Specific  & Local  & Data & Makeup  \\
Dong et al.\cite{dong2019efficient}                          & CVPR 2019   & Face Recognition & Digital  & Visible & Meaningless & Black & Both     & Specific  & Global & Data & Pixel   \\
ALPA\cite{nguyen2020adversarial}                                 & CVPR 2020   & Face Recognition & Physical & Visible & Meaningless & Both  & Both     & Universal & Local  & Data & Light   \\
GenAP\cite{xiao2021improving}                                & CVPR 2021   & Face Recognition & Both     & Visible & Meaningful  & Both  & Targeted & Universal & Local  & Data & Patch   \\
Adversarial Sticker\cite{wei2022adversarial}                  & TPAMI 2022  & Face Recognition & Both     & Visible & Meaningful  & Both  & Both     & Universal & Local  & Data & Sticker \\
Wei et al.\cite{wei2022simultaneously}                           & TPAMI 2022  & Face Recognition & Both     & Visible & Meaningless & Both  & Both     & Universal & Local  & Data & Patch   \\
Face3DAdv\cite{yang2022controllable}                            & arXiv 2022  & Face Recognition & Both     & Visible & Meaningful  & Both  & Both     & Universal & Local  & Data & Patch   \\
Lin et al.\cite{lin2022real}                           & ICASSP 2022 & Face Recognition & Both     & Visible & Meaningful  & White & Both     & Universal & Local  & Data & Makeup  \\
Singh el al.\cite{singh2022powerful}                         & WACV 2022   & Face Recognition & Both     & Visible & Meaningless & Both  & Targeted & Universal & Local  & Data & Patch   \\
CAA\cite{zheng2023robust}                                  & PR 2023     & Face Recognition & Physical & Visible & Meaningless & White & Both     & Universal & Local  & Data & Patch   \\
AT3D\cite{yang2023towards}                                 & CVPR 2023   & Face Recognition & Both     & Visible & Meaningful  & Black & Both     & Universal & Local  & Data & Patch   \\
Wenger et al.\cite{wenger2021backdoor}                        & CVPR 2021   & Face Recognition & Physical & Visible & Meaningful  & White & Targeted & Universal & Local  & Both & Trigger
\\                 
\hline\hline
\end{tabular*}
\end{threeparttable}
\end{table*}

\subsubsection{Others}

To investigate how adversarial examples affect deep product quantization networks (DPQNs), \cite{chen2022adversarial} propose to perturb the probability distribution of centroids assignments for a clean query to attack DPQNs-based retrieval systems.
\cite{chen2020universal} introduces the Attack on Attention (AoA) technique, which exploits the semantic property shared by DNNs. 
AoA demonstrates a marked increase in transferability when attention loss is employed in place of the traditional cross-entropy loss. 
Since AoA only modifies the loss function, it can be readily combined with other transferability-enhancing methods to achieve SOTA performance.
In study \cite{fu2022robust}, the authors develop novel techniques to generate robust unlearnable examples that are resistant to adversarial training.
For the first time, paper \cite{zhao2022clpa} introduces a clean-label approach for the poisoning availability attack, which reveals the intrinsic imperfection of classifiers.
Paper \cite{lovisotto2022give} highlights how the global reasoning of (scaled) dot-product attention can represent a significant vulnerability when faced with adversarial patch attacks.
The current study puts forth a novel interactive visual aid, DetectorDetective \cite{vellaichamy2022detectordetective}, which seeks to enhance users' comprehension of a model's behavior during the traversal of adversarial images through an object detector. 
The primary goal of DetectorDetective is to provide users with a deeper understanding of how object detectors respond to adversarial attacks.
Work \cite{mathov2022enhancing} represents an initial stride towards implementing physically viable adversarial attacks on visual tracking systems in real-life scenarios. 
Specifically, the authors accomplished this by developing a universal patch that serves to camouflage single-object trackers.
To attack depth estimation, Cheng \etal \cite{cheng2022physical} employ an optimization-based technique for systematically creating stealthy physical-object-oriented adversarial patches.
Research \cite{sava2022assessing} assesses the effects of the chosen transformations on the efficacy of physical adversarial attacks. 
Moreover, they measure attack performance under various scenarios, including multiple distances and angles.

Finally, we summarize other adversarial attacks (\cite{bai2020targeted,bai2021universal,bai2020adversarial,wei2022sparse,gu2022segpgd,xu2022rethinking,zheng2023u,zhu2023multi,wei2023efficient,wang2023targeted,chen2023rethinking,liu2020bias,wang2021universal,mopuri2018generalizable,tang2019adversarial,ma2021understanding,chen2022adversarial,cao2022stylefool,deng2022frequency,lovisotto2022give,eykholt2018robust,zeng2019adversarial,doan2022tnt,xie2017adversarial,wang2019advpattern,ding2021towards,cheng2022physical,han2022physical,zhong2022shadows,jia2022fooling,zhutpatch,wang2021dual,fu2022ad,gu2019badnets,sun2020towards,tu2020physically,cao2021invisible,zhu2021can,jin2022pla}) in Table \ref{table:others}.

\begin{table*}[t!]
\scriptsize
    \renewcommand{\arraystretch}{1.25}
\caption{Other adversarial attack methods.}
\label{table:others}
\centering
\setlength{\tabcolsep}{0.35mm}
\begin{threeparttable}
\begin{tabular*}{\hsize}{ccccccccccccc}
\hline\hline
\thead{\textbf{Methods}} & {\textbf{Venue}} & {\textbf{Tasks}} & {\textbf{Domain}} & {\textbf{Visibility}} & {\textbf{Semantics}} & {\textbf{Knowledge}} & {\textbf{Purpose}} & {\textbf{Universality}} & {\textbf{Position}} & {\textbf{Distortion}} & {\textbf{Form}} 
\\ \hline
DHTA\cite{bai2020targeted} & ECCV 2020            & Image Retrieval                                         & Digital  & Invisible         & Meaningless & White & Targeted   & Specific           & Global       & Data  & Pixel       \\
UAH\cite{bai2021universal}                         & ICML 2021            & Video Retrieval                                         & Digital  & Invisible         & Meaningless & Both  & Untargeted & Specific           & Global       & Data  & Pixel       \\
Bai et al.\cite{bai2020adversarial}                  & TPAMI 2021           & Person Re-Identification                                & Digital  & Invisible         & Meaningless & Both  & Both       & Specific           & Global       & Data  & Pixel       \\
SVA\cite{wei2022sparse}                         & IJCV 2022            & Video Recognition                                       & Digital  & Invisible         & Meaningless & Black & Both       & Specific           & Global       & Data  & Pixel       \\
SegPGD\cite{gu2022segpgd}                      & ECCV 2022            & SS                                   & Digital  & Invisible         & Meaningless & Both  & Untargeted & Specific           & Global       & Data  & Pixel       \\
Xu et al.\cite{xu2022rethinking}                   & TPAMI 2022           & Crowdsourced Ranking                                    & /        & /                 & /           & Black & Targeted   & /                  & /            & Data  & /           \\
ODFA\cite{zheng2023u}                        & IJCV 2022            & Image Retrieval                                         & Digital  & Invisible         & Meaningless & Both  & Untargeted & Specific           & Global       & Data  & Pixel       \\
Zhu et al.\cite{zhu2023multi}                  & TIFS 2023            & Multi-spectral PR                    & Digital  & Invisible         & Meaningless & Both  & Untargeted & Specific           & Global       & Data  & Pixel       \\
AstFocus\cite{wei2023efficient}                    & TPAMI 2023           & Video Recognition                                       & Digital  & Invisible         & Meaningless & Black & Both       & Specific           & Local        & Data  & Pixel       \\
TA-DCH\cite{wang2023targeted}                      & TCSVT 2023           & CHR                           & Digital  & Invisible         & Meaningless & Black & Targeted   & Specific           & Global       & Data  & Pixel       \\
CWA\cite{chen2023rethinking}                         & arXiv 2023           & IC/OD                   & Digital  & Both & Meaningless & Black & Untargeted & Both & Global/Local & Data  & Pixel/Patch \\
Liu et al.\cite{liu2020bias}                  & ECCV 2020            & IC/AC                & Both     & Visible           & Meaningless & Both  & Untargeted & Universal          & Local        & Data  & Patch       \\
Wang et al.\cite{wang2021universal}                 & TIP 2021             & IC/AC                & Both     & Visible           & Meaningless & Both  & Untargeted & Universal          & Local        & Data  & Patch       \\
GD-UAP\cite{mopuri2018generalizable}                      & TPAMI 2018           & OD/SS/DE & Digital  & Invisible         & Meaningless & Both  & Untargeted & Universal          & Global       & Data  & Pixel       \\
Tang et al.\cite{tang2019adversarial}                 & TPAMI 2019           & IC/Face Recognition                   & Digital  & Visible           & Meaningful  & White & Targeted   & Specific           & Global       & Data  & Pixel       \\
Ma et al.\cite{ma2021understanding}                   & PR 2020              & Medical IC                            & Digital  & Invisible         & Meaningless & White & Untargeted & Specific           & Global       & Data  & Pixel       \\
$S^2AS$\cite{chen2022adversarial}                        & TPAMI 2022           & Image Retrieval                                         & Digital  & Invisible         & Meaningless & Both  & Untargeted & Specific           & Global       & Data  & Pixel       \\
StyleFool\cite{cao2022stylefool}                   & arXiv                & Video Classification                                    & Digital  & Visible           & Meaningless & Black & Both       & Specific           & Global       & Data  & Style       \\
FTGAP\cite{deng2022frequency}                       & TIP 2022             & Texture Recognition                                     & Digital  & Invisible         & Meaningless & Black & /          & Specific           & Global       & Data  & Pixel       \\
Lovisotto et al.\cite{lovisotto2022give}            & CVPR 2022            & IC/OD                   & Digital  & Visible           & Meaningless & White & Untargeted & Specific           & Local        & Data  & Patch       \\
$RP_2$\cite{eykholt2018robust}                         & CVPR 2018            & TSR                                & Physical & Visible           & Meaningless & White & Targeted   & Universal          & Local        & Data  & Sticker     \\
Zeng et al.\cite{zeng2019adversarial}                 & CVPR 2019            & OC/VQA         & Physical & Visible           & Meaningless & White & Untargeted & Universal          & Local        & Data  & Pixel       \\
TnT\cite{doan2022tnt}                         & TIFS 2022            & IC/Face Recognition                   & Physical & Visible           & Meaningful  & Both  & Both       & Universal          & Local        & Data  & Patch       \\
DAG\cite{xie2017adversarial}                         & ICCV 2017            & OD/SS                  & Digital  & Invisible         & Meaningless & Both  & Untargeted & Specific           & Global       & Data  & Pixel       \\
advPattern\cite{wang2019advpattern}                  & ICCV 2019            & Person Re-Identification                                & Physical & Visible           & Meaningless & White & Both       & Universal          & Local        & Data  & Patch       \\
MTD\cite{ding2021towards}                         & AAAI 2021            & OT                                         & Physical & Visible           & Meaningless & White & /          & Universal          & Local        & Data  & Patch       \\
Cheng et al.\cite{cheng2022physical}                & ECCV 2022            & AD/MDE           & Physical & Visible           & Meaningless & White & /          & Universal          & Local        & Data  & Patch       \\
Han et al.\cite{han2022physical}                  & ACM MM 2022          & AD/Lane Detection                       & Physical & Visible           & Meaningful  & White & /          & Universal          & Local        & Model & Trigger     \\
Zhong et al.\cite{zhong2022shadows}                & CVPR 2022            & TSR                                & Both     & Visible           & /           & Black & Untargeted & Universal          & Local        & Data  & Light       \\
Jia et al.\cite{jia2022fooling}                  & arXiv                & TSR                                & Physical & Visible           & Meaningless & Black & Both       & Universal          & Local        & Data  & Patch       \\
Tpatch\cite{zhutpatch}                      & /                    & IC/OD                   & Physical & Visible           & Meaningful  & Both  & Both       & Universal          & Local        & Data  & Patch       \\
DAS\cite{wang2021dual}                         & CVPR 2021            & IC/OD                   & Both     & Visible           & Meaningful  & Both  & Untargeted & Universal          & Local        & Data  & Patch       \\
$Ad^2Attack$\cite{fu2022ad}                   & ICRA 2022            & RS/OT                          & Digital  & Invisible         & Meaningless & White & Untargeted & Specific           & Global       & Data  & Pixel       \\
BadNets\cite{gu2019badnets}                     & IEEE Access 2019     & IC/TSR           & Both     & Visible           & Both        & White & Targeted   & Universal          & Local        & Both  & Trigger     \\
Sun et al.\cite{sun2020towards}                  & USENIX Security 2020 & AD/LiDAR OD               & Physical & /                 & /           & Black & /          & /                  & /            & Data  & /           \\
Tu el al.\cite{tu2020physically}                   & CVPR 2020            & AD/LiDAR OD               & Physical & Visible           & Meaningless & Both  & Untargeted & Universal          & Local        & Data  & 3D Object   \\
MSF-ADV\cite{cao2021invisible}                     & S\&P 2021            & AD/LiDAR\&Camera OD       & Physical & Visible           & Meaningless & White & /          & Universal          & Local        & Data  & 3D Object   \\
Zhu et al.\cite{zhu2021can}                  & CCS 2021             & AD/LiDAR OD               & Physical & Visible           & Meaningful  & Black & Untargeted & Universal          & Local        & Data  & Location    \\
PLA-LiDAR\cite{jin2022pla}                   & S\&P 2023            & AD/LiDAR OD               & Physical & /                 & Meaningless & White & Both       & Universal          & /            & Data  & Laser
\\                 
\hline\hline
\end{tabular*}
    \begin{tablenotes}
        \footnotesize      
        \item "AC", "AD", "CHR", "DE", "IC", "PR", "MDE", "OC", "OD", "OT", "SS", "TSR", and "VQA" represent Automatic Check-out, Autonomous Driving, Depth Estimation, Cross-modal Hashing Retrieval, Image Classification, Palmprint Recognition, Monocular Depth Estimation, Object Classification, Object Detection, Object Tracking, Semantic Segmentation, Traffic Sign Recognition, and Visual Question Answering, respectively.
        \item "/"  represents "Not applicable".        
    \end{tablenotes}
\end{threeparttable}
\end{table*}

\subsection{Survey of Robustness in RS}

In the subsequent subsections, we undertake a meticulous appraisal of adversarial attacks as they relate to RS, with a particular focus on tasks such as image classification, object detection, and additional relevant applications. 
Our objective is to provide a systematic and exhaustive analysis of the current literature, thereby fostering a deeper understanding of the principles, techniques, and ramifications of adversarial attacks in the context of RS research.

\subsubsection{Image Classification}

The majority of attacks against RS imagery classifiers stem from the field of CV, thus most of the existing research focuses on digital attacks.
Czaja \etal \cite{czaja2018adversarial} first considers attacks against machine learning algorithms used in RS applications.
Specifically, they present a new study of adversarial examples in satellite image classification problems.
In \cite{chen2020attack}, the authors investigate the properties of adversarial examples in RSI scene classification. 
To this end, they create several scenarios by employing two popular attack algorithms, \ie FGSM and BIM are trained on various RSI benchmark datasets to fool DNNs.
The authors of \cite{xu2020assessing} perform a systematic analysis of the potential threat posed by adversarial examples to DNNs used for RS scene classification. 
They conduct both targeted and untargeted attacks to generate subtle adversarial perturbations that are imperceptible to human observers but can easily deceive DNNs-based models.
Paper \cite{du2021adversarial} proposes a UNet-based \cite{ronneberger2015u} GAN to enhance the optimizing efficiency and attack efficacy of the generated adversarial examples for Synthetic Aperture Radar Automatic Target Recognition (SAR-ATR) models.
\cite{chen2021empirical} aims to provide a thorough evaluation of the effects of adversarial examples on RSI classification. 
Technically, eight of the most advanced classification DNNs are tested on six RSI benchmarks. 
These data sets consist of both optical and synthetic-aperture radar (SAR) images with varying spectral and spatial resolutions.
The study \cite{burnel2021generating} introduces a novel approach for generating adversarial examples to fool RSI classifiers in black-box conditions by utilizing a variant of the Wasserstein generative adversarial network.
To enhance the success rate of adversarial attacks against scene classification, Jiang \etal \cite{jiang2021project} propose the use of the projected gradient descent method to create adversarial RSIs.
In article \cite{tian2021adversarial}, the authors analyze adversarial attacks against DL-based unmanned aerial vehicles (UAVs) and propose two novel adversarial attack methods against regression models utilized in UAVs.
\cite{peng2022empirical} presents a fully black-box universal attack (FBUA) framework for creating a single universal adversarial perturbation against SAR target recognition that can be used against a wide range of DNN architectures and a large percentage of target images.
Two variants of universal adversarial examples, called targeted universal adversarial examples and source-targeted universal adversarial examples, are proposed in work \cite{bai2022targeted}.
The proposed methods aim to extend universal adversarial perturbations to perform the targeted attack.
Xu \etal \cite{xu2022universal} present a comprehensive analysis of universal adversarial examples in RS data, without any prior knowledge of the target model. 
Furthermore, the authors introduce Mixup-Attack, a novel black-box adversarial attack method, and its simpler variant Mixcut-Attack, for RS data. 
The authors of \cite{drager2022backdoor} present a comprehensive investigation of backdoor attacks on RS data. 
Both scene classification and semantic segmentation tasks are considered, and systematic analysis is provided.
A novel approach called speckle-variant attack (SVA) is devised by Peng \etal \cite{peng2022speckle}.
The SVA consists of two major modules: an iterative gradient-based perturbation generator and a target region extractor.
\cite{wang2022universal} proposes a novel method to explore the basic characteristics of universal adversarial perturbations (UAPs) of RSIs. 
The method involves combining an encoder-decoder network with an attention mechanism to generate UAPs of RSIs.
Qin \etal \cite{qin2022universal} present a novel universal adversarial attack method for CNN-SAR image classification. 
The proposed approach aims to differentiate the target distribution by utilizing a feature dictionary model, without any prior knowledge of the classifier.

Finally, we summarize adversarial attacks against image classification in RS (\cite{czaja2018adversarial,chen2020attack,xu2020assessing,du2021adversarialSAR,du2021adversarial,chen2021empirical,burnel2021generating,jiang2021project,tian2021adversarial,peng2022empirical,bai2022targeted,xu2022universal,drager2022backdoor,peng2022speckle,wang2022universal,qin2022universal}) in Table \ref{table:classification_rs}.

\begin{table*}[t!]
\scriptsize
    \renewcommand{\arraystretch}{1.25}
\caption{Adversarial attacks against image classification in RS.}
\label{table:classification_rs}
\centering
\setlength{\tabcolsep}{1.05mm}
\begin{threeparttable}
\begin{tabular*}{\hsize}{ccccccccccccc}
\hline\hline
\thead{\textbf{Methods}} & {\textbf{Venue}} & {\textbf{Tasks}} & {\textbf{Domain}} & {\textbf{Visibility}} & {\textbf{Semantics}} & {\textbf{Knowledge}} & {\textbf{Purpose}} & {\textbf{Universality}} & {\textbf{Position}} & {\textbf{Distortion}} & {\textbf{Form}} 
\\ \hline
Czaja et al.\cite{czaja2018adversarial} & SIGSPATIAL 2018  & RS/Image Classification & Digital & Invisible & Meaningless & White & Both       & Specific  & Global & Data & Pixel   \\
Chen et al.\cite{chen2020attack}                         & IEEE Access 2020 & RS/Image Classification & Digital & Invisible & Meaningless & White & Untargeted & Specific  & Global & Data & Pixel   \\
Xu et al.\cite{xu2020assessing}                           & TGRS 2020        & RS/Image Classification & Digital & Invisible & Meaningless & White & Both       & Specific  & Global & Data & Pixel   \\
Du et al.\cite{du2021adversarialSAR}                           & RS 2021          & RS/Image Classification & Digital & Invisible & Meaningless & White & Both       & Specific  & Global & Data & Pixel   \\
Du et al.\cite{du2021adversarial}                           & arXiv 2021       & RS/Image Classification & Digital & Visible   & Meaningless & White & Targeted   & Universal & Local  & Data & Patch   \\
Chen et al.\cite{chen2021empirical}                         & TGRS 2021        & RS/Image Classification & Digital & Invisible & Meaningless & White & Untargeted & Specific  & Global & Data & Pixel   \\
ARWGAN\cite{burnel2021generating}                              & TGRS 2021        & RS/Image Classification & Digital & Invisible & Meaningless & Black & Both       & Specific  & Global & Data & Pixel   \\
Jiang et al.\cite{jiang2021project}                        & SCN 2021         & RS/Image Classification & Digital & Invisible & Meaningless & White & Both       & Specific  & Global & Data & Pixel   \\
Tian et al.\cite{tian2021adversarial}                         & IoT-J 2021       & RS/Image Classification & Digital & Invisible & Meaningless & White & Both       & Specific  & Global & Data & Pixel   \\
FBUA\cite{peng2022empirical}                                & RS 2022          & RS/Image Classification & Digital & Invisible & Meaningless & Black & Both       & Universal & Global & Data & Pixel   \\
TUAE\cite{bai2022targeted}                                & RS 2022          & RS/Image Classification & Digital & Invisible & Meaningless & White & Targeted   & Specific  & Global & Data & Pixel   \\
Mixup-Attack\cite{xu2022universal}                        & TGRS 2022        & RS/Image Classification & Digital & Invisible & Meaningless & Black & Untargeted & Specific  & Global & Data & Pixel   \\
WABA\cite{drager2022backdoor}                                & arXiv 2022       & RS/Image Classification & Digital & Invisible & Meaningless & White & Untargeted & Universal & Global & Data & Trigger \\
SVA\cite{peng2022speckle}                                 & GRSL 2022        & RS/Image Classification & Digital & Invisible & Meaningless & Black & /          & Specific  & Global & Data & Pixel   \\
Wang et al.\cite{wang2022universal}                         & IEEE MMSP 2022   & RS/Image Classification & Digital & Invisible & Meaningless & White & Untargeted & Universal & Global & Data & Pixel   \\
Qin et al.\cite{qin2022universal}                          & IGARSS 2022      & RS/Image Classification & Digital & Invisible & Meaningless & White & Untargeted & Universal & Global & Data & Pixel
\\                 
\hline\hline
\end{tabular*}
    \begin{tablenotes}
        \footnotesize      
        \item "/"  represents "Not applicable".        
    \end{tablenotes}
\end{threeparttable}
\end{table*}

\subsubsection{Object Detection}
\label{Subsubsection2.3.2}

Similarly, the adversarial attack methods are divided into digital attacks and physical attacks according to the attacked domain.

\ding{172} \textbf{Digital attack}

The authors of \cite{den2020adversarial} first investigate the use of patch-based adversarial attacks in the context of unmanned aerial surveillance. 
Specifically, they explore the application of these attacks on large military assets by laying a patch on top of them, which camouflages them from automatic detectors analyzing the imagery.
\cite{lu2021scale} introduces a novel adversarial attack method called Patch-Noobj, which is designed to address the problem of large-scale variation in aircraft in RS imagery. 
PatchNoobj is a universal adversarial method that can be used to attack aircraft of different sizes by adaptively scaling the width and height of the patch according to the size of the target aircraft.
Du \etal \cite{du2021adversarial} investigate the susceptibility of DL-based cloud detection systems to adversarial attacks. 
Specifically, they employ an optimization process to create an adversarial pattern that, when overlaid onto a cloudless scene, causes the DNNs to falsely detect clouds in the image.
In paper \cite{yuan2021generating}, the authors devise a novel approach for generating adversarial pan-sharpened images. 
To achieve this, a generative network is employed to generate the pan-sharpened images, followed by the application of shape and label loss to carry out the attack task.
In the paper \cite{van2022weaknesses}, the researchers investigate the effectiveness and limitations of adversarial camouflage in the context of overhead imagery.
Fu \etal \cite{fu2022ad} propose Ad2Attack, an Adaptive Adversarial Attack approach against UAV object tracking. 
Adversarial examples are generated online during the resampling of the search patch image, causing trackers to lose the target in the subsequent frames.
Tang \etal \cite{tang2023adversarial} propose a novel adversarial patch attack algorithm. 
In particular, unlike traditional approaches that rely on the final outputs of models, the proposed algorithm uses the intermediate outputs to optimize adversarial patches. 
The study \cite{rasol2023adaptive} introduces a novel defense mechanism based on adversarial patches that aim to disable the onboard object detection network of the LSST (Low-Slow-Small Target) recognition system by launching an adversarial attack.
\cite{wei2023adversarial} introduces a novel framework for generating adversarial pan-sharpened images. 
The proposed method employs a two-stream network to generate the pan-sharpened images and applies shape loss and label loss to carry out the attack task. 
To ensure the quality of the pan-sharpened images, a perceptual loss is utilized to balance spectral preservation and attacking performance.
Sun \etal \cite{sun2023threatening} concentrate on patch-based attacks (PAs) against optical RSIs and propose a Threatening PA without the scarification of the visual quality, dubbed TPA.

\ding{173} \textbf{Physical attack}

In work \cite{zhang2022adversarial}, the authors conduct a comprehensive analysis of the universal adversarial patch attack for multi-scale objects in the RS field.
Specifically, this study presents a novel adversarial attack method for object detection in RS data by optimizing the adversarial patch to attack as many objects as possible by formulating a joint optimization problem. 
Furthermore, it introduces a scale factor to generate a universal adversarial patch that can adapt to multi-scale objects, ensuring its validity in real-world scenarios.
Du \etal \cite{du2022physical} have developed new experiments and metrics to assess the effectiveness of physical adversarial attacks on object detectors in aerial scenes, in order to investigate the impact of physical dynamics.
In research \cite{lian2022benchmarking}, the authors propose an Adaptive Patch-based Physical Attack (AP-PA), which enables physically practicable attacks using malicious patches for both the white-box and black-box settings in real physical scenarios. 
In \cite{lian2023contextual}, Lian \etal made the inaugural effort to execute physical attacks in a contextual manner against aerial detection in the physical world.
Following their previous work, Lian \etal propose Contextual Background Attack (CBA) \cite{lian2023cba}, which can achieve high attack effectiveness and transferability in real-world scenarios, without the need to obscure the target objects.
Technically, they extract the saliency of the target of interest as a mask for the adversarial patches and optimize the pixels outside the mask area to closely cover the critical contextual background area for detection.
Additionally, the authors devised a novel training strategy, in which the patches are forced to be outside the targets during training.
As a consequence, the elaborate perturbations can successfully hide the protected objects both on and outside the adversarial patches from being recognized.
The objective of \cite{deng2023rust} is to create a natural-looking patch that has a small perturbation area. 
This patch can be used in optical RSIs to avoid detection by object detectors and remain imperceptible to human eyes.
Paper \cite{chen2023attacking} presents an approach to adversarially attack satellite RS detection using a patch-based method. 
The proposed method aims to achieve comparable attack effectiveness in the physical domain as that in the digital domain, without compromising the visual quality of the patch. 
To achieve this, the approach utilizes pairwise-distance loss to control the salience of the adversarial patch.

Finally, we summarize adversarial attacks against object detection in RS (\cite{den2020adversarial,lu2021scale,yuan2021generating,zhang2022adversarial,du2022physical,van2022weaknesses,lian2022benchmarking,tang2023adversarial,lian2023contextual,lian2023cba,deng2023rust,rasol2023adaptive,wei2023adversarial,sun2023threatening,chen2023attacking}) in Table \ref{table:detection_rs}.

\begin{table*}[t!]
\scriptsize
    \renewcommand{\arraystretch}{1.25}
\caption{Adversarial attacks against object detection in RS.}
\label{table:detection_rs}
\centering
\setlength{\tabcolsep}{1.1mm}
\begin{threeparttable}
\begin{tabular*}{\hsize}{ccccccccccccc}
\hline\hline
\thead{\textbf{Methods}} & {\textbf{Venue}} & {\textbf{Tasks}} & {\textbf{Domain}} & {\textbf{Visibility}} & {\textbf{Semantics}} & {\textbf{Knowledge}} & {\textbf{Purpose}} & {\textbf{Universality}} & {\textbf{Position}} & {\textbf{Distortion}} & {\textbf{Form}} 
\\ \hline
Adhikari et al.\cite{den2020adversarial} & AIMLDA 2020         & RS/Object Detection & Digital & Visible   & Meaningless & White & Untargeted & Universal & Local  & Data & Patch          \\
Lu et al.\cite{lu2021scale}                              & RS 2021             & RS/Object Detection & Digital & Visible   & Meaningless & Both  & Untargeted & Universal & Local  & Data & Patch          \\
APS\cite{yuan2021generating}                                    & ADVM 2021           & RS/Object Detection & Digital & Visible   & Meaningful  & White & Untargeted & Specific  & Global & Data & PS \\
Zhang et al.\cite{zhang2022adversarial}                           & RS 2022             & RS/Object Detection & Both    & Visible   & Meaningless & White & Untargeted & Universal & Local  & Data & Patch          \\
Du et al.\cite{du2022physical}                              & WACV 2022           & RS/Object Detection & Both    & Visible   & Meaningless & White & Untargeted & Universal & Local  & Data & Patch          \\
Etten\cite{van2022weaknesses}                                  & AIPR 2022          & RS/Object Detection & Digital & Visible   & Meaningless & White & /          & Universal & Local  & Data & Patch          \\
AP-PA\cite{lian2022benchmarking}                                  & TGRS 2022           & RS/Object Detection & Both    & Visible   & Meaningless & Both  & Untargeted & Universal & Local  & Data & Patch          \\
Tang et al.\cite{tang2023adversarial}                            & Neurocomputing 2023 & RS/Object Detection & Digital & Visible   & Meaningless & White & Untargeted & Universal & Local  & Data & Patch          \\
Lian et al.\cite{lian2023contextual}                            & IGARSS 2023         & RS/Object Detection & Both    & Visible   & Meaningless & Both  & Untargeted & Universal & Local  & Data & Patch          \\
CBA\cite{lian2023cba}                                    & TGRS 2023           & RS/Object Detection & Both    & Visible   & Meaningless & Both  & Untargeted & Universal & Local  & Data & Patch          \\
RSP\cite{deng2023rust}                                    & RS 2023             & RS/Object Detection & Both    & Visible   & Meaningful  & White & Untargeted & Universal & Global & Data & Style          \\
Rasol et al.\cite{rasol2023adaptive}                           & RS 2023             & RS/Object Detection & Digital & Visible   & Meaningless & White & Untargeted & Universal & Local  & Data & Patch          \\
APA\cite{wei2023adversarial}                                    & PR 2023             & RS/Object Detection & Digital & Visible   & Meaningful  & White & Untargeted & Specific  & Global & Data & Pans \\
TPA\cite{sun2023threatening}                                    & TGRS 2023          & RS/Object Detection & Digital & Invisible & Meaningless & White & Untargeted & Specific  & Local  & Data & Patch          \\
Chen et al.\cite{chen2023attacking}                            & ICGNC 2023          & RS/Object Detection & Both    & Visible   & Meaningless & White & Untargeted & Universal & Local  & Data & Patch
\\                 
\hline\hline
\end{tabular*}
    \begin{tablenotes}
        \footnotesize      
        \item "PS" represents Pan-Sharpening.
        \item "/"  represents "Not applicable".        
    \end{tablenotes}
\end{threeparttable}
\end{table*}

\section{Benchmark}
\label{Section3}

\begin{figure}[!t]
  \centering
  \includegraphics[width=0.999\linewidth]{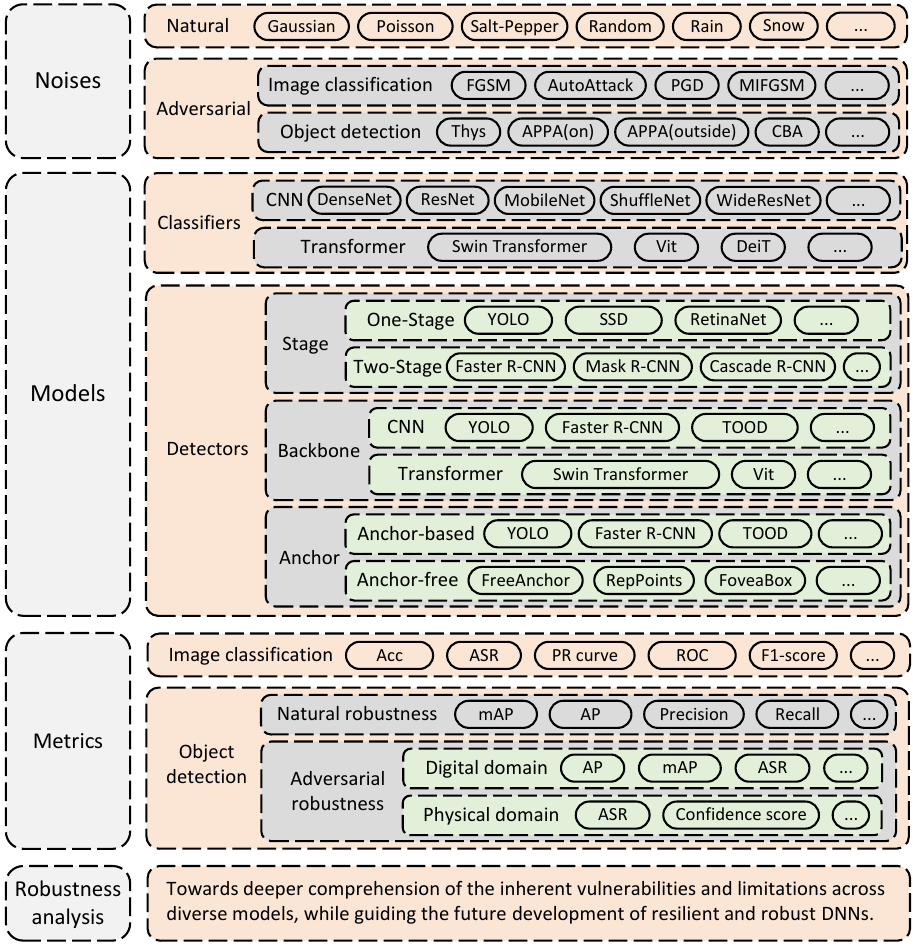}
  \caption{Overview of the benchmark assessing the robustness of image classification and object detection in RS imagery.}
  \label{fig:benchmark_overview}
\end{figure}

In this study, we introduce a comprehensive benchmark that assesses the robustness of image classification and object detection tasks in optical RSIs, as shown in Fig. \ref{fig:benchmark_overview}.
Specifically, we systematically investigate the robustness of typical DNNs-based image classifiers and object detectors against diverse natural and adversarial perturbations, which are fundamental elements of model robustness. 
Below, we give a detailed introduction to the benchmark on natural robustness and adversarial robustness in Sec. \ref{Subsection3.1} and Sec. \ref{Subsection3.2}, respectively.

\subsection{Natural Robustness}
\label{Subsection3.1}

\begin{figure}
  \centering
  \begin{subfigure}{0.24\linewidth}
    \includegraphics[width=1\linewidth]{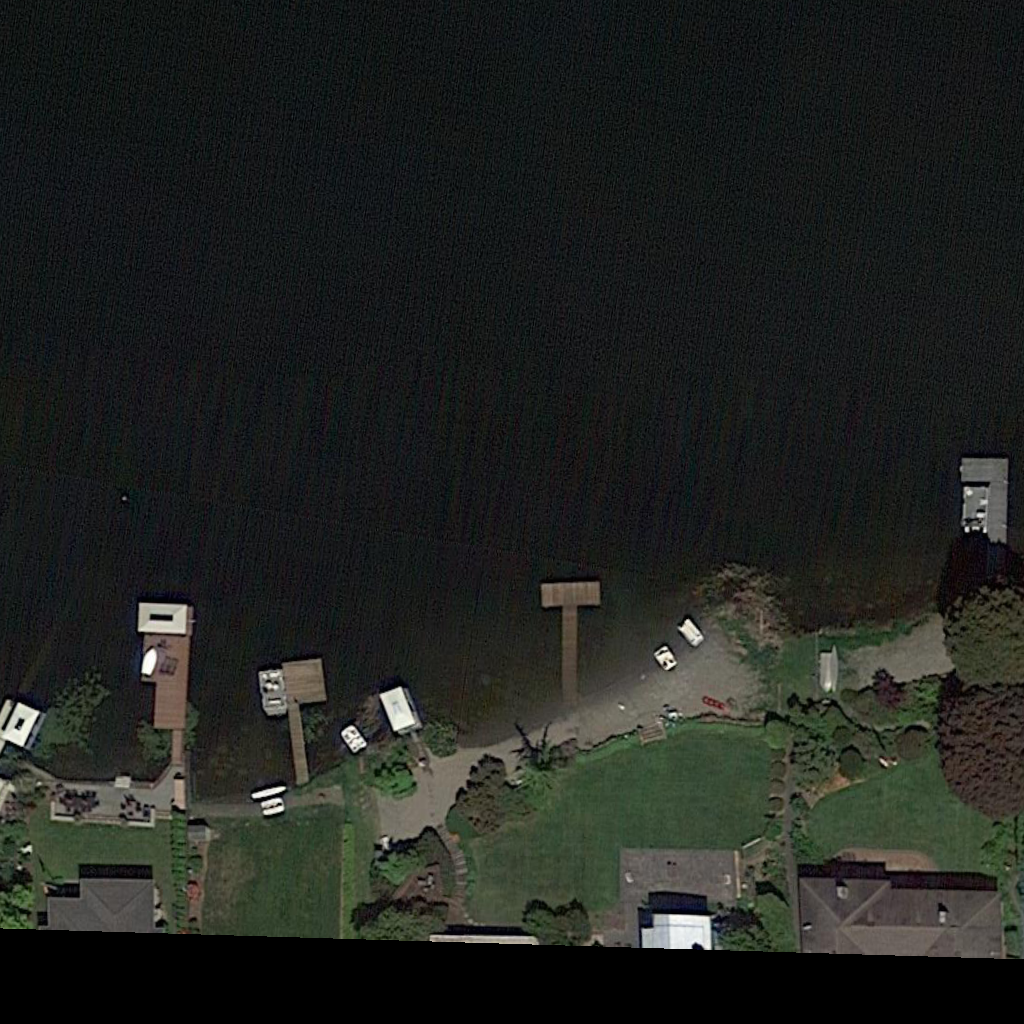}
    \caption{Clean}
  \end{subfigure}
  \begin{subfigure}{0.24\linewidth}
    \includegraphics[width=1\linewidth]{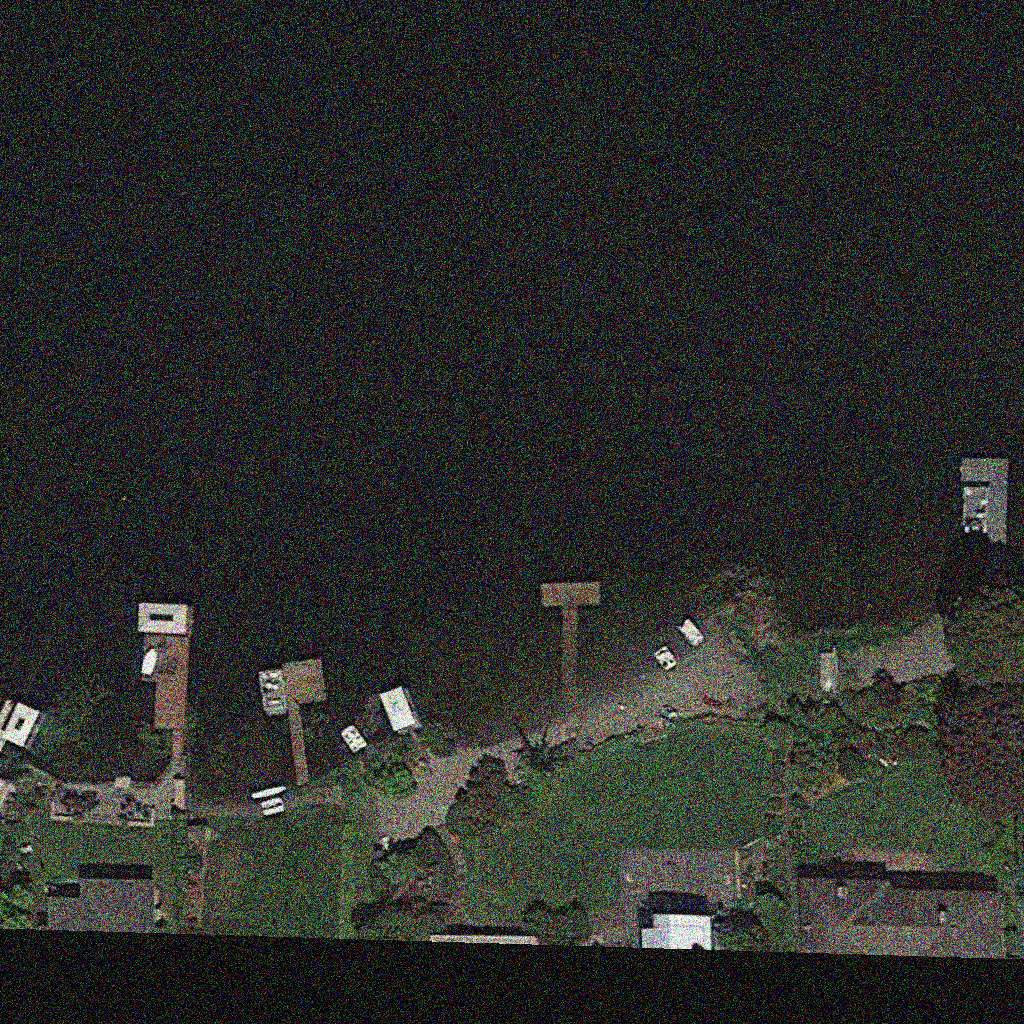}
    \caption{Gaussian}
  \end{subfigure}
  \begin{subfigure}{0.24\linewidth}
    \includegraphics[width=1\linewidth]{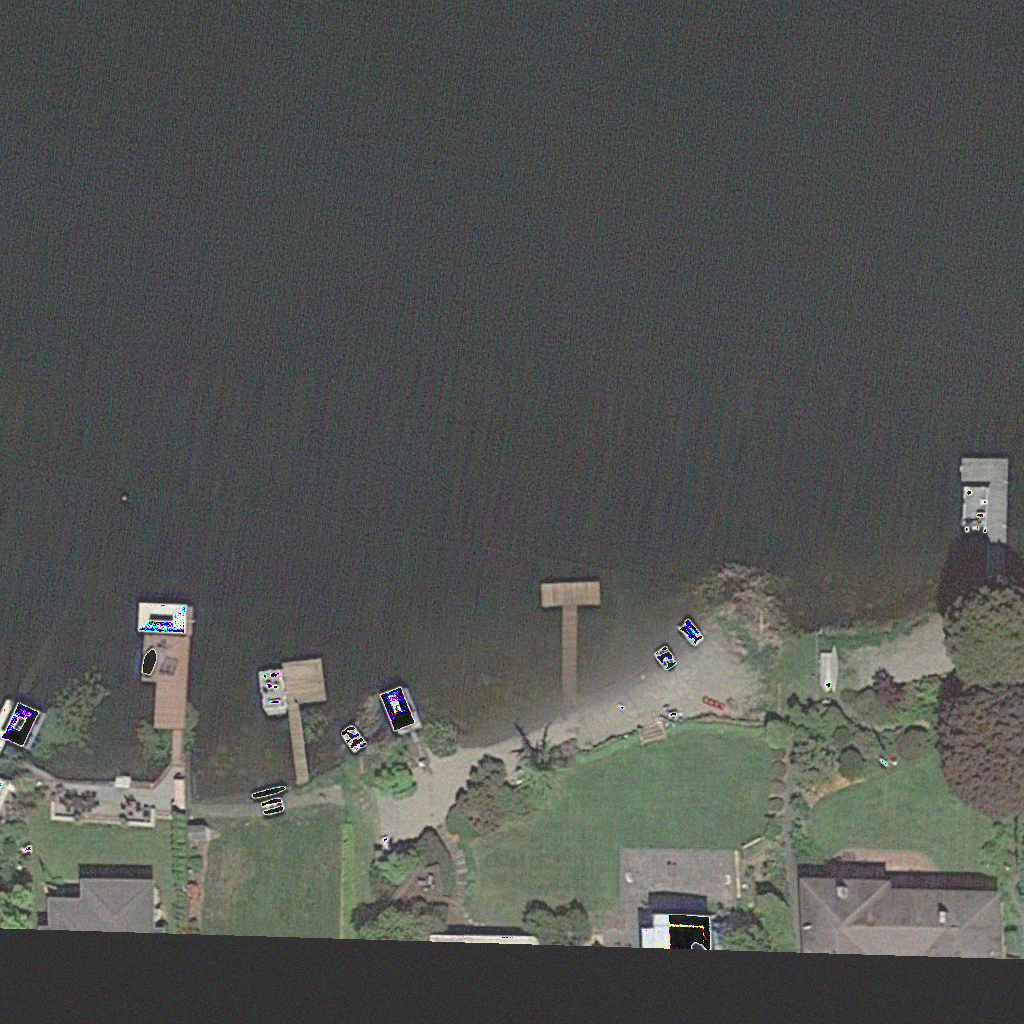}
    \caption{Poisson}
  \end{subfigure}
  \begin{subfigure}{0.24\linewidth}
    \includegraphics[width=1\linewidth]{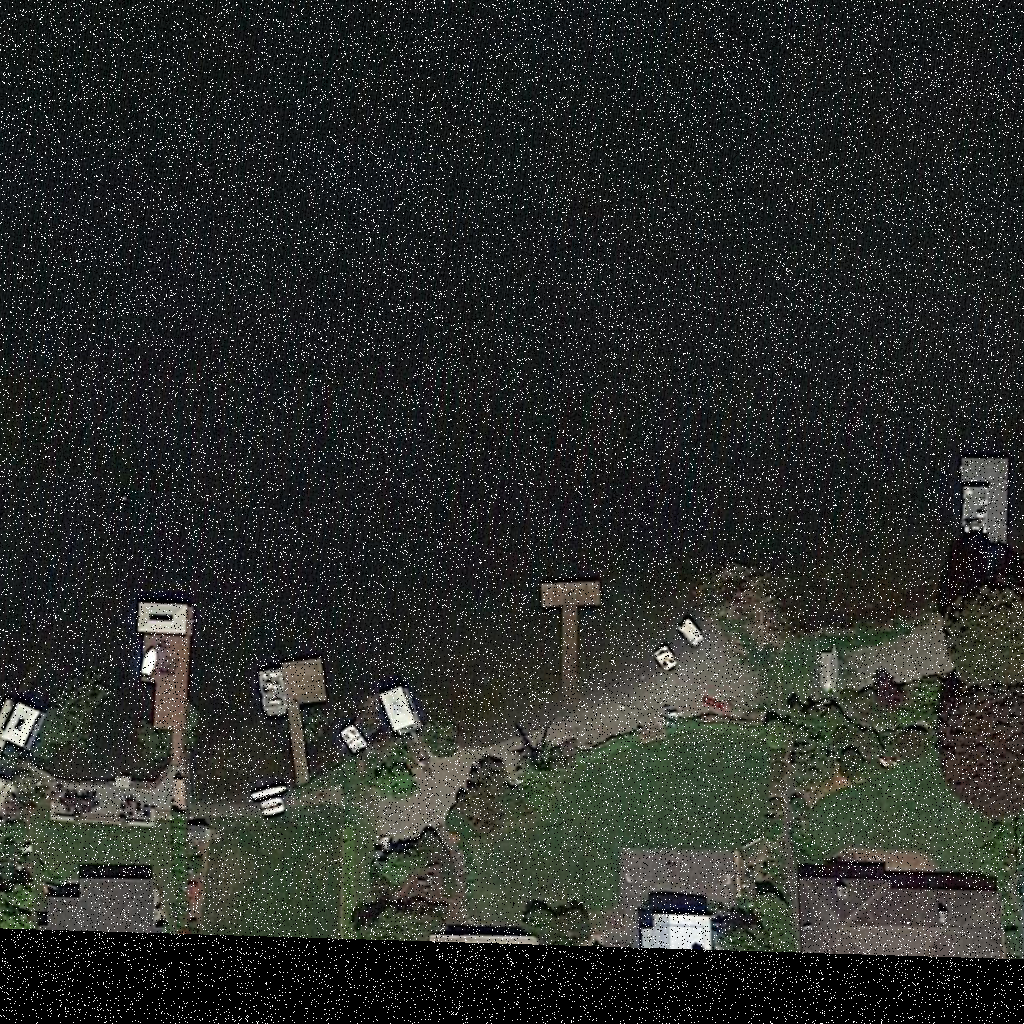}
    \caption{Salt-Pepper}
  \end{subfigure}
  \begin{subfigure}{0.24\linewidth}
    \includegraphics[width=1\linewidth]{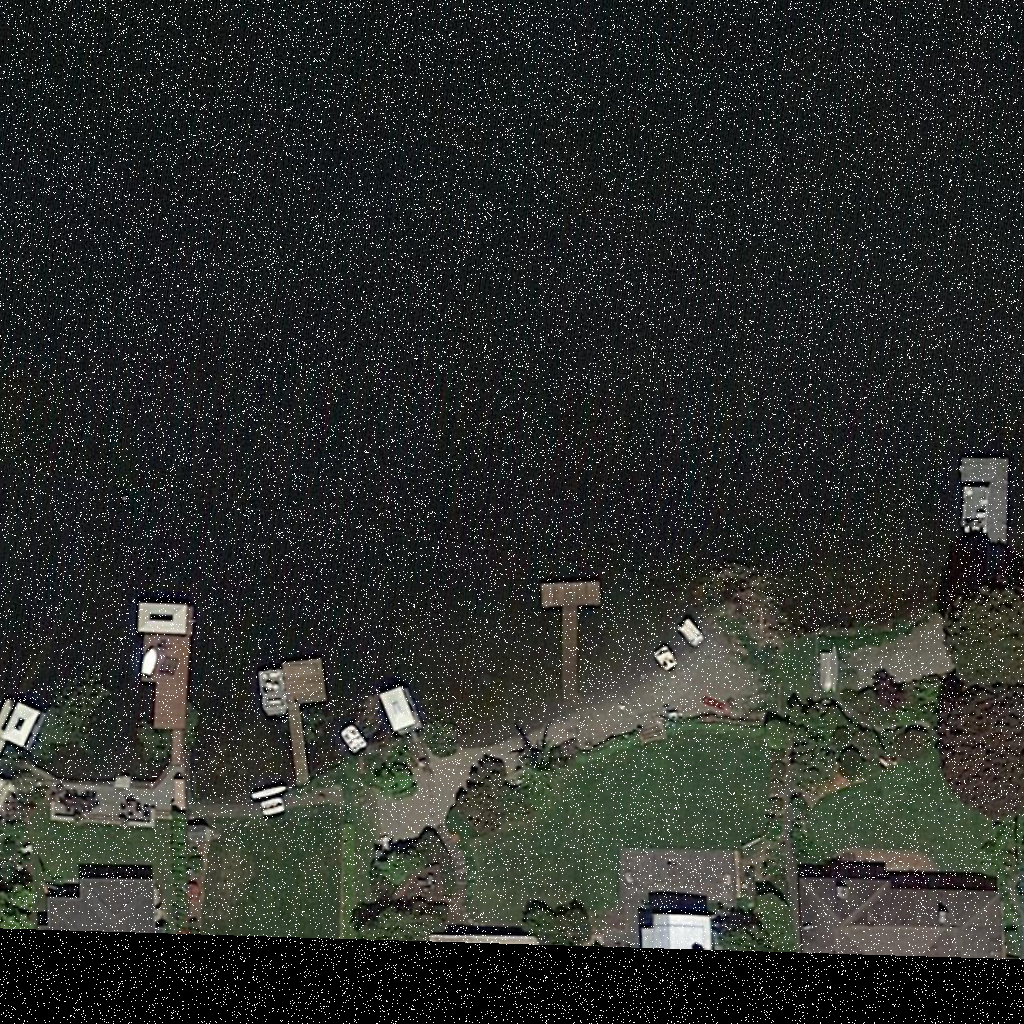}
    \caption{Random}
  \end{subfigure}
  \begin{subfigure}{0.24\linewidth}
    \includegraphics[width=1\linewidth]{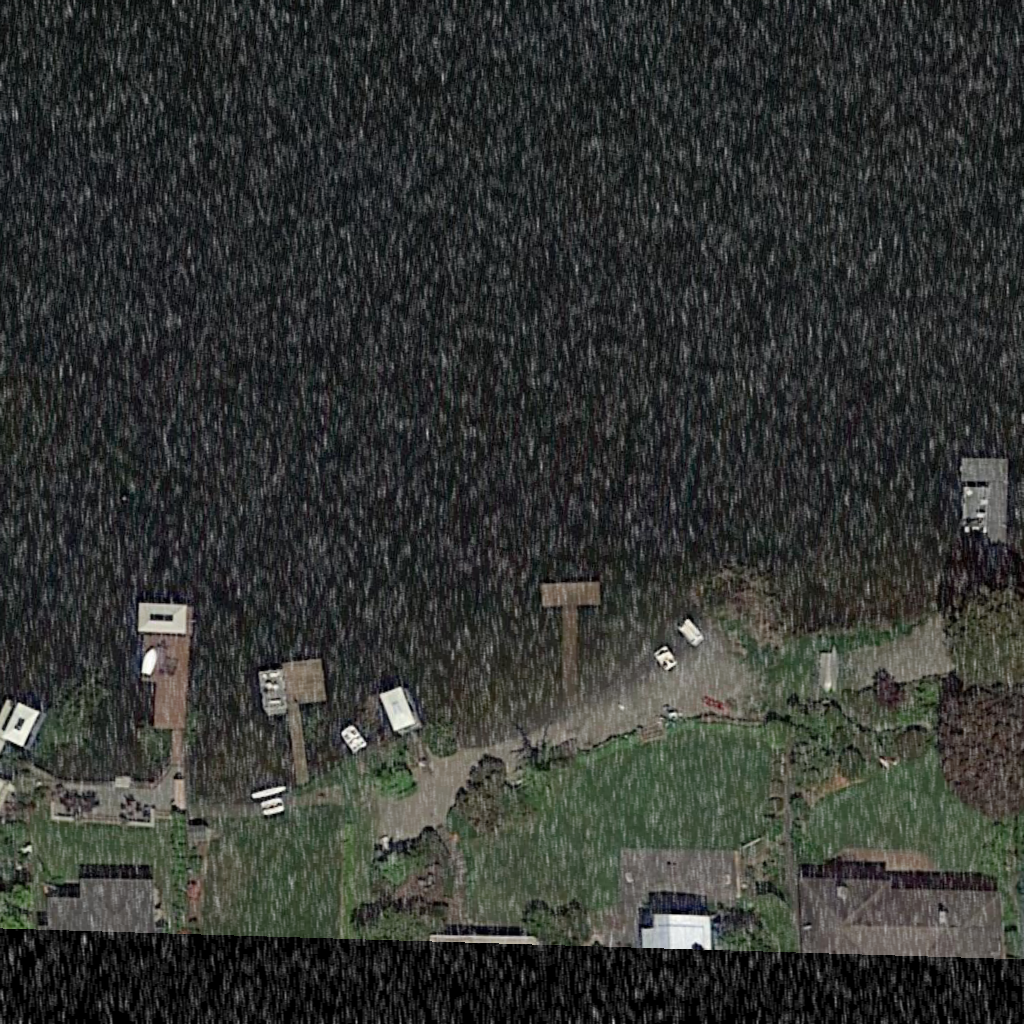}
    \caption{Rain}
  \end{subfigure}
  \begin{subfigure}{0.24\linewidth}
    \includegraphics[width=1\linewidth]{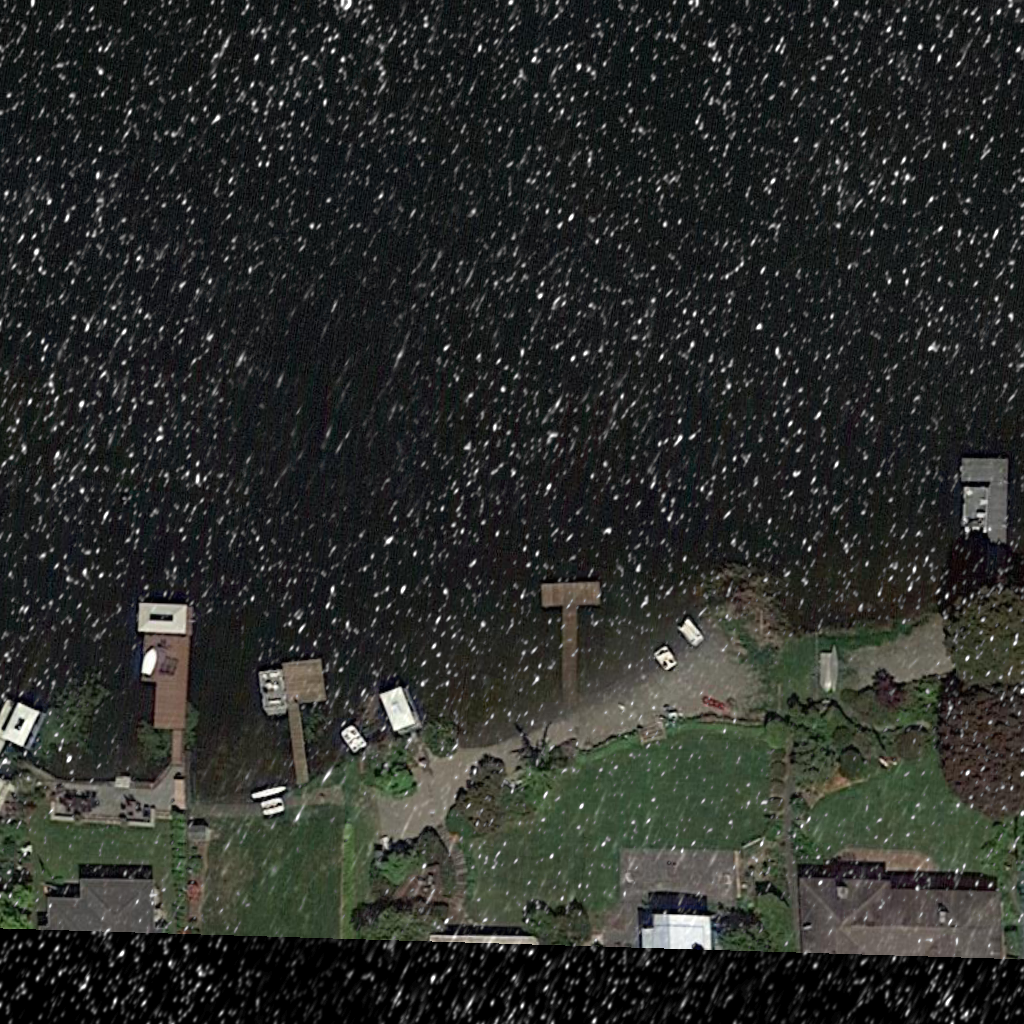}
    \caption{Snow}
  \end{subfigure}
  \begin{subfigure}{0.24\linewidth}
    \includegraphics[width=1\linewidth]{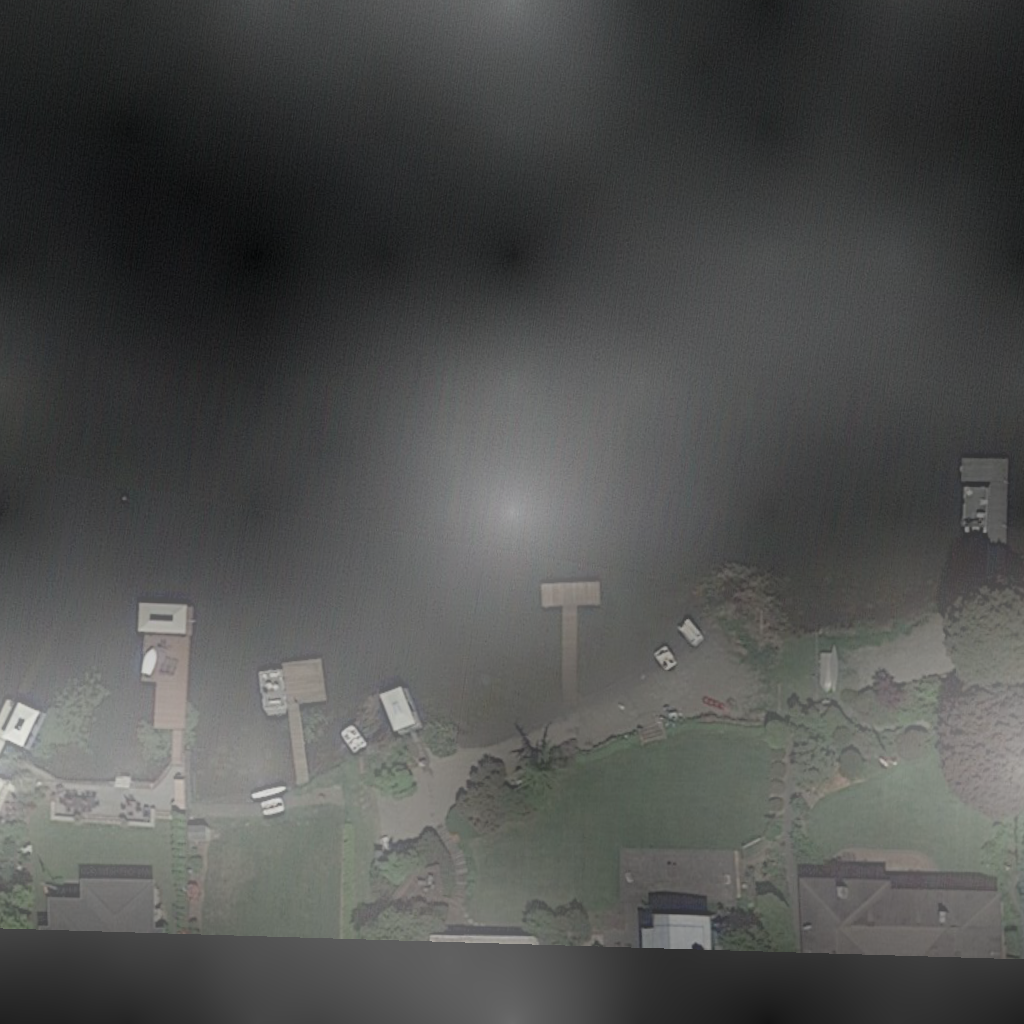}
    \caption{Fog}
  \end{subfigure}
  \caption{Images with different natural noises.}
  \label{fig:natural_noises}
\end{figure}

\begin{figure}
  \centering
  \begin{subfigure}{0.155\linewidth}
    \includegraphics[width=1\linewidth]{no-noise.png}
    \captionsetup{font=footnotesize}
    \caption{Clean}
  \end{subfigure}
  \begin{subfigure}{0.155\linewidth}
    \includegraphics[width=1\linewidth]{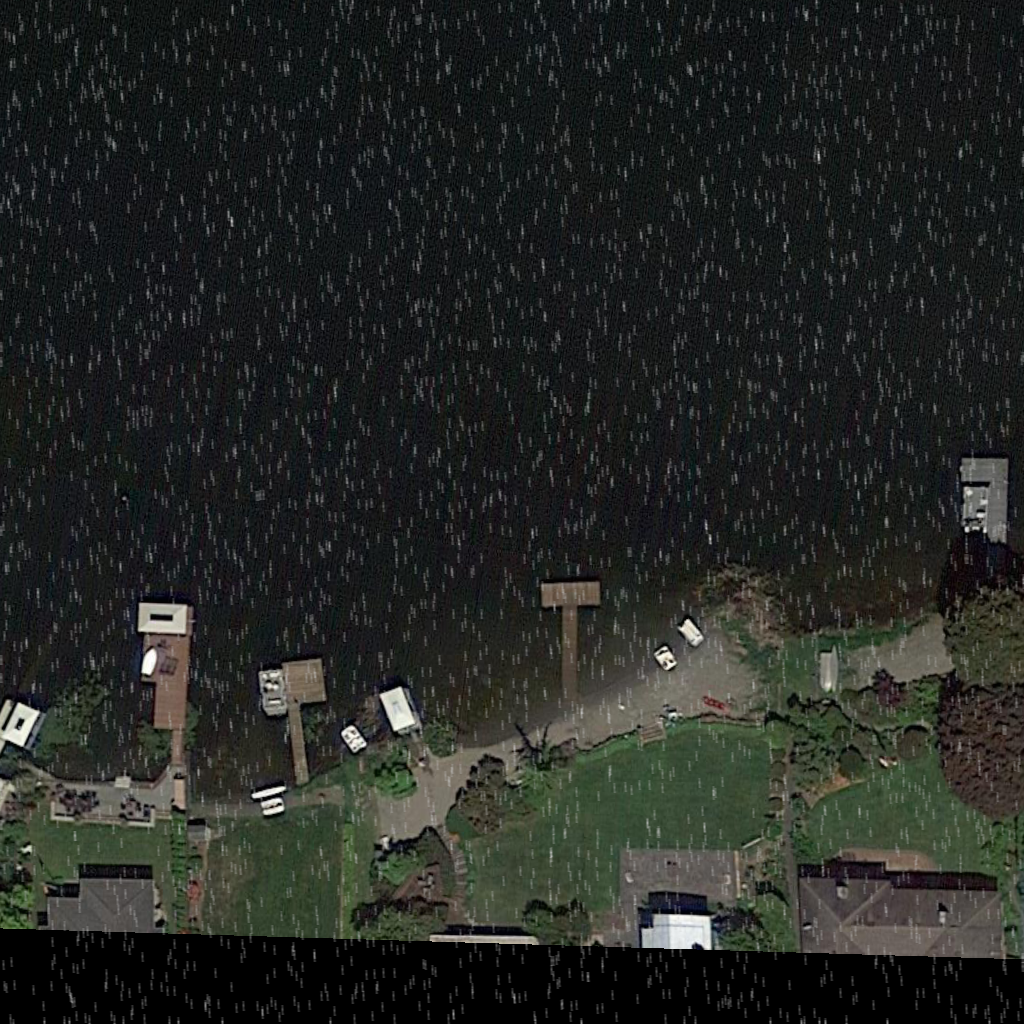}
    \captionsetup{font=footnotesize}
    \caption{Level1}
  \end{subfigure}
  \begin{subfigure}{0.155\linewidth}
    \includegraphics[width=1\linewidth]{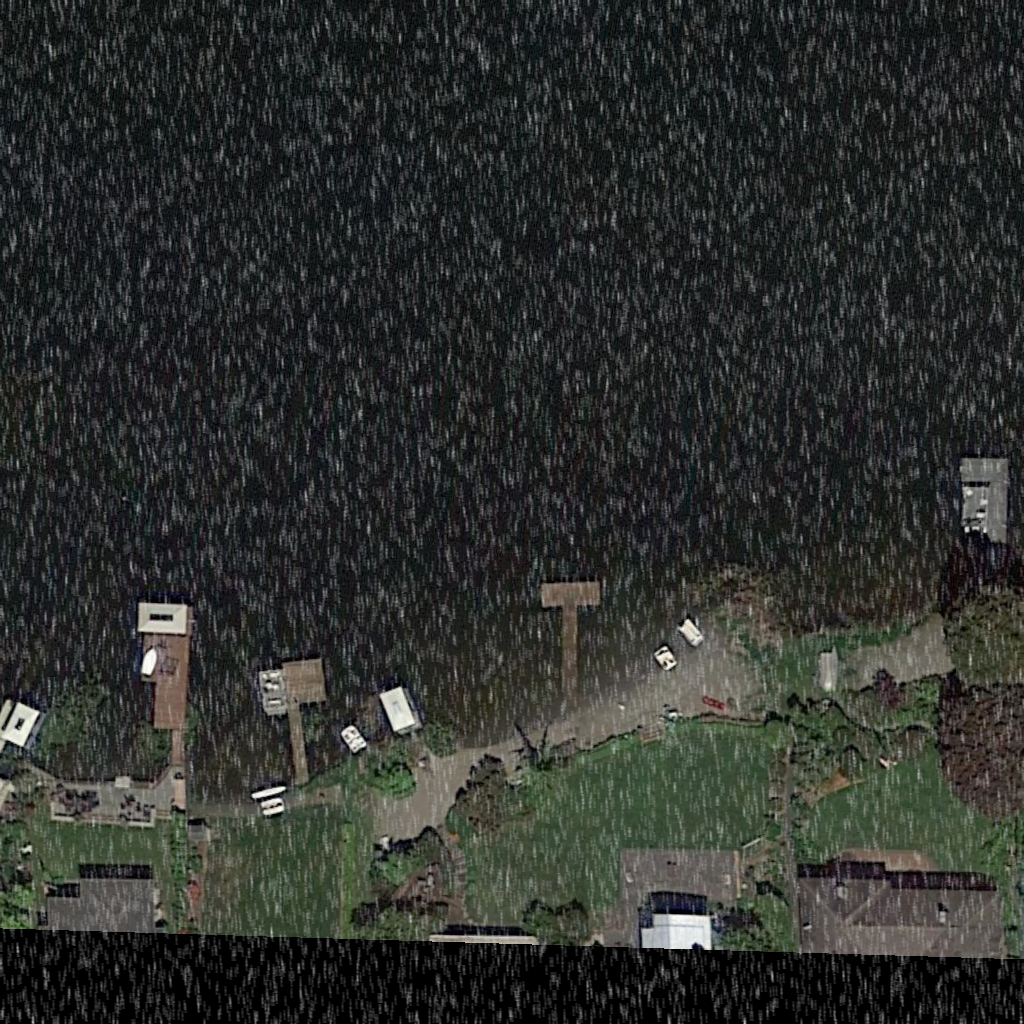}
    \captionsetup{font=footnotesize}
    \caption{Level2}
  \end{subfigure}
  \begin{subfigure}{0.155\linewidth}
    \includegraphics[width=1\linewidth]{rain3.png}
    \captionsetup{font=footnotesize}
    \caption{Level3}
  \end{subfigure}
  \begin{subfigure}{0.155\linewidth}
    \includegraphics[width=1\linewidth]{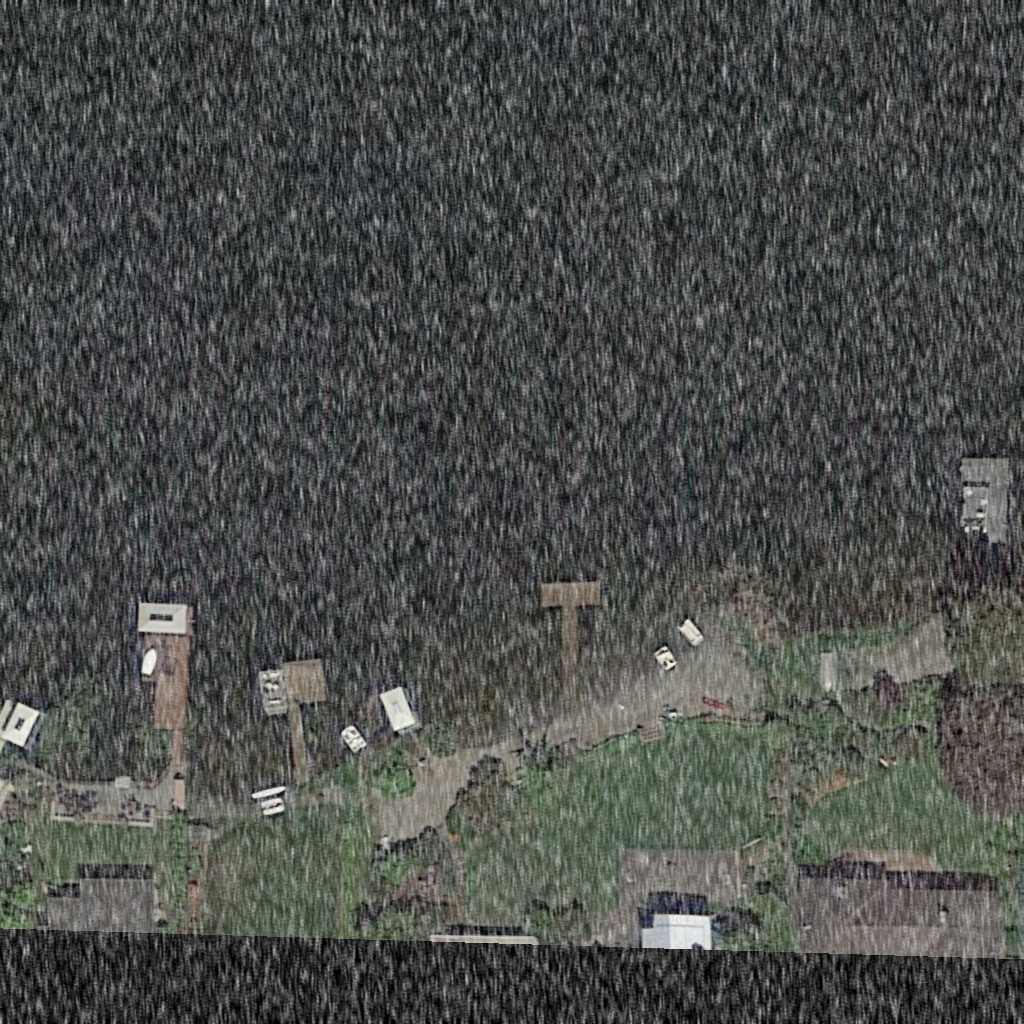}
    \captionsetup{font=footnotesize}
    \caption{Level4}
  \end{subfigure}
  \begin{subfigure}{0.155\linewidth}
    \includegraphics[width=1\linewidth]{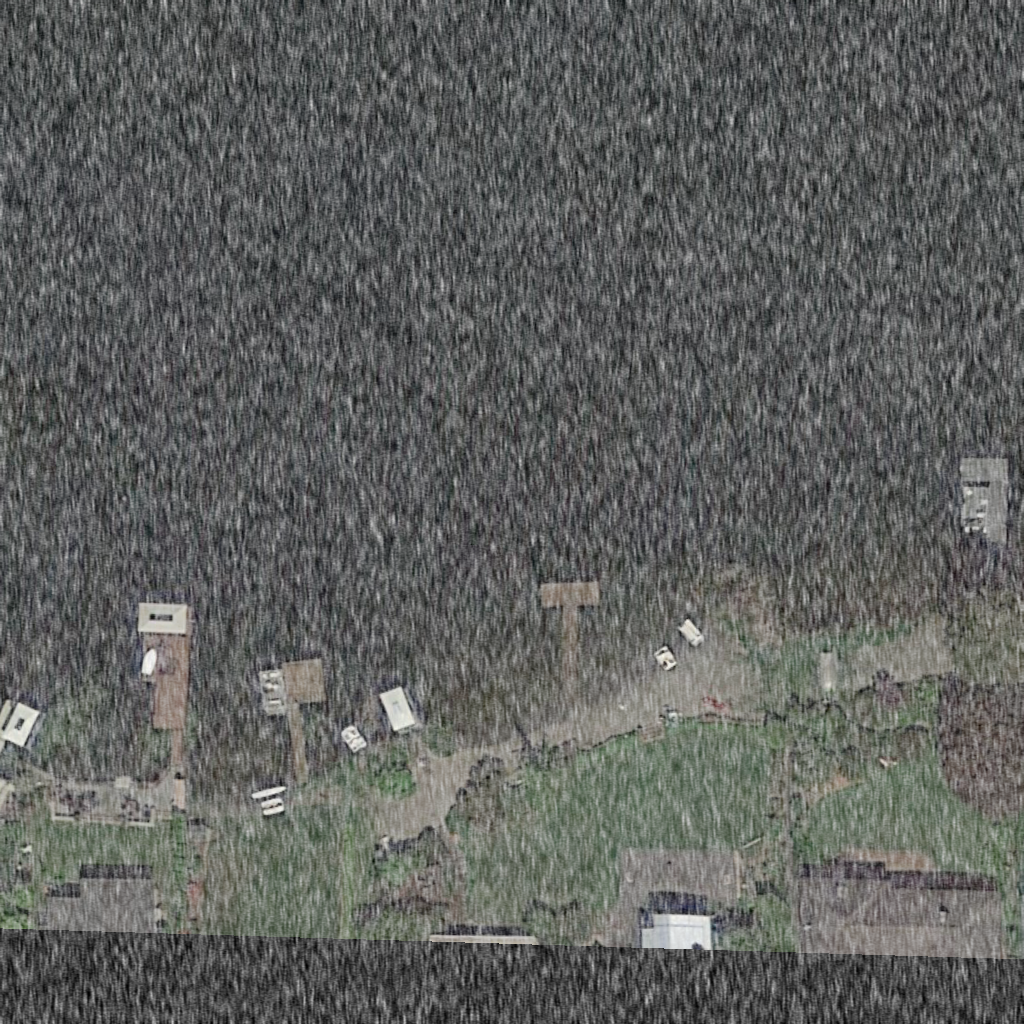}
    \captionsetup{font=footnotesize}
    \caption{Level5}
  \end{subfigure}
  \caption{Images with different levels of rain.}
  \label{fig:rain_different_levels}
\end{figure}

Various sources in the real world, such as weather fluctuations, sensor deterioration, and object deformations, generate natural noise that can be detrimental to DL models. 
These noises are inevitable, presenting a challenge in pursuing accurate and reliable artificial intelligence.
In order to undertake a thorough assessment of the inherent resilience of RSI classification and detection models in the face of varied and diverse forms of noise, it is necessary to adopt a rigorous and systematic benchmark. 
This benchmark should encompass a wide range of noise types and intensities, including those arising from natural environmental factors, sensor degradation, and varying degrees of image distortion. 
Such an all-encompassing evaluation holds the key to enhancing the practical viability of DNN-based models and to the development of more resilient and adaptable DNN architectures.
Although comprehensive benchmarks \cite{recht2019imagenet,beyer2020we,barbu2019objectnet,hendrycks2021natural,dongviewfool,hendrycks2019benchmarking,geirhos2018imagenet,wang2019learning} on natural noises have been established for the CV field, it is still lacking in the area of RS.
Consequently, we built the first benchmark and datasets on natural noises for RS tasks.
Specifically, we benchmark seven natural noises, including Gaussian noise (G), Poisson noise (P), salt-pepper noise (SP), random noise (RD), rain (R), snow (S), and fog (F), as shown in Fig. \ref{fig:natural_noises}.
Each noise is divided into five different intensities as shown in Fig. \ref{fig:rain_different_levels}.

In the following subsections, we mainly introduce the datasets, models, and metrics in our benchmark on natural robustness for image classification and object detection.

\subsubsection{Image Classification}
\label{Subsubsection3.1.1}

Benchmark on natural robustness for image classifiers:

\ding{172} \textbf{Datasets} 

\textbf{AID} \cite{xia2017aid} is a large-scale aerial image dataset for classification tasks, comprising sample images acquired from Google Earth imagery.
The AID dataset comprises 10000 aerial images from 30 different scene categories, including airport, bare land, baseball field, beach, bridge, center, church, commercial, dense residential, desert, farmland, forest, industrial, meadow, medium residential, mountain, park, parking, playground, pond, port, railway station, resort, river, school, sparse residential, square, stadium, storage tanks, and viaduct.
The AID is adopted to train aerial image classifiers.
To compute the overall accuracy, the ratios of training and testing sets are fixed at 50\% and 50\%, respectively.

\textbf{AID-NN} is introduced as a large-scale benchmark dataset to evaluate the natural robustness of image classification in aerial images, which is derived from AID by adding seven different \textbf{natural noises} in five levels, respectively. 
The rest information on AID-NN is the same as the original AID.

\ding{173} \textbf{Classifiers} 

In order to comprehensively evaluate and investigate the robustness trends across various DNN architectures for image classification, our benchmark endeavors to encompass a wide range of architectures as shown in Table \ref{table:classifiers}.
Regarding the CNNs, we select renowned and widely recognized classical network architectures, such as the ResNet series (including various versions of ResNet \cite{he2016deep}, ResNeXt \cite{xie2017aggregated}, and WRN \cite{zagoruyko2016wide}) and DenseNet \cite{huang2017densely}.
The lightweight ones including MobileNetV2 \cite{sandler2018mobilenetv2}, MobileNetV3 \cite{howard2019searching}, and ShuffleNetV2 \cite{ma2018shufflenet}.
As for the prevalent vision Transformer, Swin Transformer \cite{liu2021swin} and ViT \cite{dosovitskiy2020image} are adopted in this benchmark.

\begin{table}[t!]
\scriptsize
\caption{Classifiers of RS imagery.}
\label{table:classifiers}
\centering
\setlength{\tabcolsep}{1.5mm}
\begin{threeparttable}
\begin{tabular*}{\hsize}{ccc}
\hline\hline
\thead{\multirow{1}{*}{\textbf{Category}}} &{\multirow{1}{*}{\textbf{Architecture}}} &{\multirow{1}{*}{\textbf{Classifiers}}}

\\  \hline
    \multirow{8}{*}{\textbf{CNN}} 
                     &DenseNet &DenseNet-121,DenseNet-169,DenseNet-201\\\cline{2-3}
                     &MobileNet &MobileNetV2,MobileNetV3-S,MobileNetV3-L\\\cline{2-3}
    &{\multirow{2}{*}{ResNet}} &ResNet-18,ResNet-34,ResNet-50,\\
                      & &ResNet-101,ResNet-152\\\cline{2-3}
                     &ResNeXt &ResNeXt-101-32x8d,ResNeXt-50-32x4d\\\cline{2-3}
    &{\multirow{2}{*}{ShuffleNet}} &ShuffleNetV2-x0.5,ShuffleNetV2-x1.0,\\
                     & &ShuffleNetV2-x1.5,ShuffleNetV2-x2.0\\\cline{2-3}
                     &WideResNet &WRN-101-2,WRN-50-2
\\   \hline

\multirow{2}{*}{\textbf{Transformer}} 
                     &Swin Transformer &Swin-S,Swin-T\\\cline{2-3}
                     &Vit &Vit-B/16,Vit-B/32

\\                 
\hline\hline
\end{tabular*}
\end{threeparttable}
\end{table}

\ding{174} \textbf{Metric} 

\textbf{Acc:}
The mathematical formula to define the image classification evaluation index "Acc" (Accuracy) is as follows:
\begin{equation}
    \label{eq_acc}
    Acc = \frac{TP + TN}{TP + TN + FP + FN},
\end{equation}
where True Positive (TP) and True Negative (TN) are the numbers of correctly classified positive and negative samples, respectively;
False Positive (FP) and False Negative (FN) are the numbers of incorrectly classified positive and negative samples, respectively.

\subsubsection{Object Detection}
\label{Subsubsection3.1.2}

Benchmark on natural robustness for object detectors:

\ding{172} \textbf{Datasets} 

\textbf{DOTA} \cite{xia2018dota} is a large-scale benchmark dataset for object detection in aerial images, which contains 15 common categories, 2,806 images (image width range from 800 to 4,000), and 188,282 instances. 
The proportions of the training set, validation set, and testing set in DOTA are 1/2, 1/6, and 1/3, respectively.
DOTA is adopted to train aerial detectors after cropping\footnote{Image cropping tool: \url{https://github.com/CAPTAIN-WHU/DOTA_devkit}} the images as 1024$\times$1024.

\textbf{DOTA-NN} is introduced as a large-scale benchmark dataset to evaluate the natural robustness of object detection in aerial images, which is derived from DOTA (after cropping) by adding seven different \textbf{natural noises} in five levels, respectively. 
The rest information on DOTA-NN is the same as the original DOTA.

\ding{173} \textbf{Detectors} 

\begin{table}[t!]
\scriptsize
    \renewcommand{\arraystretch}{1.25}
\caption{Object detectors of RS imagery.}
\label{table:detectors}
\centering
\setlength{\tabcolsep}{0.1mm}
\begin{threeparttable}
\begin{tabular*}{\hsize}{ccc}
\hline\hline
\thead{\multirow{1}{*}{\textbf{Classification}}} &{\multirow{1}{*}{\textbf{Category}}} &{\multirow{1}{*}{\textbf{Detectors}}}

\\  \hline
    \multirow{3}{*}{\textbf{Stage}} 
                     &{\multirow{2}{*}{One-Stage}} &YOLOv2,YOLOv3,YOLOv5n,YOLOv5s,YOLOv5m,\\
                                                   & &YOLOv5l,YOLOv5x,SSD,RetinaNet,TOOD,ATSS\\
    \cline{2-3}    
                     &{\multirow{1}{*}{Two-Stage}} &Faster R-CNN,Cascade R-CNN,Mask R-CNN            
                     
\\   \hline
\multirow{5}{*}{\textbf{Backbone}} 
                     &{\multirow{4}{*}{CNN}} &YOLOv2,YOLOv3,YOLOv5n,YOLOv5s,YOLOv5m,\\
                                             & &YOLOv5l,YOLOv5x,SSD,Faster R-CNN,\\  
                                             & &Cascade R-CNN,RetinaNet,Mask R-CNN,FreeAnchor,\\
                                             & &FSAF,RepPoints,TOOD,ATSS,FoveaBox,VFNet\\
\cline{2-3}
                     &{\multirow{1}{*}{Transformer}} &Swin Transformer

\\    \hline
\multirow{4}{*}{\textbf{Anchor}} 
                     &{\multirow{3}{*}{Anchor-based}} &YOLOv2,YOLOv3,YOLOv5n,YOLOv5s,YOLOv5m,\\
                                                      & &YOLOv5l,YOLOv5x,SSD,RetinaNet,Mask R-CNN\\
                                                      & &Faster R-CNN,Cascade R-CNN,TOOD,ATSS,VFNet\\
\cline{2-3}
                     &{\multirow{1}{*}{Anchor-free}}  &FreeAnchor,FSAF,RepPoints,FoveaBox
\\            
\hline\hline
\end{tabular*}
\end{threeparttable}
\end{table}

Our benchmark includes an array of prominent object detectors as shown in Table \ref{table:detectors}, including YOLOv2 \cite{redmon2017yolo9000}, YOLOv3 \cite{redmon2018yolov3}, YOLOv5 \cite{jocher2020yolov5}, SSD \cite{liu2016ssd}, Faster R-CNN \cite{ren2015faster}, Swin Transformer \cite{liu2021swin}, Cascade R-CNN \cite{cai2019cascade}, RetinaNet \cite{lin2017focal}, Mask R-CNN \cite{he2017mask}, FoveaBox \cite{kong2020foveabox}, FreeAnchor \cite{zhang2019freeanchor}, FSAF \cite{zhu2019feature}, RepPoints \cite{yang2019reppoints}, TOOD \cite{feng2021tood}, ATSS \cite{zhang2020bridging}, and VarifocalNet (VFNet) \cite{zhang2021varifocalnet}.
Technically, our benchmark encompasses both one-stage (\eg YOLO \cite{redmon2017yolo9000,redmon2018yolov3,jocher2020yolov5}, SSD \cite{liu2016ssd}, \etc.) and two-stage detectors (\eg Faster R-CNN \cite{ren2015faster}, Cascade R-CNN \cite{cai2019cascade}, \etc.), as well as CNN-based (\eg YOLO \cite{redmon2017yolo9000,redmon2018yolov3,jocher2020yolov5}, Faster R-CNN \cite{ren2015faster}, \etc.) and Transformer-based (\eg Swin Transformer \cite{liu2021swin}) detectors. 
Furthermore, our benchmark also evaluates the performance of anchor-based detectors (\eg Cascade R-CNN \cite{cai2019cascade}, RetinaNet \cite{lin2017focal}, \etc.) and anchor-free detectors (\eg FreeAnchor \cite{zhang2019freeanchor}).

\ding{174} \textbf{Metric} 

\textbf{mAP:}
We use mean average precision (mAP) as the evaluation metric of object detection.
The mathematical definition formula of mAP is written as follows:
\begin{equation}
    \label{eq_mAP}
    mAP = \frac{1}{n} \sum^{n}_{i=1} AP_i,
\end{equation}
where $n$ is the number of object categories being detected, $AP_i$ is the average precision (AP) of the $i$-th category, which is calculated as:
\begin{equation}
    \label{eq_AP}
    AP_i = \int^{1}_{0} p_{interp}(r) dr,
\end{equation}
where $p_{\text{interp}}(r)$ is the interpolated precision at a certain recall level $r$, and is defined as:
\begin{equation}
    \label{eq_p_interp}
    p_{interp} = \max_{\tilde{r} \geq r} \tilde{p}(\tilde{r}).
\end{equation}
Here, $\tilde{p}(\tilde{r})$ is the precision at a certain recall level $\tilde{r}$, and the interpolation is done by taking the maximum precision value overall recall levels greater than or equal to $r$. 
The AP is calculated by averaging the precision values at all the recall levels at which there is a correct detection.
In practice, the mAP is typically calculated for a range of intersections over union (IoU) thresholds and then averaged over those thresholds. 
For example, mAP@[.50:.05:.95] means that the mAP is calculated by taking the mean of the AP scores at IoU thresholds of 0.5, 0.55, 0.6, ..., 0.95.
\subsection{Adversarial Robustness}
\label{Subsection3.2}

In this section, we mainly introduce the adversarial attacks, datasets, models, and metrics in our benchmark on adversarial robustness for image classification and object detection.

\subsubsection{Image Classification}
\label{Subsubsection3.2.1}

Benchmark on adversarial robustness for image classifiers:

\ding{172} \textbf{Attacks} 

In this benchmark, we evaluate adversarial robustness with 5 digital attacks in the same experimental settings, including Fast Gradient Sign Method (FGSM) \cite{ian2015explaining}, AutoAttack (AA) \cite{croce2020reliable}, Projected Gradient Descent (PGD) \cite{madry2018towards}, C\&W \cite{carlini2017towards}, Momentum Iterative FGSM (MIFGSM) \cite{dong2018boosting}. 
A detailed description of these attack methods is provided in Sec. \ref{Subsubsection2.2.1}.
Furthermore, we conduct the aforementioned attacks in both white-box and black-box conditions.

\ding{173} \textbf{Datasets} 

\textbf{AID} \cite{xia2017aid} dataset is used for crafting adversarial examples to test the adversarial robustness of aerial image classifiers.

\textbf{AID-AN} is introduced as a large-scale benchmark dataset to evaluate the adversarial robustness of the image classifier in RS, which is derived from AID by adding four different \textbf{adversarial noises}, respectively. 
The rest information on AID-AN is the same as the original AID.

\ding{174} \textbf{Models} and \ding{175} \textbf{Metric} are the same as the counterparts of natural robustness depicted in Sec. \ref{Subsubsection3.1.1}.

\subsubsection{Object Detection}
\label{Subsubsection3.2.2}

\begin{figure}
  \centering
    \begin{subfigure}{0.32\linewidth}
    \includegraphics[width=1\linewidth]{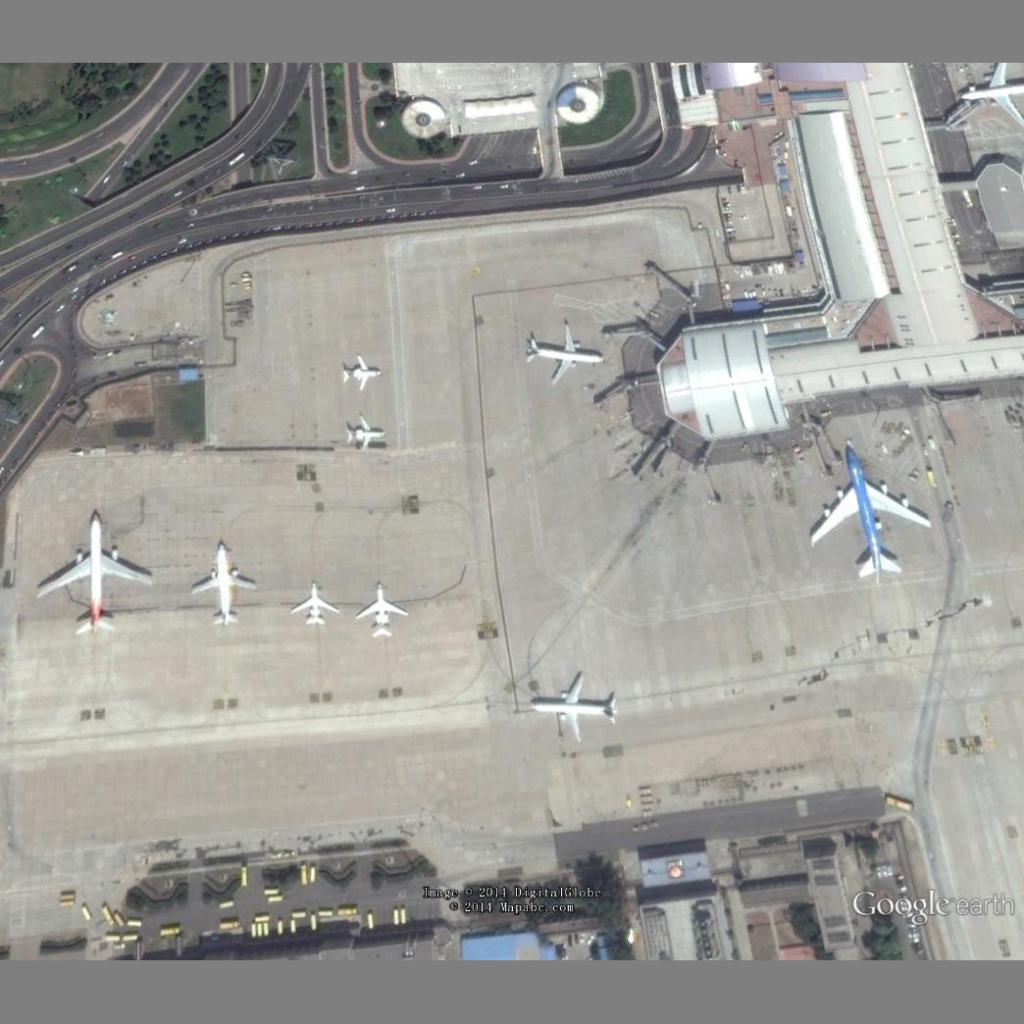}
    \captionsetup{font=small}
    \caption{Clean}
  \end{subfigure}
  \begin{subfigure}{0.32\linewidth}
    \includegraphics[width=1\linewidth]{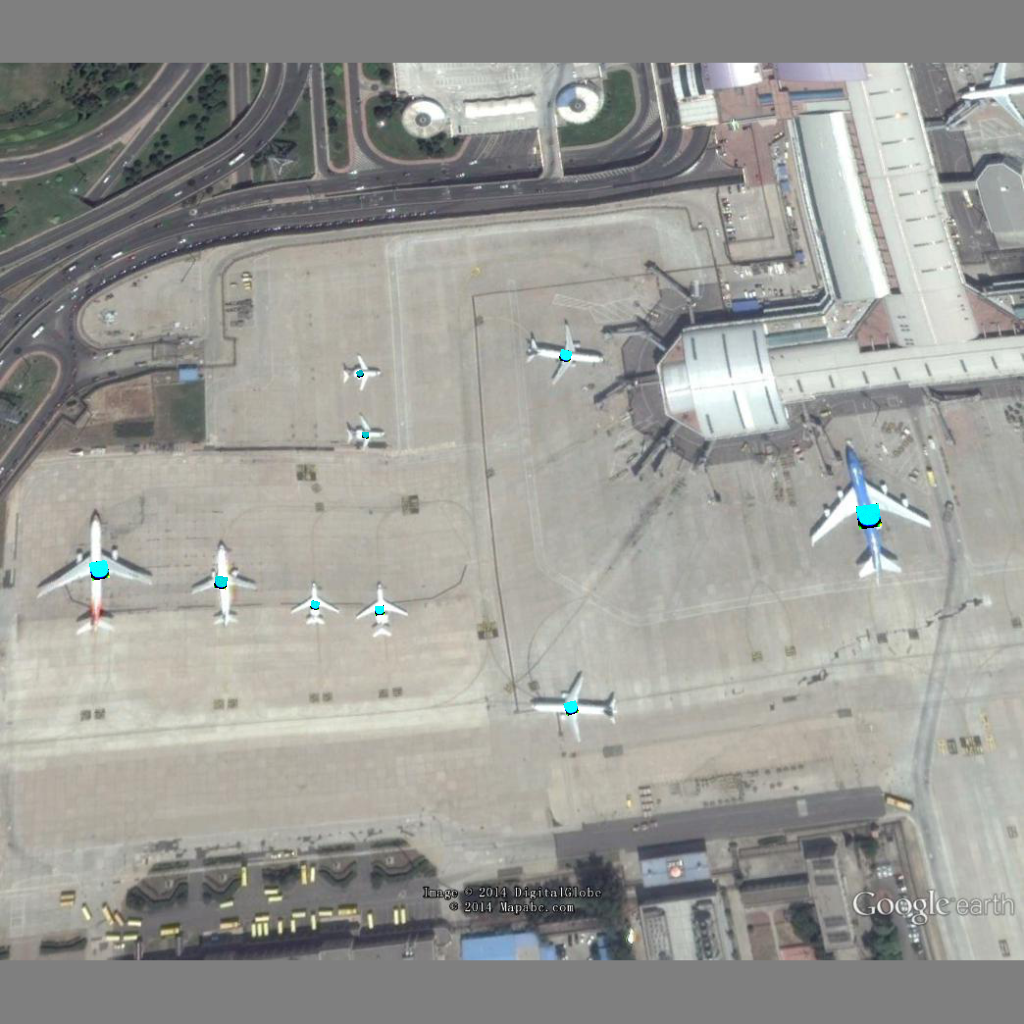}
    \captionsetup{font=small}
    \caption{On}
  \end{subfigure}
  \begin{subfigure}{0.32\linewidth}
    \includegraphics[width=1\linewidth]{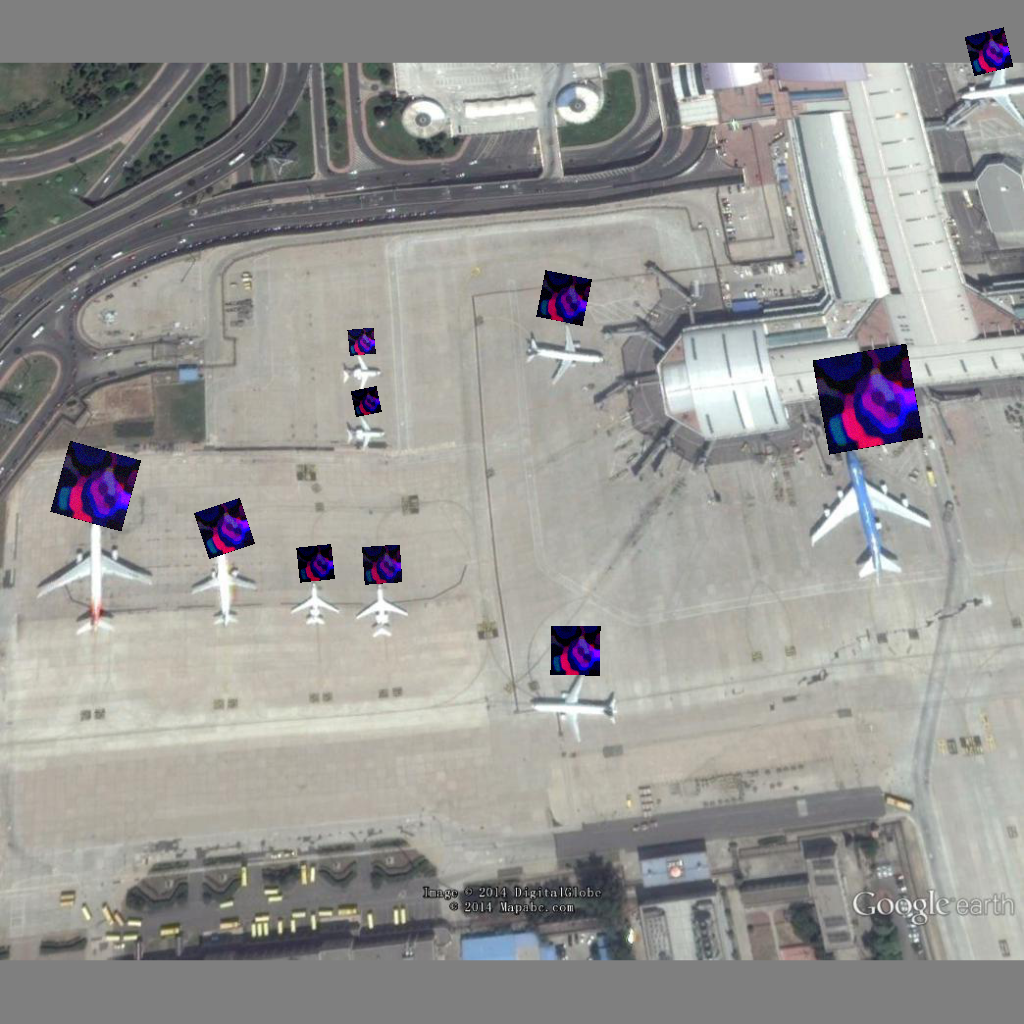}
    \captionsetup{font=small}
    \caption{Outside}
  \end{subfigure}
  \caption{Different patch settings in the digital domain.}
  \label{fig:digital_patch_settings}
\end{figure}

\begin{figure}
  \centering
    \begin{subfigure}{0.49\linewidth}
    \includegraphics[width=1\linewidth]{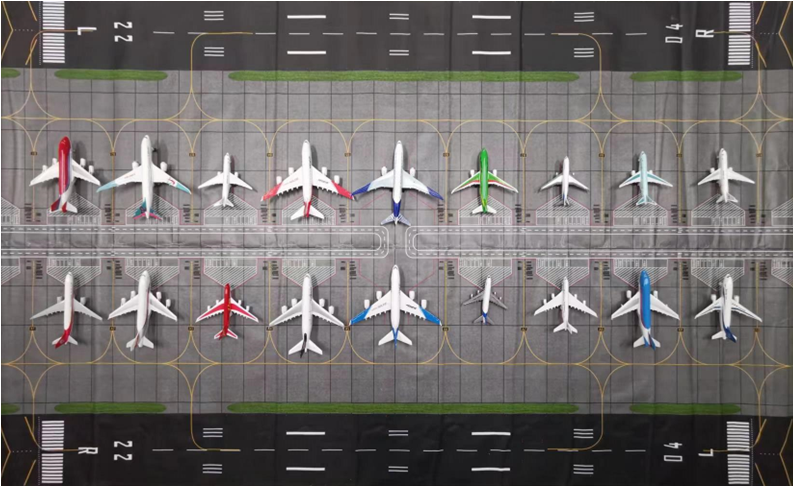}
    \caption{Clean}
  \end{subfigure}
  \begin{subfigure}{0.49\linewidth}
    \includegraphics[width=1\linewidth]{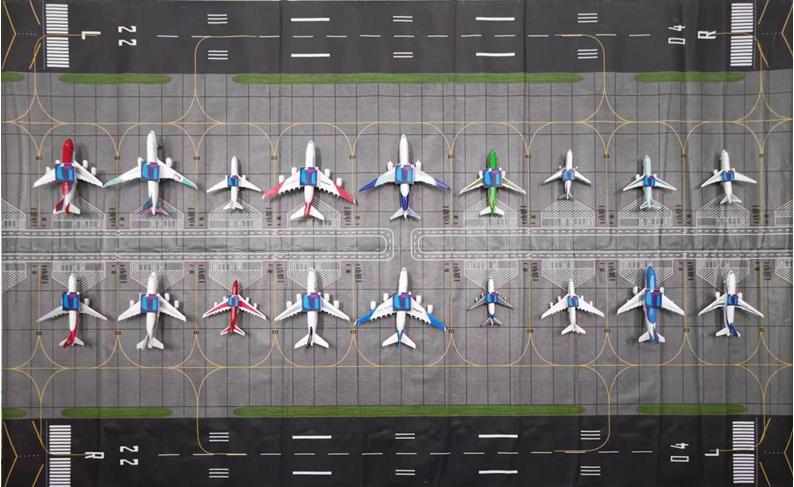}
    \caption{On}
  \end{subfigure}
  \begin{subfigure}{0.49\linewidth}
    \includegraphics[width=1\linewidth]{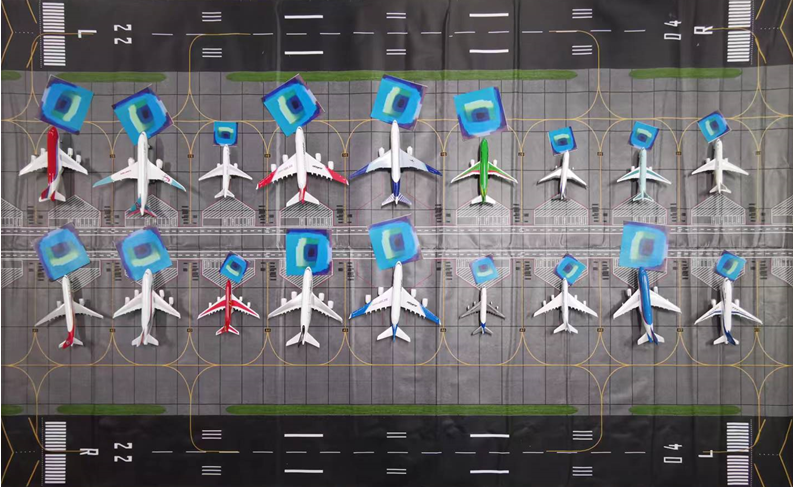}
    \caption{Outside}
  \end{subfigure}
    \begin{subfigure}{0.49\linewidth}
    \includegraphics[width=1\linewidth]{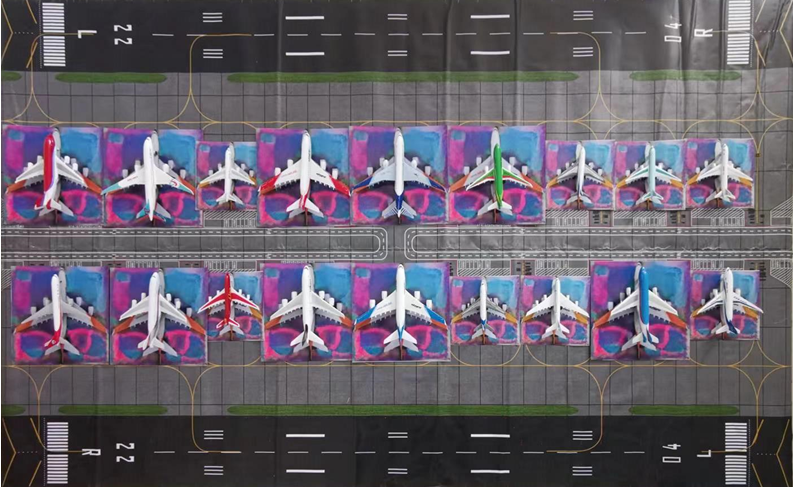}
    \caption{Background}
  \end{subfigure}
  \caption{Different patch settings in the physical domain. Note that the CBA performs digital attacks with patches outside targets same as training while performing physical attacks with patches in the background area.}
  \label{fig:physical_patch_settings}
\end{figure}

Benchmark on adversarial robustness for object detectors:

\ding{172} \textbf{Attacks} 

We evaluate adversarial robustness with 4 patch-based attacks, including CBA \cite{lian2023cba}, APPA (on) \cite{lian2022benchmarking}, APPA (outside) \cite{lian2022benchmarking}, and the method introduced by Thys \etal in \cite{thys2019fooling}.
Detailed information on these representatives and SOTA attacks against object detection is provided in Sec. \ref{Subsubsection2.3.2}.
In addition, we not only test the aforementioned SOTA methods under both white-box and black-box conditions but also conduct experiments in a different domain, \ie digital and physical domains, respectively.

\ding{173} \textbf{Datasets} 

\textbf{DOTA} \cite{xia2018dota} dataset is used for training the victim (white-box) or proxy (black-box) models, \ie the aerial detectors to be attacked, same as its role in Sec. \ref{Subsubsection3.1.2}.

\textbf{RSOD}\footnote{\url{https://github.com/RSIA-LIESMARS-WHU/RSOD-Dataset-}} is adopted to train the adversarial patches, which contains aircraft (4993 aircraft in 446 images), oil tank (1586 oil tanks in 165 images), playground (191 playgrounds in 189 images) and overpass (180 overpasses in 176 images).

In addition, we craft adversarial examples by adding adversarial patches generated by the aforementioned attack methods to perform digital attacks. The different patch settings for digital attacks are shown in Fig. \ref{fig:digital_patch_settings}.
For physical attacks, the elaborated adversarial patches are printed to disturb the targets of interest in the physical real-world scenarios. The different patch settings for physical attacks are shown in Fig. \ref{fig:physical_patch_settings}. 

\ding{174} \textbf{Models} are the same as the counterpart of natural robustness depicted in Sec. \ref{Subsubsection3.1.2}.

\ding{175} \textbf{Metrics} 

For \textbf{digital attacks}, we employ the detection results obtained from the clean images as the reference for calculating the AP. 
Specifically, the AP of the clean dataset is set as $100\%$ to ensure that targets missed by the original detector are not regarded as successful attacks.

For \textbf{physical attacks}, we conducted experiments scaled at a 1:400 proportion to verify the attack performance in the physical world. 
Technically, we trained 20 mainstream object detectors as victim or proxy models and recorded the average confidence of 18 aircraft, with the detection threshold set to 0.2. 
Targets with detection confidence lower than 0.2 are regarded as unrecognized because the confidence threshold of the object detection task is usually set to around 0.45. 

\section{Experiments}
\label{Section4}

In this part, we present the experimental results and deep analysis of benchmarking natural robustness and adversarial robustness in Sec. \ref{Subsection4.1} and Sec. \ref{Subsection4.2}, respectively.

\subsection{Natural Robustness}
\label{Subsection4.1}

\subsubsection{Image Classification}
\label{Subsubsection4.1.1}

\begin{figure}[!t]
  \centering
  \includegraphics[width=0.999\linewidth]{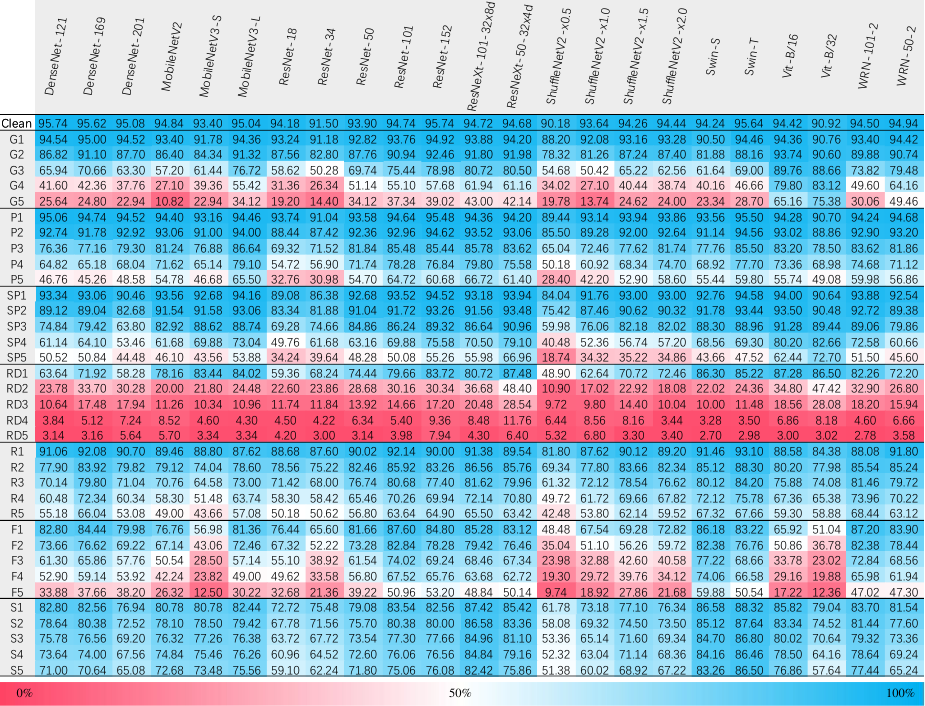}
  \caption{Benchmark on natural robustness of RSI classification.}
  \label{fig:natural_classification}
\end{figure}

\begin{figure}[!t]
  \centering
  \includegraphics[width=0.999\linewidth]{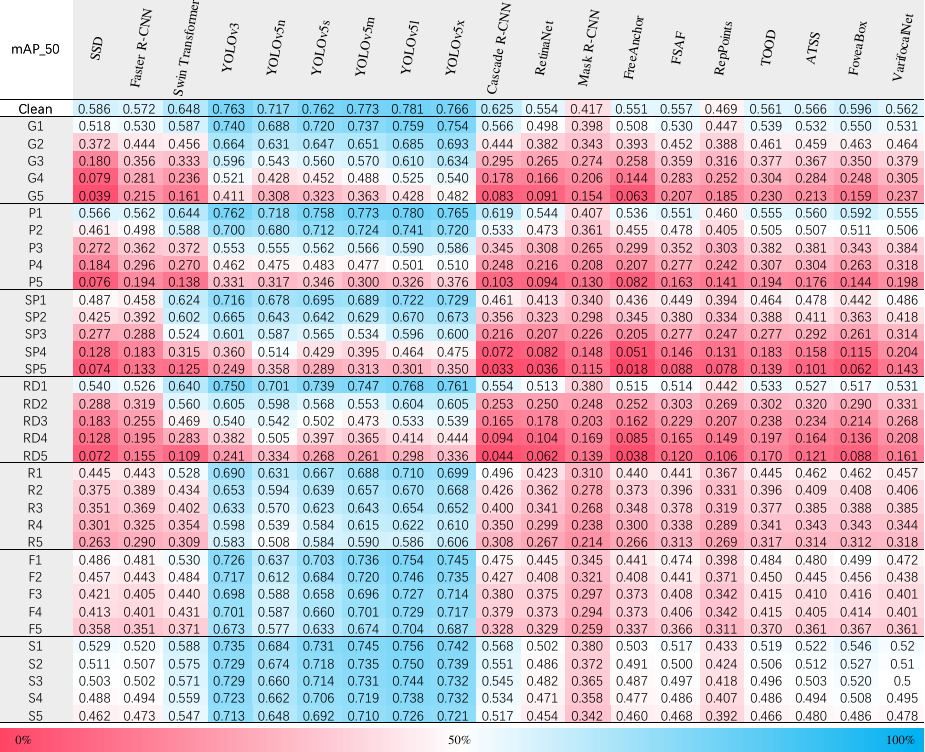}
  \caption{Benchmark on natural robustness of RS object detection (mAP@.50).}
  \label{fig:natural_detection_50}
\end{figure}

\begin{figure}[!t]
  \centering
  \includegraphics[width=0.999\linewidth]{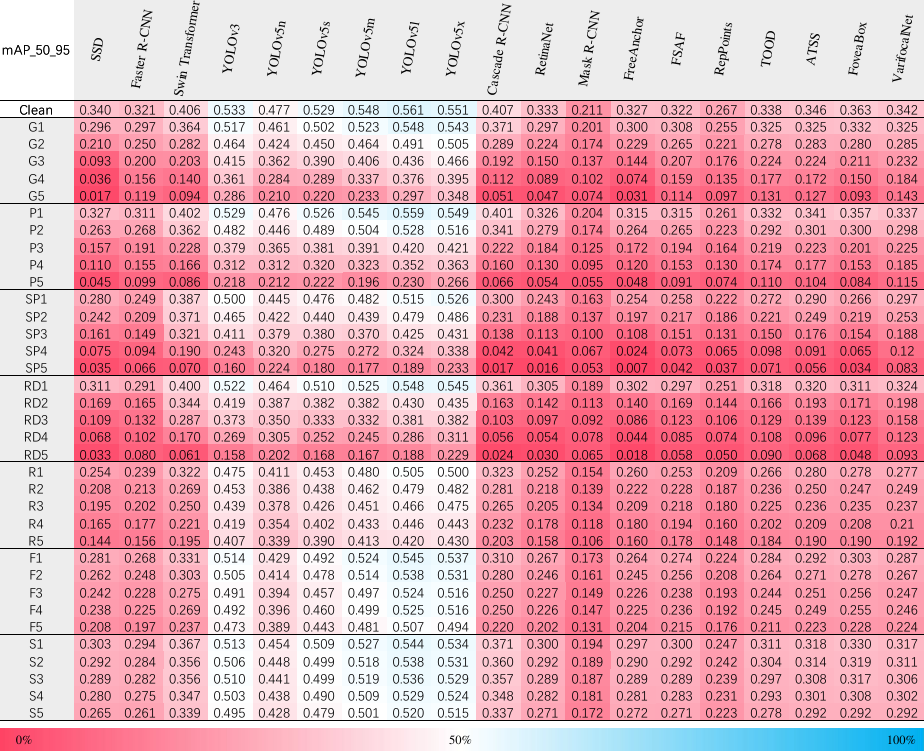}
  \caption{Benchmark on natural robustness of RS object detection (mAP@[.50:.05:0.95]).}
  \label{fig:natural_detection_50_90}
\end{figure}

In this section, we evaluate the natural robustness of the 23 RSI classifiers introduced in Sec. \ref{Subsubsection3.1.1} with AID RSI dataset \cite{xia2017aid} and its derived version with various natural noises.
We show the classification results in Fig. \ref{fig:natural_classification}.
Please note that all the evaluation results presented in this part represent the Top-1 accuracy.
Based on the experimental results, we have the following observations:
\begin{itemize}
    \item \textbf{Noise type.} The impact of various types of natural noise on classifiers exhibits varying degrees of influence. 
    Specifically, random noise exerts the most significant impact on classification accuracy, resulting in the greatest reduction in model performance. 
    In comparison, the classifiers are more robust to other noises.
    
    \item \textbf{Noise level.} As expected, for both CNNs and Transformers, an increase in the intensity of noise across all types results in a corresponding escalation of its impact on the model, thereby leading to a more pronounced reduction in classification accuracy.
    
    \item \textbf{Model type.} Evidently, when subjected to different types of noise, Transformers exhibit greater resilience compared to CNNs. 
    For CNNs, the robustness of lightweight networks (\ie MobileNet and ShuffleNet) is slightly lower than other models.
    
    \item \textbf{Model size.} When holding the model structure constant, it is apparent that larger models possess stronger robustness, such as ResNet, MobileNetv2, MobileNetv3, \textit{etc}. 
    However, it is important to note that there may be exceptions to this general trend, such as DenseNet, Swin, \etc., which could be attributed to overfitting.
\end{itemize}

\begin{figure*}[!t]
  \centering
  \includegraphics[width=0.999\linewidth]{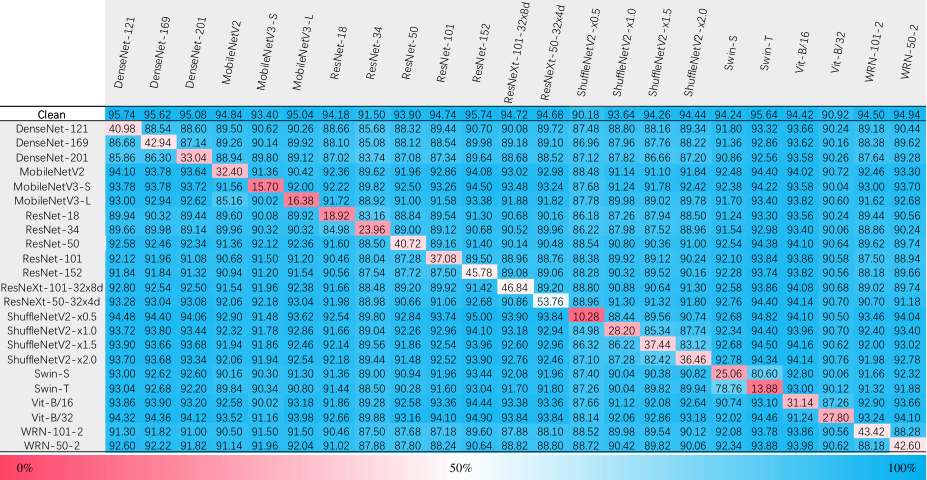}
  \caption{Benchmark on adversarial robustness of RSI classification with FGSM \cite{ian2015explaining}.}
  \label{fig:adversarial_classification_fgsm}
\end{figure*}

\begin{figure*}[!t]
  \centering
  \includegraphics[width=0.999\linewidth]{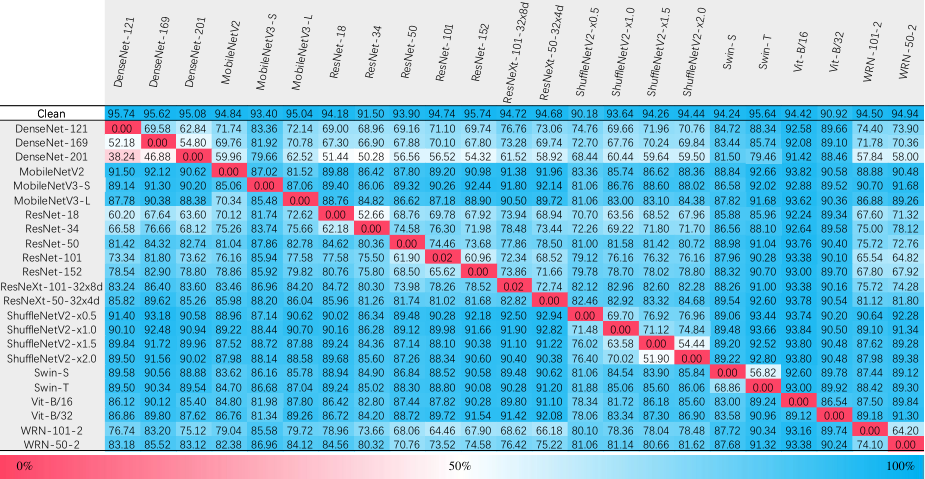}
  \caption{Benchmark on adversarial robustness of RSI classification with AA \cite{croce2020reliable}.}
  \label{fig:adversarial_classification_aa}
\end{figure*}

\begin{figure*}[!t]
  \centering
  \includegraphics[width=0.999\linewidth]{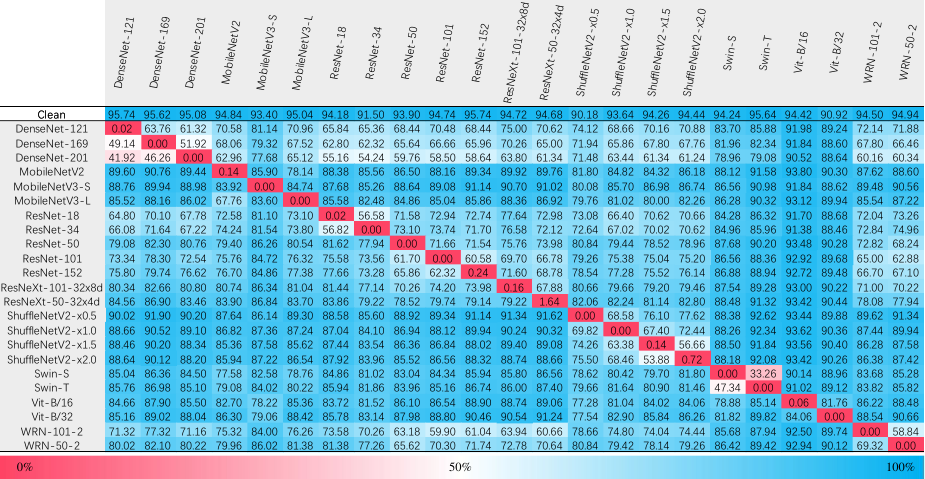}
  \caption{Benchmark on adversarial robustness of RSI classification with PGD \cite{madry2018towards}.}
  \label{fig:adversarial_classification_pgd}
\end{figure*}

\begin{figure*}[!t]
  \centering
  \includegraphics[width=0.999\linewidth]{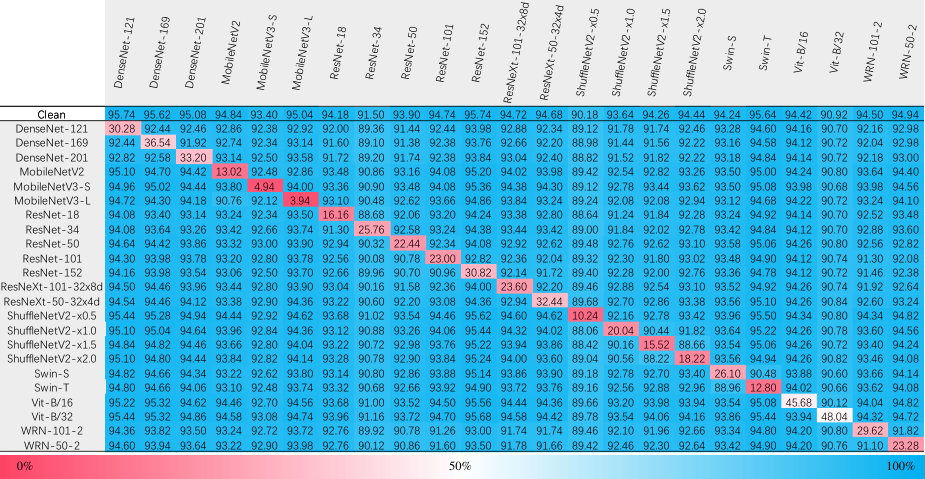}
  \caption{Benchmark on adversarial robustness of RSI classification with C\&W attack \cite{carlini2017towards}.}
  \label{fig:adversarial_classification_cw}
\end{figure*}

\begin{figure*}[!t]
  \centering
  \includegraphics[width=0.999\linewidth]{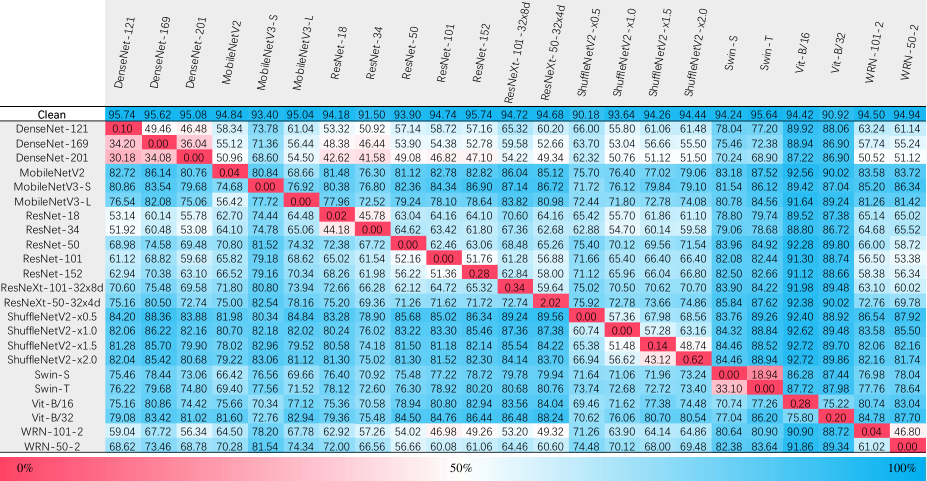}
  \caption{Benchmark on adversarial robustness of RSI classification with MIFGSM \cite{dong2018boosting}.}
  \label{fig:adversarial_classification_mifgsm}
\end{figure*}

\begin{figure*}[!t]
  \centering
  \includegraphics[width=0.999\linewidth]{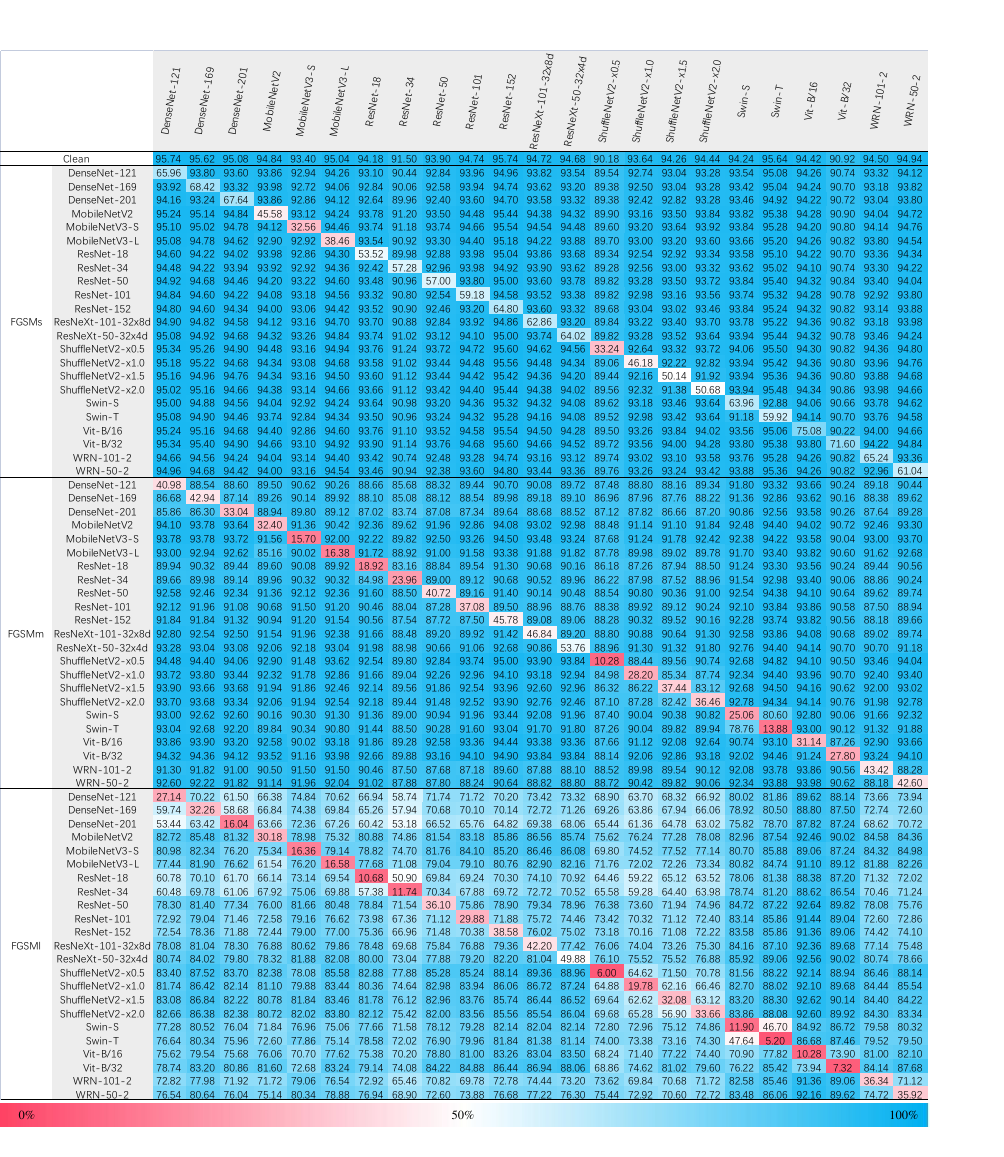}
  \caption{Benchmark on adversarial robustness of RSI classification with FGSM \cite{ian2015explaining} in different perturbation levels, \ie small, middle, and large.}
  \label{fig:adversarial_classification_fgsm_slm}
\end{figure*}

\subsubsection{Object Detection}
\label{Subsubsection4.1.2}

In this section, we evaluate the natural robustness of various mainstream RSI object detectors introduced in Sec. \ref{Subsubsection3.1.2} with large-scale aerial detection dataset DOTA and its derived version 
corrupted with various natural noises.
We show the experimental results in Fig. \ref{fig:natural_detection_50} and Fig. \ref{fig:natural_detection_50_90}.
Please note that we adopt mAP (mAP@.50 and mAP@[.50:.05:0.95]) as the evaluation metric.

Based on the experimental results, we have the following observations:
\begin{itemize}
    \item \textbf{Noise type.} Similarly, the influence of different types of natural noise on aerial detectors varies to a certain degree.
    Specifically, random noise and salt-pepper noise exert the most significant impact on aerial detectors, followed by Gaussian noise and Poisson noise. 
    In comparison, rain, snow, and fog have a relatively lesser impact on the performance of the detectors.
    
    \item \textbf{Noise level.} Consistent with expectations, all of the aerial detectors exhibit a consistent pattern: as the intensity of noise increases across all types, its impact on the detectors also intensifies, resulting in a more significant decline in detection performance.
    In comparison with natural weather noises, the level change of the remaining noises has a greater impact on the detection accuracy.
    
    \item \textbf{Model type.} Obviously, YOLOv5 and YOLOv3 are significantly more robust than other detectors and with better detection performance, followed by Swin Transformer, which is slightly more resilient than the rest aerial detectors.   
    In addition, it is hard to tell the difference between the robustness of different types of aerial detectors, such as CNN-based and Transformer-based, anchor-based and anchor-free, and one-stage and two-stage.
    
    \item \textbf{Model size.} Generally speaking, when the model structure is held constant, such as YOLOv5, it becomes evident that larger model sizes exhibit a greater level of robustness same as image classifiers. 
    However, in several cases, YOLOv5l (the second largest detector) outperforms YOLOv5x (the largest detector), overfitting may be a contributing factor to this phenomenon.
\end{itemize}

\subsection{Adversarial Robustness}
\label{Subsection4.2}

\subsubsection{Image Classification}
\label{Subsubsection4.2.1}

In this section, we evaluate the adversarial robustness of the 23 RSI classifiers introduced in Sec. \ref{Subsubsection3.2.1} with AID RSI dataset \cite{xia2017aid} and its derived version with various adversarial noises.
We show the classification attack results of FGSM, AutoAttack, PGD, C\&W, and MIFGSM in Fig. \ref{fig:adversarial_classification_fgsm},\ref{fig:adversarial_classification_aa}, \ref{fig:adversarial_classification_pgd}, \ref{fig:adversarial_classification_cw}, and \ref{fig:adversarial_classification_mifgsm}, respectively.
Moreover, we also conduct experiments on FGSM with different perturbation sizes, \ie small, middle, and large, as shown in Fig. \ref{fig:adversarial_classification_fgsm_slm}.
Please note that in the attack results figure, the diagonal position indicates the cases where the victim model and the proxy model are consistent, reflecting the results of white-box attacks. 
In contrast, the remaining positions in the figure correspond to the results of black-box attacks, where the victim model and the proxy model do not align.

Based on the experimental results, we have the following observations:
\begin{itemize}
    \item \textbf{Noise type.} The impact of various types of adversarial noise on classifiers exhibits varying degrees of influence. 
    Specifically, MIFGSM shows the best attack performance in both white-box and black-box conditions, followed by AutoAttack.
    In contrast, FGSM and C\&W attacks are found to have the least detrimental impact on classifiers, particularly in black box scenarios, rendering them nearly ineffective.
    
    \item \textbf{Noise level.} Consistent with expectations, both CNNs and Transformers demonstrate an increased attack efficacy to adversarial noise as its intensity escalates, regardless of whether the attacks are conducted in white-box or black-box settings. 
    Specifically, under white-box settings, FGSMs can successfully execute attacks, while the classification accuracy of most classifiers remains higher than 50\%. 
    However, in black-box conditions, FGSMs are found to be largely ineffective. 
    As the perturbation amplitude increases, the accuracy of most classifiers drops below 30\%, indicating a substantial reduction in classification accuracy. 
    Even under black-box attacks, the classification accuracy is moderately compromised to some extent.
    
    \item \textbf{Model type.} Transformers, such as Swin Transformer and ViT, exhibit a higher level of resilience compared to CNNs when facing various adversarial attacks, particularly in black-box scenarios. 
    This implies that perturbations trained on CNNs do not transfer well to Transformers, and vice versa. 
    For all classifiers, white-box attacks consistently outperform black-box attacks, and the generated perturbations demonstrate superior attack transferability across different versions of the training model. Specifically, perturbations trained on ResNet-50 exhibit good transferability to ResNet-18, ResNet-34, ResNet-101, and ResNet-152. 
    Furthermore, lightweight networks are easier to fool in white-box settings, while the corresponding perturbations exhibit lower attack transferability compared to other models.
    
    \item \textbf{Model size.} It is observed that model size does not affect attack efficacy in white-box conditions. 
    Under black-box settings, when keeping the classifiers' network structure constant, it is intuitive that the bigger the neural networks, the stronger the adversarial robustness. 
    However, it is important to note that the most robust model is usually not the biggest version of the classifiers but the second largest one, this phenomenon could be attributed to overfitting.
\end{itemize}

\subsubsection{Object Detection}
\label{Subsubsection4.2.2}

\begin{figure}
  \centering
  \begin{subfigure}{0.99\linewidth}
    \includegraphics[width=1\linewidth]{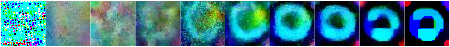}
  \end{subfigure}
  \caption{Visualization of patch evolution.}
  \label{fig:patch_evolution}
\end{figure}

\begin{figure}
  \centering
  \begin{subfigure}{0.685\linewidth}
    \includegraphics[width=1\linewidth]{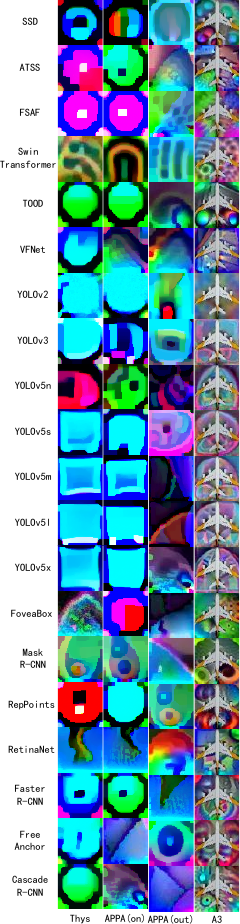}
  \end{subfigure}
  \caption{The generated adversarial patches.}  
  \label{fig:80patches}
\end{figure}

\begin{figure*}[!t]
  \centering
  \includegraphics[width=0.999\linewidth]{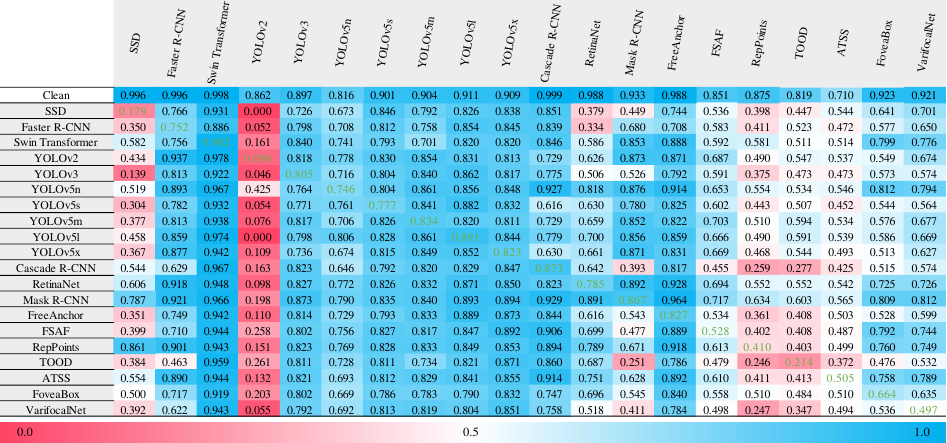}
  \caption{Benchmark on adversarial robustness of RS object detection with adversarial patches generated by \cite{thys2019fooling} in the physical world.}
  \label{fig:adversarial_detection_thys}
\end{figure*}

\begin{figure*}[!t]
  \centering
  \includegraphics[width=0.999\linewidth]{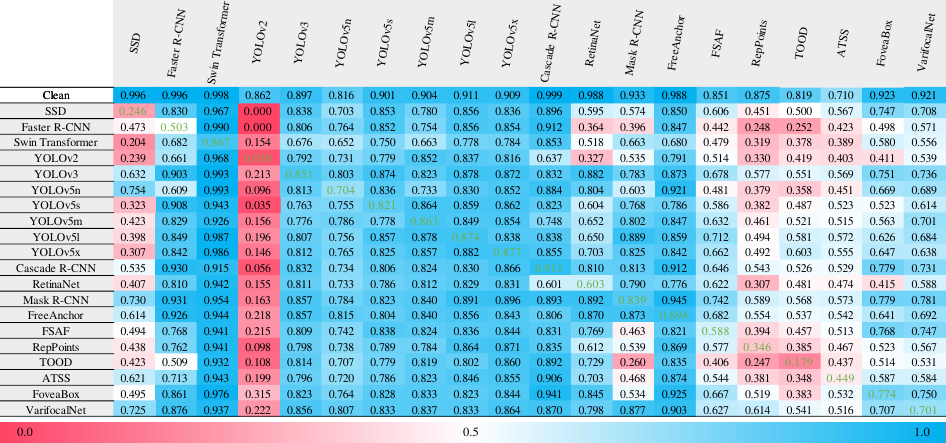}
  \caption{Benchmark on adversarial robustness of RS object detection with APPA (on) \cite{lian2022benchmarking} in the physical world.}
  \label{fig:adversarial_detection_appa_on}
\end{figure*}

\begin{figure*}[!t]
  \centering
  \includegraphics[width=0.999\linewidth]{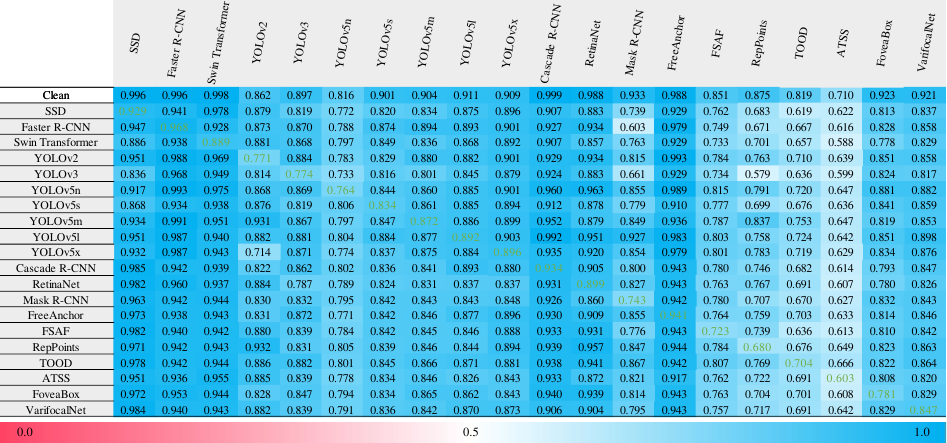}
  \caption{Benchmark on adversarial robustness of RS object detection with APPA (outside) \cite{lian2022benchmarking} in the physical world.}
  \label{fig:adversarial_detection_appa_outside}
\end{figure*}

\begin{figure*}[!t]
  \centering
  \includegraphics[width=0.999\linewidth]{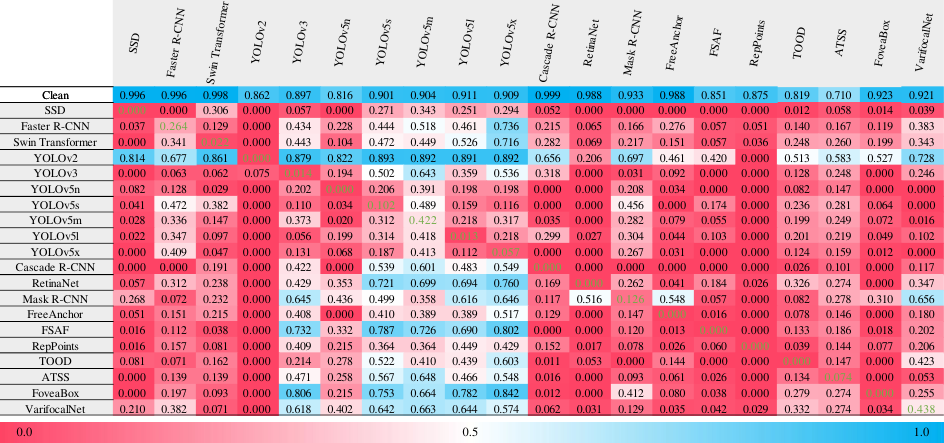}
  \caption{Benchmark on adversarial robustness of RS object detection with CBA \cite{lian2023cba} in the physical world.}
  \label{fig:adversarial_detection_cba}
\end{figure*}

\begin{figure}[!t]
  \centering
  \includegraphics[width=0.999\linewidth]{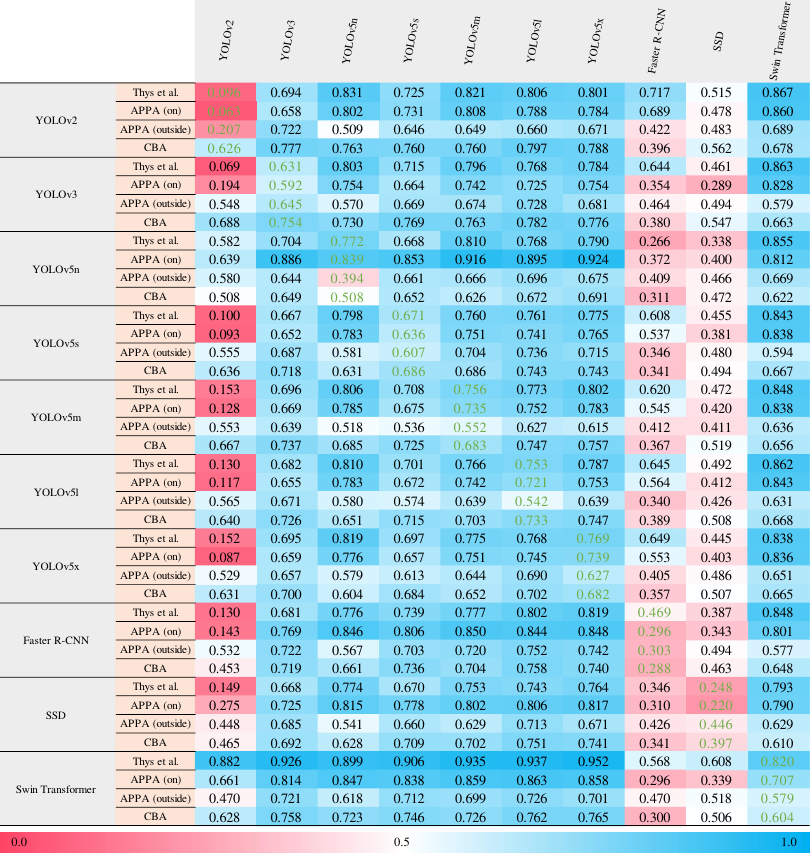}
  \caption{Benchmark on adversarial robustness of RS object detection in the digital domain.}
  \label{fig:adversarial_detection_digital}
\end{figure}

In this section, we evaluate the adversarial robustness of the 20 RSI object detectors introduced in Sec. \ref{Subsubsection3.1.2} with the DOTA RSI dataset \cite{xia2018dota} and its derived version with various adversarial noises.
We illustrate the evolutionary progression of the adversarial patch in Figure \ref{fig:patch_evolution}, demonstrating its dynamic development over time.
The generated adversarial patches are shown in Fig. \ref{fig:80patches}.
We show the detection attack results of four physical attack methods in Fig. \ref{fig:adversarial_detection_thys}, \ref{fig:adversarial_detection_appa_on}, \ref{fig:adversarial_detection_appa_outside}, and \ref{fig:adversarial_detection_cba}, respectively.
In addition, the digital attack performance is also exhibited in Fig. \ref{fig:adversarial_detection_digital}.
Evaluation metrics are introduced in detail in \ref{Subsubsection3.2.2}.
Please note that the experimental results presented in this section are partially derived from our previous works \cite{lian2022benchmarking} and \cite{lian2023cba}.

Based on the experimental results, we have the following observations:
\begin{itemize}
    \item \textbf{Digital attack.} 
    \ding{172} For attack methods, the attack effects of the four methods exhibit minimal variation in general, but for YOLOv2, the attack methods with the patch outside the target, \ie APPA(outside) and CBA, is less effective than the attack methods with the patch on the target, \ie \cite{thys2019fooling} proposed by Thys \etal and APPA(on).
    Notably, the background patch is positioned outside the targeted objects in the digital test, and a portion of the patch area is sacrificed to mask targeted objects in the physical world. 
    \ding{173} For detection methods, YOLOv2 is found to be the most vulnerable to attack, even in black-box scenarios. 
    On the other hand, different versions of YOLOv5 demonstrate robustness across diverse attack settings. 
    However, detectors such as Faster R-CNN and SSD are comparatively easier to be attacked and compromised. 
    In general, the Swin Transformer stands out as the most resilient detector, exhibiting a higher level of resistance against various attacks.
    
    \item \textbf{Physical attack.} 
    \ding{172} For attack methods, CBA exhibits a notable physical attack effect, causing a significant number of detectors to fail in detecting any objects in real-world scenarios, which is seldom observed for APPA and \cite{thys2019fooling}.
    In addition, CBA also shows the best attack transferability, even for some robust detectors, \eg YOLOv3, YOLOv5, Swin Transformers, \etc.
    \ding{173} For detection methods, YOLOv5 continues to demonstrate remarkable resilience against attacks compared to other aerial detectors. 
    However, CBA has proven highly effective in impairing the detection performance of YOLOv5 and shows strong generalization across different versions of YOLOv5.
    Similar to the digital attack scenario, YOLOv2 remains the most vulnerable detector in the physical world as well. 
    Interestingly, it exhibits a certain degree of immunity to adversarial patches placed outside the targets of interest.

    \item \textbf{White-box.} 
    \ding{172} The contextual background patches demonstrate a remarkable ability to completely impair the detection performance of several aerial detectors. 
    Specifically, detectors such as SSD, YOLOv2, YOLOv5n, Cascade R-CNN, RetinaNet, FreeAnchor, FSAF, RepPoints, TOOD, and FoveaBox are unable to recognize any of the protected targets when confronted with these patches.
    \ding{173} The remaining detectors are capable of correctly recognizing some of the protected objects; however, they exhibit a significantly lower average confidence level, averaging below 0.438.
    \ding{174} In contrast, the patches generated by APPA and \cite{thys2019fooling} have a smaller impact on misguiding the detectors, resulting in only a slight deviation in the confidence of correct detection. 

    \item \textbf{Black-box.} 
    \ding{172} Even in the black-box setting, the CBA demonstrates effective transferability of its attack efficacy across different aerial detectors, which significantly outperforms APPA and adversarial patches generated by \cite{thys2019fooling}.
    \ding{173} The CBA trained on YOLOv5n effectively safeguards all the targeted objects, preventing their recognition by YOLOv2, Cascade R-CNN, RetinaNet, FSAF, RepPoints, FoveaBox, and VFNet. 
    Other attacks, however, do not achieve the same level of success. 
    \ding{174} Under the attack of CBA, the average confidences of all detectors are below the threshold of 0.208, demonstrating its remarkable superiority over other physical attack methods.
\end{itemize}

\section{Discussions}
\label{Section5} 

The investigation into the robustness of DNNs has witnessed rapid advancements in recent years, particularly in the field of CV and its related applications such as RS. 
Despite the significant progress made, there remain several challenging issues that demand further examination and discussion. 
In this section, we delve into these challenges and offer insights into potential research directions of CV and RS as follows:
\begin{enumerate}
    \item \textbf{Explain the generation of adversarial perturbations with neural network training.}
    The process of training adversarial perturbations shares great similarities with training neural networks. 
    The key distinction lies in the update mechanism, where pixels within perturbations are adjusted during adversarial perturbation training, while network parameters are updated during network training. 
    Consequently, the generation of adversarial perturbations is influenced by various factors, including training samples, victim network models, and optimization strategies. 

    \item \textbf{Enlighten adversarial attacks with victim model.}
    The victim model plays a crucial role in determining the characteristics of the generated adversarial perturbations, particularly when the training samples and optimization process are fixed. 
    As a result, when targeting weak detectors like YOLOv2, an attack method may only learn limited information, which might be sufficient for white-box attacks, \ie successfully attack YOLOv2, but inadequate for targeting more robust models. 
    This analysis also sheds light on why adversarial patches trained on different versions of the same model exhibit similar pattern styles while differing from others.

    \item \textbf{Enlighten adversarial attacks with strategies for strengthening DNNs' performance.}
    Considering the similarities between perturbation generation and model training, it is worthwhile to explore the effective application of techniques that enhance the performance of DNNs in adversarial attacks. 
    For instance, methods such as "momentum" introduced in \cite{dong2018boosting} and "dropout" discussed in \cite{huang2022t} have shown the potential in boosting attack efficacy. 
    Investigating how these techniques, such as training strategies, test augmentations, and so on, can be appropriately utilized in the context of adversarial attacks could provide valuable insights for strengthening attack effectiveness, to further improve the security and resilience of DNN models.

    \item \textbf{Bridge the gap between digital and physical attacks.}
    The majority of existing research primarily concentrates on theoretical analyzes of adversarial attacks and their transferability in the digital domain, rendering them ineffective when confronted with real-world physical applications. 
    However, physical attacks raise substantial security concerns due to their potential implications in practical scenarios. 
    Therefore, it becomes imperative to bridge the gap between digital and physical attacks by developing techniques capable of effectively translating digital attack strategies into real-world settings.

    \item \textbf{Bridge the gap between attacks against different tasks.}
    The essence of DNNs-based models for visual perception is extracting features, progressing from shallow to deep concepts and from simple to abstract representations.
    As a consequence, how to interfere with the feature extraction process in various visual tasks to achieve a universal attack effect is an important and promising direction for research. 
    By understanding the underlying mechanisms of feature extraction in DNNs, researchers can develop strategies to manipulate and disrupt this process to generate effective adversarial attacks across different visual tasks. 
    This line of research has the potential to uncover vulnerabilities and weaknesses in DNN models, leading to the development of robust defense mechanisms and improved security in various applications of CV.

    \item \textbf{The background features matter more than you think.}
    The background features of a target are widely acknowledged to play a crucial role in its correct recognition. 
    However, recent studies \cite{lian2023cba,xu2022universalviaBackground} have demonstrated that intelligent recognition systems based on DNNs can be easily deceived solely by manipulating the background features of the target, even without distorting the target itself at all. 
    This raises the question of why a well-elaborated intelligent algorithm is so vulnerable to such manipulation. 
    It suggests that the influence of background features may be more significant than initially anticipated. 
    Consequently, there is a pressing need to delve deeper into the pivotal role that background features play in CV tasks and to understand their underlying mechanisms. 
    Such research can provide valuable insights to guide the design of more robust visual perception algorithms and models.

    \item \textbf{Background attack in the physical world.}
    The prevailing physical attacks directed at object detectors primarily focus on the development of perturbations in patch form. 
    These elaborated adversarial patches are printed and affixed to the surfaces of targeted objects through painting or pasting techniques, thereby compromising the recognition capabilities of intelligent systems operating in real-world environments. 
    However, the application of patch-based perturbations in the physical realm is accompanied by significant costs and time requirements. 
    As a viable alternative, background attacks emerge as a promising approach, wherein only the background regions surrounding the targeted objects are manipulated, without any direct alteration of the protected objects themselves. 
    This approach proves particularly advantageous for scenarios involving small targets, such as object detection in RS applications, where the effectiveness and practicality of adversarial patches are limited.
    Furthermore, the practical value of adversarial camouflage in background attacks is of utmost importance to ensure adversarial perturbations' inconspicuousness.

\end{enumerate}

\section{Conclusions}
\label{Section6} 

In this study, we present a comprehensive investigation into the robustness of image classification and object detection in the context of RS. 
Our work encompasses an extensive review of existing literature in both CV and RS domains, providing a comprehensive understanding of the research landscape in this area.
Furthermore, we perform a series of extensive experiments to benchmark the robustness of image classifiers and object detectors specifically designed for RS imagery. 
We also release the corresponding datasets with various types of noise to facilitate future research and evaluation in this field.
To the best of our knowledge, this study represents the first comprehensive review and benchmarking of the robustness of different tasks in optical RS.
Additionally, we conduct a deep analysis of the experimental results and outline potential future research directions to further enhance the understanding and development of model robustness.
Overall, our work offers a systematic perspective on the robustness of RS models, enabling readers to gain a comprehensive overview of this field and guiding the calibration of different approaches to accelerate the advancement of model robustness.
We also plan to continually update this work by incorporating more details and the latest advancements in the field, to enrich the benchmarking of model robustness in RS.

\bibliographystyle{IEEEtran}
\bibliography{references}



\end{document}